\pdfoutput=1

\documentclass[a4paper, 11pt, abstract=true, titlepage, oneside]{scrartcl}
\makeatletter

\usepackage[T1]{fontenc}
\usepackage[utf8]{inputenc}
\usepackage[onehalfspacing]{setspace}
\usepackage[headsepline]{scrlayer-scrpage}
\usepackage{layout}
\usepackage{geometry}
\usepackage{authblk}
\usepackage[titletoc,title]{appendix}
\usepackage[hyphens]{url}
\usepackage{color}

\usepackage{amsmath,amssymb,amsfonts,amsthm, mathtools, bm}
\usepackage{mathrsfs}
\usepackage{accents}
\usepackage{thmtools, thm-restate}
\usepackage{nicefrac}
\usepackage{array}

\usepackage{tabularx}
\usepackage{longtable}
\usepackage{booktabs}
\usepackage{colortbl}
\usepackage{multirow}
\usepackage{makecell}

\usepackage[bottom,hang,multiple]{footmisc}
\usepackage[acronym]{glossaries}

\usepackage{import}
\usepackage{soul}
\usepackage{graphicx}

\usepackage{placeins}

\usepackage{tikz}
\usetikzlibrary{arrows.meta}
\usetikzlibrary{math}
\usetikzlibrary{shapes}
\usetikzlibrary{patterns,shapes.arrows}
\usepackage{pgfplots}
\pgfplotsset{compat=1.18} 
\usepgfplotslibrary{groupplots}
\usepgfplotslibrary{groupplots,dateplot}
\pgfplotsset{compat=newest}
\usepgfplotslibrary{dateplot}
\usetikzlibrary{backgrounds}
\pgfdeclarelayer{background}
\pgfdeclarelayer{foreground}
\pgfsetlayers{background,main,foreground} 
\usepackage{tikzscale}

\usepackage{enumerate}
\usepackage{multicol}
\usepackage[author=Patrick]{pdfcomment}
\usepackage{xparse}
\usepackage{twoopt}
\usepackage{etoolbox}

\usepackage{eurosym}
\usepackage{ellipsis}

\usepackage{natbib}
\usepackage{paralist}
\usepackage{ulem}
\usepackage{overpic} 

\usepackage{soul}
\usepackage{threeparttable} 
\usepackage{physics}
\usepackage{algorithmic}
\usepackage[linesnumbered,ruled,vlined]{algorithm2e}
\usepackage{comment}
\usepackage{bbm}
\usepackage{adjustbox}   
\usepackage{import}      
\usepackage{subcaption}  
\usepackage{caption}
\usepackage{txfonts}

\usepackage{tikz}
\usetikzlibrary{positioning, arrows.meta, fit, calc}
\usepackage{relsize} 

\usepackage{multirow} 
\usepackage{amsfonts} 
\usepackage{float}

\AtBeginDocument{
	\newgeometry{
		textheight=9in,
		textwidth=6.5in,
		top=1in,
		headheight=16.49675pt,
		headsep=25pt,
		footskip=30pt
	}
}
\newsavebox{\abstractbox}
\renewenvironment{abstract}
{\begin{lrbox}{0}\begin{minipage}{\textwidth}
			\begin{center}\normalfont\sectfont\abstractname\end{center}\quotation}
		{\endquotation\end{minipage}\end{lrbox}%
	\global\setbox\abstractbox=\box0 }

\makeatletter
\expandafter\patchcmd\csname\string\maketitle\endcsname
{\vskip\z@\@plus3fill}
{\vskip\z@\@plus2fill\box\abstractbox\vskip\z@\@plus1fill}
{}{}
\makeatother

\allowdisplaybreaks

\addtokomafont{captionlabel}{\footnotesize\bfseries} 
\addtokomafont{caption}{\footnotesize\bfseries} 
\setkomafont{section}{\bfseries\Large}  
\setkomafont{subsection}{\itshape\large} 

\bibpunct[, ]{(}{)}{,}{a}{}{,}%
\def\newblock{\ }%
\newtheoremstyle{upright} 
  {3pt}                    
  {3pt}                    
  {\upshape}               
  {}                       
  {\bfseries}              
  {.}                      
  {.5em}                   
  {}                       

\theoremstyle{upright}

\newtheorem{lemma}{Lemma}

\let\theoremstyle\relax

\counterwithin*{equation}{section}

\newenvironment{myfont}{\fontfamily{ccr}\selectfont}{\par}
\DeclareTextFontCommand{\textmyfont}{\myfont}

\AtBeginDocument{%
	\abovedisplayskip=7.5pt plus 0pt minus 0pt
	\abovedisplayshortskip=7.5pt plus 0pt minus 0pt
	\belowdisplayskip=7.5pt plus 0pt minus 0pt
	\belowdisplayshortskip=7.5pt plus 0pt minus 0pt
}

\newcolumntype{L}[1]{>{\raggedright\let\newline\\\arraybackslash\hspace{0pt}}p{#1}}
\newcolumntype{C}[1]{>{\centering\let\newline\\\arraybackslash\hspace{0pt}}p{#1}}
\newcolumntype{R}[1]{>{\raggedleft\let\newline\\\arraybackslash\hspace{0pt}}p{#1}}
\renewcommand{\emph}[1]{\textit{#1}}

\definecolor{legendfill}{rgb}{0.9,0.9,0.9}  

 \tikzset{
    block/.style={draw, rectangle, minimum height=1cm, minimum width=2cm, align=center},
    arrow/.style={-Latex},
    line/.style={draw, -Latex},
    fontscale/.style = {font=\relsize{#1}}
}
\begin{document}
\emergencystretch 3em
\newacronym{acr:mnl}{MNL}{Multinomial Logit model}
\newacronym{acr:pd-vfa}{PD-VFA}{post-decision value function approximation}
\newacronym{acr:p-pd-vfa}{Pert. $\epsilon$}{PD-VFA using perturbed MNL parameters}
\newacronym{acr:p-pd-vfa-s}{Pert std $\sigma$}{}
\newacronym{acr:pp}{PP}{Reward percentage compensation}
\newacronym{acr:fp}{FP}{Formula-based compensation}
\newacronym{acr:rmse}{RMSE}{Root Mean Squared Error}
\newacronym{acr:drl}{DRL}{Deep Reinforcement Learning}
\newacronym{acr:mdp}{MDP}{Markov Decision Process}
\newacronym{acr:mbe}{MBE}{Mean Bias Error}


\title{\large Preference-aware compensation policies for crowdsourced on-demand services}

\author[1]{\normalsize Georgina Nouli}
\author[2]{\normalsize Axel Parmentier}
\author[1,3]{\normalsize Maximilian Schiffer}
\affil{\small 
	School of Management, Technical University of Munich, Germany

	\scriptsize georgina.nouli@tum.de,
	
	\small
\textsuperscript{2}
        CERMICS, École des Ponts, Marne-la-Vallée, France

        \scriptsize axel.parmentier@enpc.fr

 \small
	\textsuperscript{3}Munich Data Science Institute, Technical University of Munich, Germany
	
	\scriptsize schiffer@tum.de}

\date{}

\lehead{\pagemark}
\rohead{\pagemark}

\begin{abstract}
\begin{singlespace}
{\small\noindent Crowdsourced on-demand services offer benefits such as reduced costs, faster service fulfillment times, greater adaptability, and contributions to sustainable urban transportation in on-demand delivery contexts. However, the success of an on-demand platform that utilizes crowdsourcing relies on finding a compensation policy that strikes a balance between creating attractive offers for gig workers and ensuring profitability. In this work, we examine a dynamic pricing problem for an on-demand platform that sets request-specific compensation of gig workers in a discrete-time framework, where requests and workers arrive stochastically. The operator's goal is to determine a compensation policy that maximizes the total expected reward over the time horizon. Our approach introduces compensation strategies that explicitly account for gig worker request preferences. To achieve this, we employ the Multinomial Logit model to represent the acceptance probabilities of gig workers, and, as a result, derive an analytical solution that utilizes post-decision states. Subsequently, we integrate this solution into an approximate dynamic programming algorithm. We compare our algorithm against benchmark algorithms, including formula-based policies and an upper bound provided by the full information linear programming solution. Our algorithm demonstrates consistent performance across diverse settings, achieving improvements of at least 2.5–7.5\% in homogeneous gig worker populations and 9\% in heterogeneous populations over benchmarks, based on fully synthetic data. For real-world data, it surpasses benchmarks by 8\% in weak and 20\% in strong location preference scenarios.
\smallskip}
{\\ \footnotesize\noindent \textbf{Keywords:} Crowdsourced on-demand platforms, Approximate dynamic programming, Dynamic gig worker compensation}
\end{singlespace}
\end{abstract}

\maketitle
\section{Introduction}

\noindent In today's fast-paced world, consumers are increasingly time-sensitive when utilizing services, leading to substantial growth in on-demand service platforms. These platforms face the significant challenge of managing rapid fluctuations in demand and meeting the urgency of customer requests, while simultaneously keeping operational costs low. Within this context, platforms increasingly rely on using gig workers via crowdsourcing, i.e., independent workers that service on-demand tasks in exchange for a compensation, instead of employing traditional contracted personnel. Crowdsourcing has gained traction across various sectors, including last-mile delivery services such as grocery delivery (e.g., Instacart) and food delivery (e.g., DoorDash, UberEats), parcel delivery (e.g., Amazon Flex, UberFreight, UberRUSH, Roadie), and ride-hailing services (e.g., Uber, Lyft, Didi). Besides offering benefits in managing demand fluctuations \citep{luy2023strategic}, the utilization of gig workers is also appealing due to its potential to reduce costs \citep{fatehi2022crowdsourcing} and, in the case of crowdsourced delivery, it can contribute to sustainable urban logistics by minimizing emissions and road congestion \citep{yuen2023sustainable}. However, integrating gig workers into on-demand service models introduces complexity to the platform’s decision-making process, primarily due to the uncertainty in gig workers' availability and willingness to accept requests. In this context, the compensation offered by platforms plays a crucial role in influencing the likelihood of request acceptance \citep{barbosa2023data, bathke2023occasional}. Therefore, the success of a crowdsourcing platform relies on finding a compensation policy that balances two key objectives: creating offers attractive enough to engage gig workers, while also maintaining profitability for the platform. Achieving this balance is challenging as gig workers tend to show strong preferences based on specific request characteristics, such as the type of service, and the request's deadline. Additionally, gig workers with different characteristics, such as age and employment status can show different request preferences. Accordingly, it is essential for platforms to account for these preferences when designing compensation policies to ensure optimal outcomes. The goal of this study is to introduce compensation strategies for crowdsourced on-demand service platforms, which directly consider gig worker preferences in shaping offers for on-demand requests. 

\subsection{Related work}

Our work relates to two streams of research: studies on  gig worker preferences and behavior as well as work on dynamic pricing strategies for crowdsourced on-demand service platforms. In the following, we review both works concisely.

\noindent \textbf{Gig worker preferences}:
Multiple studies investigate gig worker preferences and behavior across various on-demand service platforms. In the context of crowd-shipping, where workers bid for delivery requests, \cite{ermagun2018bid} apply logistic regression to analyze how request characteristics influence attractiveness. They find that characteristics such as the size of the package, the type of service, the delivery deadline, and the distance of delivery are significantly correlated with a request's probability of receiving a bid. In a similar context, \cite{hou2023order} study the crowd-shipper's acceptance behavior by utilizing discrete choice models and machine learning models applied to survey data. Their findings suggest that factors such as gig workers’ age, income, and the compensation offered per request are key predictors of acceptance behavior. \cite{bathke2023occasional} perform a choice-based conjoint analysis to understand how various crowd-shipping delivery request attributes, in conjunction with gig worker characteristics, affect their willingness to accept requests. Their results indicate that delivery time and compensation are the primary drivers of gig worker decisions. Moreover, the influence of these factors varies across different worker demographics, such as age, and employment status. In the context of crowdsourced ride-sourcing platforms,  \cite{ashkrof2022ride} use a choice modeling approach based on the Random Utility Model (RUM) to analyze driver decisions using a driver-decision survey dataset of Uber and Lyft drivers. Their study highlights that driver-specific factors, including employment status, experience, and work shifts, significantly impact acceptance rates. Additionally, longer travel distances between a driver’s location and the pick-up point reduce the likelihood of job acceptance. In a similar context, \cite{xu2018empirical} utilize logistic regression on a dataset from a major ride-hailing platform in Beijing to show that driver behavior is strongly influenced by economic incentives, request characteristics (especially trip distance), and supply-demand intensities. 

This body of literature highlights the variety and complexity of factors influencing gig worker behavior as well as the variability of preferences. Together, these studies underscore the importance of incorporating gig worker preferences into the decision-making process to ensure effective compensation policies. Doing so remains the scope of this work.

\noindent \textbf{Pricing in crowdsourcing platforms}: A large body of literature exists on compensation strategies for crowdsourced on-demand service platforms.

\noindent \textit{Crowdsourced on-demand transportation platforms:} \cite{bai2019coordinating} study time-based pricing for an on-demand service platform with price and waiting time-sensitive requests and price-sensitive gig workers. \cite{sun2019optimal} consider ride-sourcing platforms and study how pricing per individual ride request can be optimized based on factors like distance and time, while considering that gig workers and customers maximize utility and therefore have the right to reject the platform's offer. \cite{cachon2017role} analyze pricing strategies for crowdsourcing platforms, focusing on how different contracts, such as fixed commission and surge pricing policies, affect provider participation and consumer welfare. They conclude that, while fixed pricing achieves near-optimal profits in most scenarios, surge pricing policies, can further enhance efficiency by better aligning supply with demand during peak periods. \cite{wang2022optimal} study dynamic pricing strategies in crowdsourcing logistics, using optimal control theory to maximize platform revenue considering fluctuating social delivery capacities and stochastic order demands. \cite{bimpikis2019spatial} study spatial pricing for ride-sharing platforms that serve requests on a network of locations. They find that maximum profit is achieved when the demand is balanced across all network locations and that using origin-specific ride prices can lead to significantly higher profits compared to using a fixed price across the network, especially when the demand pattern is highly unbalanced. Similarly, \cite{meskar2023spatio} optimize pricing and matching rates in ride-hailing platforms considering spatial and temporal network characteristics while also allowing the possibility of rejection from both the customer and the gig worker side. They also find that a balanced demand pattern across network locations yields maximum profit.

\noindent \textit{Crowdsourced last-mile delivery:}
\cite{yildiz2019service} present a stylized equilibrium model for optimizing service coverage and capacity planning in on-demand meal delivery, and derive results on the optimal gig worker compensation and service area. They study model variations where the probability of acceptance of gig workers is either fixed or dependent on the distance of the request's drop-off location from the restaurant. In the context of crowdsourced last-mile delivery, several studies consider joint decisions of gig worker compensation and routing for delivery requests. To this end, \cite{fatehi2022crowdsourcing} study crowdsourced last-mile delivery with guaranteed delivery time windows using robust optimization approaches based on robust queuing and routing theory. They derive analytical results for optimal hourly compensation and labor planning. Considering the use of a privately owned fleet, \cite{barbosa2023data} and \cite{silva2022deep} explore data-driven dynamic pricing schemes under uncertainty of gig workers' willingness to accept offers, focusing on last-mile delivery from a single store with gig workers who are in-store customers. The work of \cite{barbosa2023data} uses logistic regression on data collected through a questionnaire to model gig workers' willingness to undertake a request and develops a direct search algorithm to determine the optimal compensation. \cite{silva2022deep} use a \gls{acr:drl} approach to solve a two-stage optimization problem, where the first stage involves decisions on request fulfillment order and gig worker compensation, and the second stage involves routing decisions.  

Most of the existing literature on compensation strategies for on-demand service platforms either overlooks gig worker preferences and acceptance probabilities or, to the best of our knowledge, addresses them only in highly restricted or stylized settings, see e.g., \cite{yildiz2019service}, \cite{barbosa2023data} and \cite{silva2022deep}. However, ignoring gig worker preferences when designing compensation policies inevitably leads to suboptimal outcomes. Against this background, we develop an algorithmic paradigm for creating preference-aware compensation policies that can generally be applied in the context of crowdsourced on-demand service platforms. Furthermore, we aim to learn these preferences from data using statistical models, ensuring that the algorithm is both data-driven and scalable, making it suitable for real-world applications within dynamic and large-scale environments.

\subsection{Contributions}

In this paper, we propose a general algorithmic paradigm for computing preference-aware compensation policies to incentivize gig workers on on-demand service platforms to accept request offers. Our contributions are as follows: (1) We formalize the underlying compensation problem as a \gls{acr:mdp} and prove that the inner optimization problem of the Bellman equation can be solved optimally under the assumption that gig worker behavior follows a \gls{acr:mnl}. This allows us to derive a profit maximizing policy for the operator. (2) Based on the closed-form expression of this policy, we derive a practical algorithm for making compensation decisions, utilizing an approximate value iteration method that incorporates an \gls{acr:mnl} approximation learned in a predict-then-optimize fashion. (3) We present new benchmark datasets to evaluate the effectiveness of our algorithm and, through extensive numerical experiments, compare its performance against formula-based benchmark policies, a full-information upper bound, and our approach using a perfect \gls{acr:mnl} approximation.

Our experimental results underscore the critical role of incorporating gig worker preferences in achieving near-optimal results. In synthetic scenarios with a homogeneous gig worker population, our algorithm achieved performance ratios between 88.5\% and 94.7\% outperforming two benchmark policies by~\mbox{2.5-7.5\%}. In scenarios with a heterogeneous gig worker population, our approach achieved an average performance improvement of 9\% over benchmark policies. However, when gig worker preferences were not properly accounted for (i.e., using a single \gls{acr:mnl} for all groups), performance dropped significantly, confirming the necessity of accurately capturing worker heterogeneity to prevent revenue loss. In tests on real-world data simulating an on-demand ride-hailing platform, our algorithm achieved a performance improvement of 8\% with gig workers showing weak location preferences and 20\% with strong location preferences. We attribute this improvement largely to our algorithm being capable of avoiding the excessive compensation observed in the benchmark policies. Overall, these results demonstrate that a preference-aware approach is essential for balancing gig worker engagement and platform profitability, leading to superior performance across varied operational conditions.

\subsection{Organization}

The remainder of this paper is structured as follows. Section \ref{sec:prob_form} presents our problem setting and formulates it as an \gls{acr:mdp}, while Section \ref{sec:method} outlines our methodology for solving the  \gls{acr:mdp}. Section \ref{sec:exp_des} describes our experimental design while Section \ref{sec:res} presents the results of our experiments, including a comparative analysis of our algorithm against benchmark policies, sensitivity analysis over the gig worker utility estimation, and managerial insights. Finally, Section \ref{sec:conc} concludes our paper by summarizing its key findings and suggesting avenues for future work.

\section{Problem formulation}
\label{sec:prob_form}
We consider an on-demand service platform which operates on a discrete time horizon consisting of $T$ decision time steps $\{1,2,...,T\}$. Within this time horizon, on-demand requests and gig workers arrive in the system stochastically. Each gig worker is willing to serve requests in exchange for a fee. The decision to accept a request is based on the gig worker's individual utility function, which is unknown to the platform operator. Generally, a higher compensation increases the likelihood of a gig worker accepting a request. Gig workers have the flexibility to choose any of the offered requests or reject all offers. The operator’s task is to set a compensation for each offered request, with the objective of maximizing net profit over the complete time horizon.

\subsection{Problem formulation as an \gls{acr:mdp}}
\begin{figure}[t!]%
    \centering
    \fontsize{10}{10}\selectfont
    \includegraphics[width=15cm]{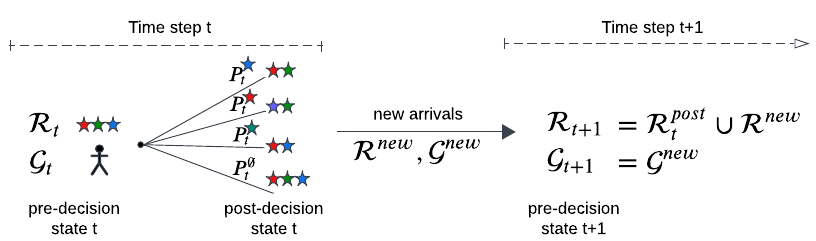}
    \caption{\textnormal{State transitions.}}
    \label{fig:state_trans}%
\end{figure}

The problem setting outlined above unfolds as an \gls{acr:mdp} as follows:\\
\noindent \underline{System State:} 
At any decision time step $t \in \{1,...,T\}$ the system state  $S\hspace{-0.1em}_t = (\mathcal{R}\hspace{0.05em}_{t},\mathcal{G}_t)$ is a pair where $\mathcal{R}\hspace{0.05em}_t$ is the set of active on-demand requests and $\mathcal{G}_t$ is the set of available gig workers. For simplicity, we will refer to $\mathcal{R}\hspace{0.05em}_{t}$ and $\mathcal{G}_t$ as the request and gig worker states, correspondingly. We assume that the gig worker state $\mathcal{G}_t$ can accommodate at most one gig worker and that they remain in the system for only one time step.  \\
\noindent \underline{On-demand requests:} 
We describe each on-demand request $i \in \mathcal{R}\hspace{0.05em}_{t}$ by a set of characteristics $\mathbf{x}_i =(x_{i1},...,x_{ik})$, a reward $r_i>0$ which the operator will receive for servicing the request, an expiration step $t_i^{\mathrm{exp}}$ after which the request will leave the system without being serviced and a penalty $\beta_i<0$ which the operator has to pay if the request is not serviced. \\
\noindent \underline{Gig workers:} 
Gig workers are willing to serve a single on-demand request in return for some compensation~$c_i$. The willingness of gig worker $j \in \mathcal{G}_t$ to service an on-demand request $i \in \mathcal{R}\hspace{0.05em}_{t}$ is stochastic and can be described by a stochastic function:
\begin{align}
f_{ij} = h_j(\mathbf{x}_i,c_i) + e_{ij} \label{eq:util}
\end{align}
where $\mathbf{x}_i = (x_{i1}, ..., x_{ik})$ is a feature vector that describes the characteristics of request $i \in \mathcal{R}\hspace{0.05em}_{t}$, $c_i$ is the compensation for servicing request $i$, and $e_{ij}$ is a noise variable. This function is unknown to the system operator. \\
\noindent \underline{Feasible Decisions:} 
At every decision time step $t \in \{1,...,T\}$ the central operator must offer a compensation~$c_i$ to a gig worker for servicing each request $i \in \mathcal{R}\hspace{0.05em}_{t}$. A feasible compensation for servicing a request~$i \in \mathcal{R}\hspace{0.05em}_{t}$ must be non-negative ($c_i \geq 0$). We define the set of feasible compensations for a state $S\hspace{-0.1em}_t$ as:
\begin{align}
\mathcal{C}(S\hspace{-0.1em}_t) = \{\mathbf{c_t} = (c_i)_{i \in \mathcal{R}\hspace{0.05em}_{t}} :  c_i \geq 0  \ \ \forall i \in \mathcal{R}\hspace{0.05em}_{t} \}
\end{align} \\
\noindent \underline{Evolution of the system:} 
At each decision time step $t \in \{1,...,T\}$, the operator observes the system state $S\hspace{-0.1em}_t = (\mathcal{R}\hspace{0.05em}_{t},\mathcal{G}_t)$. If a gig worker is in the system, the central operator decides on a compensation $c_i$ for each request $i \in \mathcal{R}\hspace{0.05em}_{t}$ which will be offered to the gig worker as a compensation for fulfilling the request. Subsequently, the gig worker receives these offers and either selects to service the offer which maximizes their utility or, in case all offers have low utility, rejects all offers. We denote by $P^i_{t}(\mathbf{c_t})$ the probability of a gig worker accepting request $i$ for compensation decision $\mathbf{c_t} = (c^i_t)_{i \in \mathcal{R}\hspace{0.05em}_{t}}$ in system state $\mathcal{R}\hspace{0.05em}_{t}$. We denote by $P^\emptyset_{t}(\mathbf{c_t})$ the alternative where the gig worker does not select any request. In case the gig worker selects a request, the gig worker receives a compensation $c_i$, the operator receives a reward of $r_i - c_i$, and both the gig worker and the request leave the system. If the gig worker did not select any offer, the gig worker leaves the system without compensation. On-demand requests that expire in $t$, i.e., such that $t_i^{\mathrm{exp}} = t$, also leave the system and the operator receives the corresponding penalties $\beta_i$. So far, the system has evolved due to the realization of the operator's compensation decision and the gig worker's acceptance decision reaching a post decision state. We refer to the \textit{post-decision state} at time step $t$ as the intermediate state after a gig worker has decided which request offer to accept and after removing expired requests, but before new requests and gig workers arrive. We define the post-decision state $\mathcal{R}\hspace{0.05em}_t^{p}$ as a subset of the request state $\mathcal{R}\hspace{0.05em}_{t}$. To this end, we note that, the probabilities $P^i_{t}(\mathbf{c_t})$ and $P^\emptyset_{t}(\mathbf{c_t})$ establish a probability distribution for the transition into post-decision states. Figure~\ref{fig:state_trans} illustrates the state transitions, including the distribution over the post-decision states. The system then evolves between decision states $t$ and $t+1$ as new requests $\mathcal{R}^{\mathrm{new}}$ and new gig workers $\mathcal{G}^{\mathrm{new}}$ arrive, and transitions into the next state~$S\hspace{-0.1em}_{t+1}$ (see Figure~\ref{fig:state_trans}). Lastly, we assume that all requests expire by the last decision time step~$T$.

\noindent \underline{Expected immediate reward:} 
The expected immediate reward for the operator at decision time step $t$ for state $S\hspace{-0.1em}_t = (\mathcal{R}\hspace{0.05em}_{t},\mathcal{G}_t)$ when considering a compensation decision $\mathbf{c_t} = (c^i_t)_{i \in \mathcal{R}\hspace{0.05em}_{t}}$ reads:
\begin{align}
R_t(\mathbf{c_t}) = & \mathbbm{1}_{|\mathcal{G}_t|=1} [\sum_{i \in \mathcal{R}\hspace{0.05em}_{t}}{P^i_{t}(\mathbf{c_t})(r_i - c_t^i - \beta_i \mathbbm{1}_{i \in \mathcal{R}\hspace{0.05em}_t^{\mathrm{exp}}}}) ] + \! \sum_{i \in \mathcal{R}\hspace{0.05em}_t^{\mathrm{exp}}}\!\beta_i
\label{eq:imm_rew}
\end{align}
\noindent where $r_i - c_t^i$ is the profit for request $i$ being serviced for a compensation of $c_t^i$. The term $\sum_{i \in \mathcal{R}\hspace{0.05em}_t^{\mathrm{exp}}}\!\beta_i$ accounts for all penalties that result for all expiring requests while the term $- \beta_i \mathbbm{1}_{i \in \mathcal{R}\hspace{0.05em}_t^{\mathrm{exp}}}$ ensures that we omit the penalty of an expiring request if it gets accepted. \\
\noindent \underline{Compensation policy:} 
We assume that the operator takes decisions based on a deterministic compensation policy which we define as a function $\pi_p: S\hspace{-0.1em}_t \mapsto \pi_p(S\hspace{-0.1em}_t)$ that maps any state $S\hspace{-0.1em}_t$ to a feasible compensation decision $c \in \mathcal{C}(S\hspace{-0.1em}_t)$. \\
\noindent \underline{Objective:} 
We aim to study an online learning problem from the perspective of the central operator which is responsible for assigning compensations to on-demand requests. Our objective is to find a compensation policy $\pi_p$ that maximizes the expected reward for the complete time horizon:
\begin{equation}
  \max_{\pi_p}\mathbb{E}\left(\sum_{t=1}^{T}{R_t(\mathbf{c_t})}|\pi_p\right)
\end{equation} where $R_t(\mathbf{c_t})$ is the expected immediate reward at decision time step $t$ for compensation decision $\mathbf{c_t}$.

\subsection{Discussion}

Three comments on our problem are in order. First, our model processes offers by considering one gig worker at a time. The assumption that $\mathcal{G}_t$ can accommodate at most one gig worker can be justified in two ways. First, by adjusting the frequency of decision time steps so that at most one gig worker arrives per step, or second, by allowing multiple gig workers to arrive simultaneously and queuing them for sequential consideration in subsequent steps. Formally, let $\mathcal{Q}$ represent the queue of arriving gig workers. At each time step $t$, all newly arrived gig workers are added to $\mathcal{Q}$ for processing. The operator then observes the first gig worker in the queue (FIFO) and includes them in $\mathcal{G}_t$. If $\mathcal{Q}$ is not empty, the operator will observe the next gig worker in the queue in the next time step. This sequential processing ensures that $\mathcal{G}_t$ contains at most one gig worker at any time step.
Second, we assume gig workers accept one request at a time in exchange for a fee, which aligns with many real-world applications. However, an alternative approach could involve offering bundles of requests to gig workers. Extending the problem to include bundles instead of individual offers would introduce additional complexity due to the combinatorial nature of the decision-making process, making the corresponding \gls{acr:mdp} more challenging to solve. Nonetheless, we believe that with proper algorithm implementation and an efficient \gls{acr:mnl} model estimator, handling bundles of requests could be a feasible extension.
Lastly, our methodology focuses on a deterministic compensation policy, however, the platform operator could alternatively opt for a stochastic compensation policy. We opted for a deterministic policy to ensure consistency and predictability for gig workers. This consistency is particularly valuable in a practical context, as it fosters trust and transparency, reducing uncertainty for workers and enabling more reliable income expectations.

\section{Methodology}
\label{sec:method}

Solving the \gls{acr:mdp} as defined in Section \ref{sec:prob_form} presents several key challenges from a dynamic decision-making perspective. A primary challenge is the need for a model that describes the environment's transition dynamics, which is often complex and challenging to characterize. Another significant challenge is the intractable state space arising from the stochastic arrival of requests and gig workers, which renders exhaustive computation of optimal policies infeasible. Therefore, effective generalization across states is essential, as explicitly enumerating all possible states is unrealistic in many real-world scenarios.

Our proposed method offers several solutions to address these challenges. First, we utilize the \gls{acr:mnl} model to describe gig workers' decision-making process, capturing essential transition dynamics without requiring a fully known environment model. To manage the complexity of large state spaces, we leverage a post-decision state formulation, which eases computing optimal policies from a tractability perspective. This formulation also facilitates solving the inner optimization problem and deriving an exact form for optimal prices based on the post-decision state values. Our method is designed to ensure scalability, making it suitable for real-world applications where exhaustive enumeration of all possible states is infeasible. To this end, we incorporate value function approximators to generalize effectively across states, which enables us to estimate value functions without exhaustive state enumeration.  

This section details our approach. Section \ref{sec:bellman} introduces the Bellman equation for the \gls{acr:mdp} as defined in Section~\ref{sec:prob_form}. Section \ref{sec:mnl} discusses the \gls{acr:mnl} model for capturing essential transition dynamics, addresses the state space intractability by deriving the post-decision state formulation and presents an analytical solution to the optimization problem with optimal prices based on post-decision state values. Section \ref{sec:statistical_models} presents value function approximators and an algorithm for learning appropriate parameterizations. Section \ref{sec:training} details the statistical model used for gig worker utility estimation and presents the training procedure for the full algorithm. Finally, Section \ref{sec:meth_as} discusses methodological assumptions. 

\subsection{Bellman equation}
\label{sec:bellman}
In the following, we derive the Bellman equation for the \gls{acr:mdp} as defined in Section \ref{sec:prob_form}. To do so, we first elaborate on optimal value functions in general to elucidate the connection between pre- and post-decision states, which then allows us to derive the pre- and post-decision state value function accordingly. 

\noindent \textit{Optimal value functions:} The optimal value function of a pre-decision state $S\hspace{-0.1em}_t = (\mathcal{R}\hspace{0.05em}_{t},\mathcal{G}_t)$ or a post-decision state $\mathcal{R}\hspace{0.05em}_t^{p}$, denoted as $V_t(\mathcal{R}\hspace{0.05em}_{t},\mathcal{G}_t)$ and $V^{\mathrm{p}}_{t}(\mathcal{R}\hspace{0.05em}_{t}^{p})$, describes the maximum expected reward achievable from that state onward by following the optimal policy. In this context, the transition from a post-decision state to a successor pre-decision state exclusively involves the stochastic realization of the arrival processes of on-demand requests and gig workers. Therefore, a straightforward relationship between the two value functions is the following: 
\begin{align}
V^{\mathrm{p}}_{t}(\mathcal{R}\hspace{0.05em}_t^{p}) = \mathbb{E}_{\mathcal{R}^{\mathrm{new}},\mathcal{G}^{\mathrm{new}}}[V_{t+1}(\mathcal{R}\hspace{0.05em}_t^{p} \cup \mathcal{R}^{\mathrm{new}},\mathcal{G}^{\mathrm{new}})] \label{eq:value_function}
\end{align}
where the expectation over $\mathcal{R}^{\mathrm{new}},\mathcal{G}^{\mathrm{new}}$ indicates the dynamics of new request and new gig worker arrivals in the system correspondingly (see Figure~\ref{fig:state_trans}). Essentially, Equation \ref{eq:value_function} states that the value of a post-decision state $\mathcal{R}\hspace{0.05em}_{t}$ is equal to the expected value over all possible successor pre-decision states. 

\noindent \textit{Bellman equation using the pre-decision state value function:}
The classic definition of the Bellman equation of the pre-decision state $S\hspace{-0.1em}_t = (\mathcal{R}\hspace{0.05em}_{t},\mathcal{G}_t)$ at time step $t = 0,1,...,T$ reads: 
\begin{align}
V_t(S\hspace{-0.1em}_t) & = \max_{\mathbf{c_t}} \{  R_t(\mathbf{c_t}) + \mathbb{E}_{S' \sim P(\cdot|S\hspace{-0.1em}_t,\mathbf{c_t})}[V_{t+1}(S')] \}
\end{align}
where $R_t(\mathbf{c_t})$ accounts for the expected immediate reward of being in state $S\hspace{-0.1em}_t = (\mathcal{R}\hspace{0.05em}_{t},\mathcal{G}_t)$ and following action $\mathbf{c_t}$, while the expectation $\mathbb{E}_{S' \sim P(\cdot|s,\mathbf{c_t})}[V_{t+1}(S')]$ accounts for the expected future reward that can be obtained from successor pre-decision states $S'$. In this standard definition, the value of each pre-decision state is decomposed using the value of the successor pre-decision state. Notably, the transitions from one pre-decision state to the next include two different sources of stochasticity. The first source of stochasticity results from the utility of the gig workers while the second source results from the stochastic nature of the request and gig worker arrivals. 
This formulation of the Bellman equation is highly intractable due to the complexities introduced by both sources of stochasticity. The primary challenge arises from the second source of stochasticity, as the arrival of new requests and gig workers leads to an explosion in the state transition space, making it impossible to model or track transitions to the next state accurately. In contrast, the first source of stochasticity, which depends solely on the gig workers' choice model, is less complex. To address this issue, we can alternatively formulate the Bellman equation using the post-decision state value function.

\noindent \textit{Bellman equation using the post-decision state value function:} Let us denote as $\mathcal{R}\hspace{0.05em}_t^{\mathrm{exp}}$ the set of requests in~$\mathcal{R}\hspace{0.05em}_{t}$ that expire, and let $\mathcal{R}\hspace{0.05em}_t^{'}$ correspond to the set of requests in $\mathcal{R}\hspace{0.05em}_t$ that do not expire after time step $t$, i.e., $\mathcal{R}\hspace{0.05em}_t^{'} = \mathcal{R}\hspace{0.05em}_{t} \setminus \mathcal{R}\hspace{0.05em}_t^{\mathrm{exp}}$. Furthermore, from here on, we assume for ease of notation that $\mathcal{R}\hspace{0.05em}_t^{'} \setminus \{\emptyset\}$ is equivalent to $\mathcal{R}\hspace{0.05em}_t^{'} $. Then, the Bellman equation of the pre-decision state $S\hspace{-0.1em}_t = (\mathcal{R}\hspace{0.05em}_{t},\mathcal{G}_t)$ at time step $t = 0,1,...,T$ is given by: 
\begin{align}
V_t(S\hspace{-0.1em}_t) & = \max_{\mathbf{c_t}} \big\{  R_t(\mathbf{c_t}) + \underbrace{(1 - \mathbbm{1}_{|\mathcal{G}_t|=1})  V^{\mathrm{p}}_{t}(\mathcal{R}\hspace{0.05em}_t^{'})\vphantom{\sum_{i \in \mathcal{R}\hspace{0.05em}_{t} \cup \{\emptyset\}}}}_{\text{(I)}}  + \underbrace{\mathbbm{1}_{|\mathcal{G}_t|=1} \! \sum_{i \in \mathcal{R}\hspace{0.05em}_{t} \cup \{\emptyset\}} \! P^i_{t}(\mathbf{c_t}) V^{\mathrm{p}}_{t}(\mathcal{R}\hspace{0.05em}_t^{'} \setminus \{i\})}_{\text{(II)}}  \big\}
\label{eq:bell_post}
\end{align}

\noindent where the term $R_t(\mathbf{c_t})$ accounts for the expected immediate reward of being in state $S\hspace{-0.1em}_t = (\mathcal{R}\hspace{0.05em}_{t},\mathcal{G}_t)$ and following action $\mathbf{c_t}$, while the remaining terms account for the expected future reward obtained from the successor post-decision state. Specifically, term (I) accounts for the expected future reward in the case where the pre-decision state does not contain any gig workers. Conversely, term (II) represents the expected future reward in the case where the pre-decision state contains a gig worker. In this case, the probabilities of request selection made by the gig worker determine the transition to post-decision states. Consequently, with a probability of \(P^i_{t}\), we transition to the post-decision state \(\mathcal{R}_{t}^{'} \setminus \{i\}\). 

Bellman Equation \ref{eq:bell_post} makes an explicit distinction between the two different sources of stochasticity in the \gls{acr:mdp}. The source of uncertainty which results from the gig workers' stochastic utility is reflected by the transition to all possible post-decision states, while the second source of uncertainty is embedded within the value function of the post-decision states. By separating the transition dynamics in this manner, the complexity of the Bellman equation is reduced, making the problem more tractable by directly depending only on the transition into the post-decision states.

\subsection{Representing gig worker's utility using the \gls{acr:mnl} model and deriving an analytical solution}
\label{sec:mnl}

To effectively utilize the tractability offered by the post-decision state formulation of the Bellman equation, it is essential to model the probability distribution for transitioning to each post-decision state. This necessitates a model that accurately describes gig worker behavior, and the \gls{acr:mnl} model is one of the most commonly used approaches in the literature on discrete choice behavior for both research and practical applications. Its popularity stems from the fact that it offers a closed-form expression for acceptance probabilities and its balance of simplicity and performance, which renders it both versatile and effective. Accordingly, we adopt the \gls{acr:mnl} model as the foundation for our analysis.

\noindent\textit{Multinomial Logit models:} In the \gls{acr:mnl} model, the random variables $\{e_{ij}\}_{i \in \mathcal{R}\hspace{0.05em}_{t}}$ of the utility function (see Equation \ref{eq:util}) are independent and identically distributed (i.i.d), following a Gumbel distribution. We utilize a special case of the MNL model which assumes a linear relationship between the offered compensation and Gumbel-distributed random variables with zero mean. Formally, this variation considers that the utility function of a gig worker $j$ for a request $i$ is:
\begin{equation} U_{ij} = u_{ij} + c_{i} + e_i \end{equation} where $u_{ij}$ is a value indicating the attractiveness of request $i$ to gig worker $j$, and whose value is determined by observable characteristics of the request, $c_i$ is the offered compensation to the gig worker for accepting the request, and $e_i$ are i.i.d. zero mean Gumbel variables with variance equal to $(\mu_j\pi)^2/6$ for some $\mu_j> 0$. Under this \gls{acr:mnl} model, when the request state is $\mathcal{R}\hspace{0.05em}_{t}$, the probability of a gig worker $j$ in time step $t$ accepting request $i$ for offered compensations $\mathbf{c_t} = (c^i_t)_{i \in \mathcal{R}\hspace{0.05em}_{t}}$ equals:
\begin{align} P^i_{t}(\mathbf{c_t}) = \frac{\exp((u_{ij} + c_t^i)/\mu_j)}{\sum_{l \in \mathcal{R}\hspace{0.05em}_{t}}(\exp((u_{lj} + c_t^j)/\mu_j) + \exp(u_0/\mu_j)},\label{eq:mnl1}\end{align} 
while the probability of gig worker $j$ not accepting any requests reads:
\begin{align} P^\emptyset_{t}(\mathbf{c_t}) = \frac{ \exp(u_0/\mu_j)}{\sum_{l \in \mathcal{R}\hspace{0.05em}_{t}}(\exp((u_{lj} + c_t^j)/\mu_j) + \exp(u_0/\mu_j)},\label{eq:mnl2}\end{align} 
where $u_0 \geq 0$ is a constant indicating the attractiveness of the all-reject alternative. 

In the simplest case, all gig workers share the same utility function. Formally, $u_{ij} = u_i$ for all requests $i \in \mathcal{R}\hspace{0.05em}_{t}$ and for all gig workers $j \in \mathcal{G}_t$, and $\mu_j = \mu$ for all gig workers $j \in \mathcal{G}_t$. However, to introduce variability in preferences among the population while maintaining model simplicity, we can assume that the population is divided into $D$ gig worker groups. This formulation corresponds to a Mixed Multinomial Logit (Mixed MNL) model, where within each group, individuals exhibit similar preferences. Formally, for each group $d \in [1,\dots,D]$, the utility function is $u_{ij} = u_i^d$  for all requests $i \in \mathcal{R}\hspace{0.05em}_{t}$ and all gig workers $j \in \mathcal{G}^d_t$, and $\mu_j = \mu_d$ for all gig workers $j \in \mathcal{G}^d_t$.

To solve Bellman equation \ref{eq:bell_post}, we adapt and extend the theory from \cite{dong2009dynamic}, which studies dynamic pricing in the context of inventory control of substitute products, to our problem setting. Specifically, we demonstrate that there exists an alternative formulation of the optimization problem defined by the post-decision state Bellman equation. This alternative formulation uses acceptance probabilities as decision variables and is concave. We then derive the optimal solution by solving the first-order condition.

To this end, we first reformulate the value function by applying Equation \eqref{eq:imm_rew} and representing $P^\emptyset_{t}(c_t^i)$ as $ 1 - \sum_{i \in \mathcal{R}\hspace{0.05em}_{t}}P^i_{t}(\mathbf{c_t})$ and derive the following lemma.
\begin{lemma}\label{lem:dynamic_programming}
The value of the pre-decision state $S\hspace{-0.1em}_t = (\mathcal{R}\hspace{0.05em}_{t},\mathcal{G}_t)$ is equal to:
$$ V_t(\mathcal{R}\hspace{0.05em}_{t},\mathcal{G}_t) = \max_{c_t} \{ \phi_t(\mathcal{R}\hspace{0.05em}_{t},\mathcal{G}_t,\mathbf{c_t}) \} + V^{\mathrm{p}}_{t}(\mathcal{R}\hspace{0.05em}_t^{'})  + \sum_{i \in \mathcal{R}\hspace{0.05em}_t^{\mathrm{exp}}}\! \beta_i $$
where 
$\phi_t(\mathcal{R}\hspace{0.05em}_{t},\mathcal{G}_t,\mathbf{c_t}) = \mathbbm{1}_{|\mathcal{G}_t|=1} \mathbb{E}_{i \sim P_t(\mathbf{c_t})} [r_i - c_t^i - \Delta^i_{V_{t}}(\mathcal{R}\hspace{0.05em}_t^{'}) -\beta_i\mathbbm{1}_{i \in \mathcal{R}\hspace{0.05em}_t^{\mathrm{exp}}}]$ and
$ \Delta^i_{V_{t}}(\mathcal{R}\hspace{0.05em}_t^{'}) = V^{\mathrm{p}}_{t}(\mathcal{R}\hspace{0.05em}_t^{'}) - V^{\mathrm{p}}_{t}(\mathcal{R}\hspace{0.05em}_t^{'} \setminus \{i\}).$

\end{lemma}
The proof of Lemma 1 can be found in Appendix \ref{sec:p_1}. 

As a result of this reformulation, we observe that finding the optimal prices $\mathbf{c_t}$ reduces to optimizing $\phi_t(\mathcal{R}\hspace{0.05em}_{t},\mathcal{G}_t,\mathbf{c_t})$. The function $\phi_t(\mathcal{R}\hspace{0.05em}_{t},\mathcal{G}_t,\mathbf{c_t})$ concerns only scenarios where a gig worker is available, as no decision is required otherwise. It accounts for the probability of each request $i \in \mathcal{R}\hspace{0.05em}_{t}$ being accepted given a compensation $c_t^i$, the corresponding reward and penalty, and an additional term $\Delta^i_{V_{t}}(\mathcal{R}\hspace{0.05em}_t^{'})$. The term $\Delta^i_{V_{t}}(\mathcal{R}\hspace{0.05em}_t^{'})$ reflects the difference in expected reward when request $i \in \mathcal{R}\hspace{0.05em}_{t}$ is accepted immediately versus when it remains unfulfilled, commonly referred to as the opportunity cost of request $i \in \mathcal{R}\hspace{0.05em}_{t}$. As demonstrated by \cite{hanson1996optimizing}, $\phi_t(\mathcal{R}\hspace{0.05em}_{t},\mathcal{G}_t,\mathbf{c_t})$ is not guaranteed to be concave in $\mathbf{c_t}$. 

We can address such optimization problems by expressing the compensation decision $c^t_i$ in terms of the acceptance probabilities. Using the \gls{acr:mnl} model to describe gig worker behavior, we combine Equations \eqref{eq:mnl1} and \eqref{eq:mnl2} to establish a bijection between the compensation \(c^t_i\) and the acceptance probabilities \(P_t^i\). Specifically, the compensation for each request can be expressed as a function of acceptance probabilities as follows:
\begin{align}
\frac{P_t^i(\mathbf{c_t})}{P_t^{\emptyset}(\mathbf{c_t})} = \exp((u_{ij} + c_t^i - u_0)/\mu_j) \Leftrightarrow c_t^i = -u_{ij} + u_0 + \mu_j \ln{P_t^i} - \mu_j \ln{P_t^{\emptyset}}.
\label{eq:mnl3}
\end{align}
Consequently, using the above expression we reformulate $\phi_t$ as a function of $P_t$ and establish the following result.
\begin{lemma}\label{lem:concavity}
$\phi_t(\mathcal{R}\hspace{0.05em}_{t},\mathcal{G}_t,P_t) = \mathbbm{1}_{|\mathcal{G}_t|=1}\mathbb{E}_{i \sim P_t}[r_i + u_{ij} - u_0 - \mu_j \ln{P_t^i} + \mu_j \ln{P_t^{\emptyset}}  - \Delta^i_{V_{t}}(\mathcal{R}\hspace{0.05em}_t^{'}) -\beta_i\mathbbm{1}_{i \in \mathcal{R}\hspace{0.05em}_t^{\mathrm{exp}}}]$ is concave in~$P_t$.
\end{lemma}
The proof of Lemma 2 can be found in Appendix \ref{sec:p_2}. 

By proving the concavity of $\phi_t$ as a function of $P_t$ we can solve the first order condition of $\phi_t$ and derive an optimal solution as a function of the acceptance probabilities $P_t$.
\begin{lemma}\label{lem:exact}
The optimal price $c^{i*}_t$ is given by:
$c^{i*}_t = r_i - \beta_i \mathbbm{1}_{i \in \mathcal{R}\hspace{0.05em}_t^{\mathrm{exp}}} 
 - \Delta^i_{V_{t}}(\mathcal{R}\hspace{0.05em}_t^{'}) - m_t$ where $m_t$ results from solving: 
 $(\frac{m_t}{\mu_j} - 1) \exp\{ \frac{m_t}{\mu_j} - 1\} = \sum_{i \in \mathcal{R}\hspace{0.05em}_{t}} \exp\{ \frac{1}{\mu_j} (r_i + u_{ij} - u_0 - \beta_i \mathbbm{1}_{i \in \mathcal{R}\hspace{0.05em}_t^{\mathrm{exp}}} - \Delta^i_{V_{t}}(\mathcal{R}\hspace{0.05em}_t^{'}) - \mu_j)\}$
\end{lemma}
The proof of Lemma \ref{lem:exact} can be found in Appendix \ref{sec:p_3}. 

Lemma \ref{lem:exact} states that the optimal price for a request~$i$ is equal to the reward of that request, reduced by the following terms: its penalty in the case that it expires within the current time step, the expected value discrepancy $\Delta^i_{V_{t}}(\mathcal{R}\hspace{0.05em}_t^{'})$ and the term $m_t$, which is a state-specific factor that adjusts pricing based on the overall system state rather than on individual requests. In essence, $m_t$ captures system-wide influences on the optimal pricing decision at any given time.

While we define feasible compensations as non-negative in our problem setting, the optimal compensation described in Lemma \ref{lem:exact} may result in negative values under certain conditions, e.g., due to specific worker-request dynamics or model assumptions. In practice, when implementing the optimal compensations, we set any negative values of $\mathbf{c_t} = (c^i_t)_{i \in \mathcal{R}\hspace{0.05em}_{t}}$ to zero to ensure all compensations remain feasible. Since gig workers would not accept requests with either negative or zero compensation, this adjustment aligns with realistic worker behavior and does not compromise the model's practical effectiveness. This adjustment allows for practical feasibility and ensures the model’s applicability in real-world scenarios, where negative compensations would be infeasible, while retaining most of the theoretical solution’s structure and insights.

\subsection{Value function approximation}
\label{sec:statistical_models}
Although our optimization problem has an analytical solution, derived in Section \ref{sec:mnl}, its dependence on the post-decision value function makes precise computation infeasible due to the vast and intractable state space. To overcome this, we employ statistical models to approximate the post-decision value function.

\noindent\underline{Statistical models for the post-decision value function:} A statistical model for the post-decision value function is defined as a function $\hat{V}$ parameterized by $\theta$ which receives as input a post-decision state $\mathcal{R}^{post}\hspace{0.05em}$ and predicts a value $\hat{v}$ for that state: $\hat{V}^\theta: \mathcal{R}^{post}\hspace{0.05em} \mapsto \hat{v} \in \mathbbm{R}$. The primary challenge in designing an effective statistical model to approximate the post-decision state value lies in the fact that the request state is represented as a set, containing a variable number of requests. Consequently, an effective statistical model for the post-decision value function approximation must be able to handle this dynamic structure. To account for such a variability, we utilize a neural network architecture that incorporates an attention mechanism.

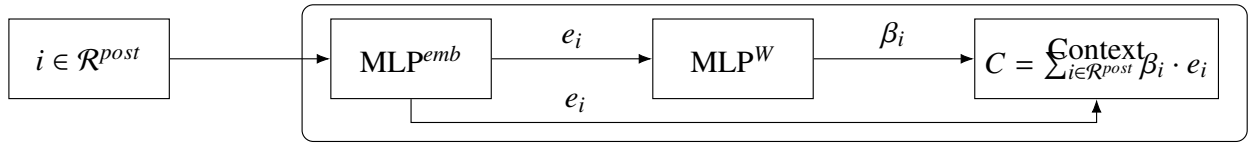
\begin{figure}[b]
\centering
\resizebox{\textwidth}{!}{ 
\begin{tikzpicture}[node distance=2cm]

    \node [block] (input) {$i \in \mathcal{R}^{post}\hspace{0.05em}$};

    \node [block, right=of input] (mlp_emb) {MLP$^{emb}$};
    \node [block, right=of mlp_emb] (mlp_w) {MLP$^W$};
    \node [block, right=of mlp_w] (cell) {Context \\[-0.3cm] $C = \sum_{i \in \mathcal{R}^{post}\hspace{0.05em}} \beta_{i} \cdot e_{i}$};

    \node[fit=(mlp_emb) (mlp_w) (cell), draw, inner sep=10pt, rounded corners,  yshift=-0.18cm] (enclosing_box) {};

    \draw [arrow] (input) -- (mlp_emb);
    \draw [arrow] (mlp_emb) -- node[above] {$e_i$} (mlp_w);
    \draw [arrow] (mlp_w) -- node[above] {$\beta_i$} (cell);
    
    \draw [arrow] (mlp_emb) -- ++(0,-0.8cm) -| node[pos=0.12, above] {$e_i$} (cell);
\end{tikzpicture}
}
\caption{\textnormal{Attention mechanism}}
\label{fig:tikz_diagram}
\end{figure}

\noindent \textit{Attention mechanism:} The attention mechanism (see Figure \ref{fig:tikz_diagram}) calculates an embedding vector $e_i$ for each request $i \in \mathcal{R}^{post}\hspace{0.05em}$ using a multi-layer perceptron (MLP). Subsequently, $e_i$ is processed by a second MLP which performs the following calculation: $\beta_{i} = \sigma(w \cdot \tanh(W \cdot e_{i}))$, where $\sigma$ is the sigmoid function, $w$ is a trainable vector of weights, and $W$ is a trainable matrix of weights. Lastly, it calculates the context vector $C$ as $C = \sum_{i \in \mathcal{R}^{post}\hspace{0.05em}} \beta_{i} \cdot e_{i}$. The context vector is then given as an input to the rest of the neural network architecture. 

\noindent \textit{Neural network architecture:} The neural network for the post-decision value function approximation consists of an attention mechanism that comprises a feedforward layer with 32 units and a Swish activation \citep{ramachandran2017searching}, which provides the embedding vector $e_i$ for each request $i \in \mathcal{R}^{post}\hspace{0.05em}$. It then calculates the context vector using weights $w \in \mathbbm{R}^{64}$ and $W \in \mathbbm{R}^{64 \times 32}$. This context vector, along with other relevant state features (e.g., the number of requests, the most urgent deadline), constitutes the input to the subsequent layers of the network. These layers include two feedforward layers, each with 16 units and a Swish activation. Additional information about neural network training and hyperparameters can be found in Appendix \ref{sec:nn_app}. 

\noindent \underline{Algorithm for training the value function approximator:} To learn the parameters of the statistical model, we follow an approximate value iteration for the post-decision state value function, adapted from \cite{powell2021reinforcement}, to accommodate the continuous action space of our problem setting and the probabilistic transitions to post-decision states. Algorithm~1 shows a pseudocode of our learning procedure: initially, the algorithm receives estimates of the \gls{acr:mnl} parameters $\hat{\mathit{u}}$ and $\hat{\mu}$ (L.1). Then we randomly initialize a parameterization $\theta$ for the statistical model of the post-decision value function (L.2), and repeat the following procedure: first, we gather experiences by interacting with a simulated or real-world environment (L.3). During this time, we select the compensation by considering the estimated optimal price $\hat{c}_{t}$ using the latest estimate of the value function approximation, and we perturb the estimated optimal value using a Gaussian perturbation (L.4) to ensure exploration. After a sufficient amount of experiences is gathered, we update the parameterization $\theta$ (L.5-L.8). For updating the post-decision value function approximation around the observed post-decision state $\mathcal{R}\hspace{0.05em}_{t-1}^{post}$ of time step $t-1$, we use the (estimated) optimal value $v_t^*$ of the observed pre-decision state $S\hspace{-0.1em}_t$ of the following time step $t$.

\begin{algorithm}[b!]
\caption{Approximate value iteration using the post-decision value function}
\begin{algorithmic}[1]  
\REQUIRE MNL model utilities $\hat{\mathit{u}}$, and Gumbel parameter $\hat{\mu}$
\STATE Initialize a value function parameterization $\theta$ for $\hat{V}^\theta$ 
\FOR{episode = 1,\ldots,M}
    \STATE Step 1. Gather experiences using the e-greedy policy:
    \STATE \hspace{\algorithmicindent} $\begin{aligned}[t]
            \mathbf{\hat{c}}_{t} = (\max(0,r_i - \beta_i \mathbbm{1}_{i \in \mathcal{R}\hspace{0.05em}_t^{\mathrm{exp}}} - \Delta_{\hat{V}}^i(\mathcal{R}\hspace{0.05em}_t^{'}) - m_t + \varepsilon_{i}))_{i \in \mathcal{R}\hspace{0.05em}_{t}} 
            \end{aligned}$
            where $\varepsilon_{i} \sim \mathcal{N}(0,\delta)$  $\forall i \in \mathcal{R}\hspace{0.05em}_{t}$
    \STATE Step 2. Update $\theta$ using the target:
    \STATE \hspace{\algorithmicindent} $\begin{aligned}[t]
            \hat{V}^{\theta}(\mathcal{R}\hspace{0.05em}_{t-1}^{post}) \leftarrow v_t^*
            \end{aligned}$
    \STATE where $\mathcal{R}\hspace{0.05em}_{t-1}^{post}$ is the \underline{observed} post-decision state at time step $t-1$ and $v_t^*$ is the estimated optimal value of the successor pre-decision state $S\hspace{-0.1em}_t$:     
    \STATE \hspace{\algorithmicindent} $\begin{aligned}
            v_t^* = \max_{\mathbf{c_t}} \{ R_t(\mathbf{c_t}) +  \gamma \mathbbm{1}_{|\mathcal{G}_t|=1} \sum_{i \in \mathcal{R}\hspace{0.05em}_t\cup\{\emptyset\}}  P_t^i(\mathbf{c_t}) \cdot \hat{V}^{\theta}_t(\mathcal{R}\hspace{0.05em}_t^{'}\backslash\{i\}) \} + (1 - \mathbbm{1}_{|\mathcal{G}_t|=1}) \cdot \hat{V}^{\theta}_t(\mathcal{R}\hspace{0.05em}_t^{'}).
            \end{aligned}$
    \ENDFOR
\end{algorithmic}
\end{algorithm}

\subsection{Training procedure}
\label{sec:training}
\noindent \underline{Statistical model of the gig worker's utility:} In practice, the true \gls{acr:mnl} model parameters for the utilities $u_{ij}$ and the Gumbel parameter $\mu$ are not known. To estimate the parameters of the \gls{acr:mnl} model for each gig worker group, we use observed accept/reject gig worker decisions based on the characteristics of the on-demand request $\mathbf{x}_i$ and offered compensation $c_i$. This data comes either from interactions with the environment under any reasonable policy or from pre-existing historical data. For each gig worker group $d~\in~[1,\dots,D]$ we train a logistic regression model using the log-odds function: $(\mathbf{\hat{w}}_d^T\mathbf{x}_i + c_i)/\beta_d$ where $\mathbf{\hat{w}}_d$ and $\beta_d$ are trainable parameters. Finally, we estimate the utility $u_{ij}$ of a gig worker $j$ for request $i$ using the mapping $\hat{\mathit{u}}_j : \mathbf{x}_i \mapsto \mathbf{\hat{w}}_d^T \mathbf{x}_i$, so that $\hat{u}_{ij} = \hat{\mathit{u}}_j(\mathbf{x}_i)$, and $\beta_d$ by $\hat{\mu}_d = \beta_d$.

\noindent \underline{Training pipeline:} Our training procedure for each scenario involves the following steps: Initially, we gather experiences by interacting with training scenarios. The compensation policy employed sets the compensation for each request $i$  by randomly selecting a value between 40\%-85\% of the request's reward. From these experiences, we train the \gls{acr:mnl} estimator as previously defined, using stochastic gradient descent for optimization. Since the gig worker always chooses the offer that maximizes their utility, the data skews toward higher-compensation offers, especially when the sampling policy is non-optimal and tends to over-offer. Therefore, we use \(L2\) regularization on the weight $\beta_d$ to prevent the weight $\beta_d$ from increasing excessively. We then proceed to train the post-decision value function approximation as outlined in Algorithm 1. In order to mitigate the effect of the network weight initialization, we repeat this process with 5 different random seeds, resulting in 5 distinct models. We select the final model based on the one that demonstrates the best performance on the validation scenarios.

\subsection{Discussion}
\label{sec:meth_as}
Our proposed algorithmic paradigm adopts a model-free approach in many aspects, while it relies on the \gls{acr:mnl} model to capture gig worker behavior. As a result, it requires some knowledge of gig worker groups or at least observable characteristics of gig workers. Below, we discuss key considerations and limitations of our algorithm.

\noindent \textit{Modeling gig worker utility:} Modeling all stochastic aspects of the environment is impractical due to the complexity of real-world scenarios. Instead, selectively modeling key elements provides a practical balance between model-free and model-based approaches. For example, while the variability in gig worker arrivals and requests (e.g., weekday vs. weekend patterns) is too complex to be modeled precisely, focusing on gig worker decision-making is reasonable. In our approach, we use the \gls{acr:mnl} model to represent gig worker decisions. This model is widely used in research and practice, offering flexibility to capture diverse preferences across gig worker groups. However, it assumes a specific mathematical structure (e.g., independence of irrelevant alternatives), which may not fully reflect real-world decision-making. Despite this, its practicality and generalizability justify its use in our algorithm.

\noindent \textit{Knowledge of gig worker groups:} When the gig worker population displays heterogeneous preferences, i.e., when more that one gig worker group exists, the performance of our algorithm relies on having knowledge of the different groups. However, the availability of such information varies across platforms, influencing their ability to identify groups with similar preferences. Platforms seeking to identify potential groups can refer to existing studies, such as \cite{marcucci2017connected}, \cite{bathke2023occasional}, and \cite{miller2017crowdsourced}, which offer valuable insights into gig worker decision-making and subgroup behavior.

\section{Experimental Design}
\label{sec:exp_des}

In the following section, we outline our numerical experiments used to evaluate the performance of our algorithm. To this end, we focus on environments that simulate delivery and ride-hailing platforms, which are two of the fastest growing sectors in the gig economy. Section~\ref{sec:syn_data} introduces our experiments for fully synthetic datasets, which model dynamic environments with stochastic on-demand request and gig worker arrivals, applicable to both delivery and ride-hailing contexts. These synthetic datasets allow us to evaluate the algorithm’s performance across diverse scenarios, including variations in request arrival rates and different compositions of the gig worker population. We consider two scenarios in terms of gig worker population: one with a homogeneous gig worker population and another with a heterogeneous gig worker population composed of three distinct groups of gig workers. Section \ref{sec:nyt_data} introduces our experiments using real-world data from the \cite{NYTdata} dataset, allowing us to assess the algorithm’s performance in a more heterogeneous and realistic setting that simulates a ride-hailing platform. This dataset captures differences between instances; for example, even within instances from the same day and hour, events like national holidays or weather conditions can introduce significant variability. We simulate gig worker behavior, considering both weak and strong location preferences across four regions of Manhattan. Section \ref{sec:benchmarks} presents benchmark policies, including the full information solution policy and formula-based policies, and introduces the performance measure used for comparative analyses. 

\subsection{Synthetic scenarios}
\label{sec:syn_data}

\begin{figure}[b!]%
    \centering
    \fontsize{10}{10}\selectfont
    \subfloat[\centering Utility of group 1]{{\includegraphics[width=0.3\linewidth]{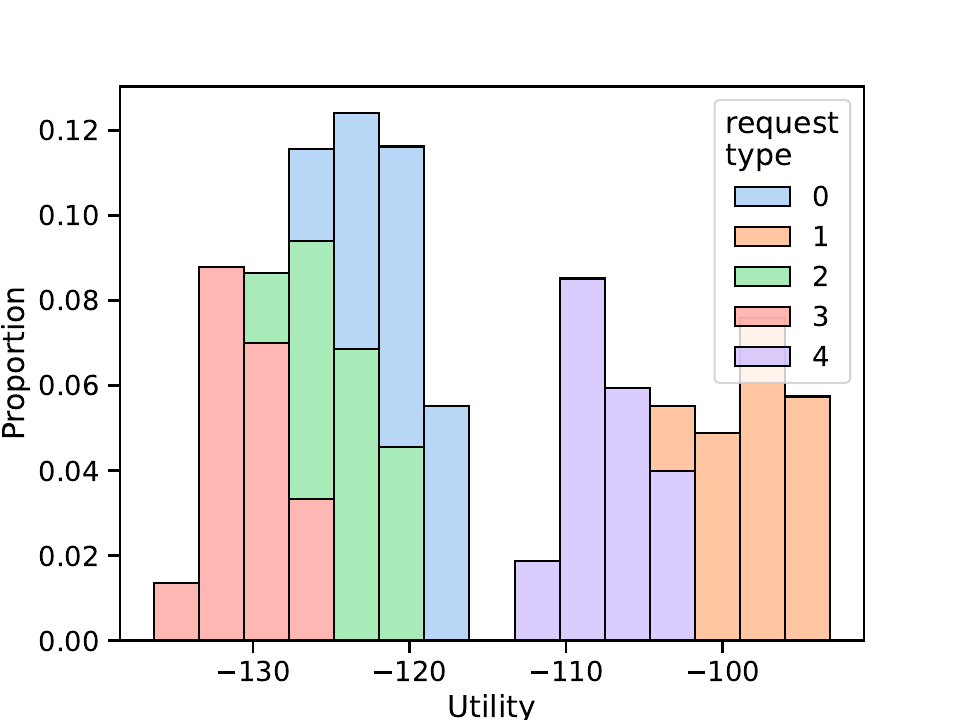} }}%
    \subfloat[\centering Utility of group 2]{{\includegraphics[width=0.3\linewidth]{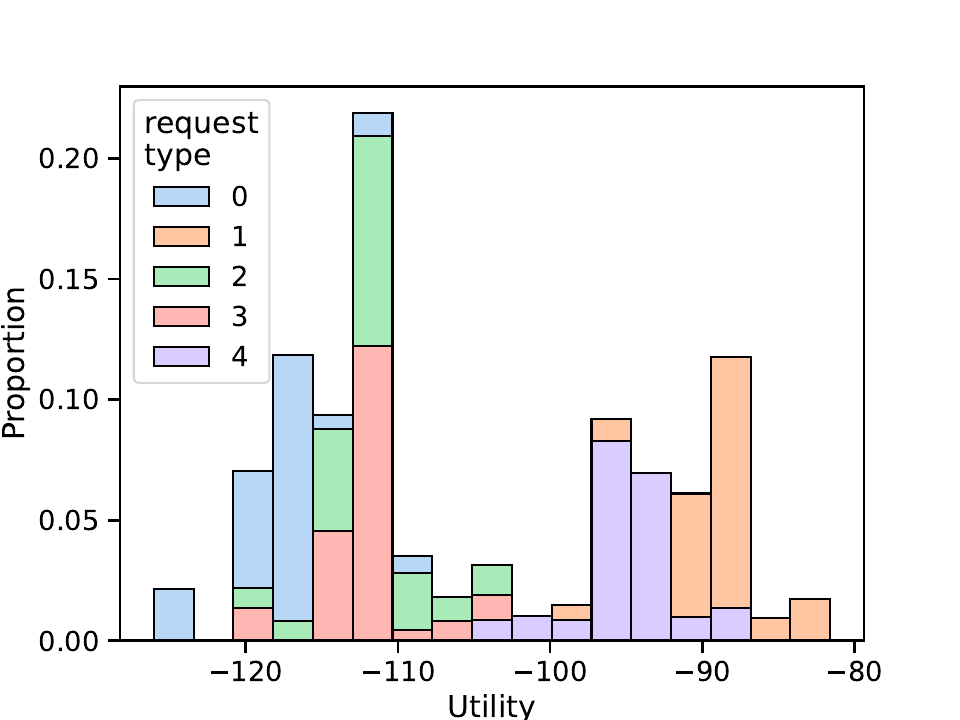} }}%
    \subfloat[\centering Utility of group 3]{{\includegraphics[width=0.3\linewidth]{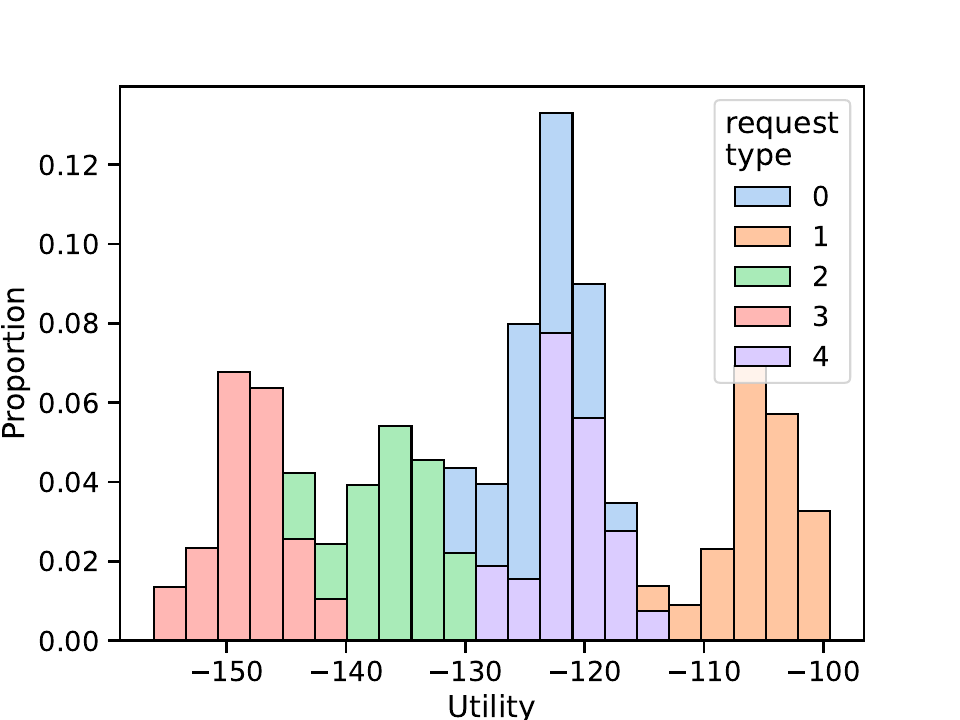} }}%
    \caption{\textnormal{Utilities of each gig worker group in Scenario II separated by the on-demand request type}}%
    \label{fig:utility_scen_2}%
\end{figure}

We simulate a dynamic environment over a horizon of $T = 50$ time steps and model request arrivals using a Poisson Point Process with an arrival rate $\lambda_r$. The maximum duration for which a request stays in the system follows an exponential distribution with parameter $\beta_{\varepsilon}$. There are $k$ types of requests, each defined by $m$ features, a pickup, and a destination location. Requests originate from $n_l^r$ possible locations and have $n_d^r$ possible destinations, both being uniformly distributed. We calculate the rewards for requests as a weighted sum of their characteristics and travel time, with an additional reward for urgency if the request has a deadline of less than three time steps. We calculate request penalties in a  similar way, with an additional penalty for requests that remain in the system for extended periods. The weights for rewards and penalties are uniformly sampled. 

For gig workers, arrivals also follow a Poisson Point Process with an arrival rate $\lambda_g$. We explore two scenarios regarding the types of gig workers. In Scenario I, the gig worker population is homogeneous, represented by a single group ($D = 1$). Within this scenario, we examine three sub-scenarios (I.1–I.3) to assess the impact of varying arrival rates of on-demand requests. Specifically, we test three different request arrival rates, which are equal to, lower or higher than the gig workers' arrival rate correspondingly. In Scenario II, there are three distinct groups of gig workers ($D = 3$), each with a different utility function. We model the utility function for each group as a weighted sum of request characteristics, travel time, preferences for pickup and dropoff locations, and an additional stochastic term. For each gig worker group $d \in [1, \dots, D]$, we denote as $\mathbf{w_d}$ the weight vector associated with this weighted sum, with its values being uniformly sampled. Figure \ref{fig:utility_scen_2} shows the utilities of each gig worker group in Scenario II grouped by each on-demand request type. We assume that the utility of the no-selection alternative is $u_0 = 0$, implying that the deterministic component of the gig worker’s utility is always negative. Consequently, a positive compensation $c > 0$ incentivizes a gig worker's request selection in case of a positive utility. Additionally, we assume that the stochastic term of the utility follows a zero-mean Gumbel distribution parameterized by $\mu=1$ for all gig worker groups. 

\begin{table}[b!]
    \centering
    \captionof{table}{Synthetic Data Generation Parameters}
    \label{tab:scen_param}
    \begin{tabular}{lllllllllllll}
        \toprule
         & $D$ & $\lambda_r$ & $\beta_{\varepsilon}$ & $k$ & $m$ & $n_l^r$ &  $n_d^r$ & $\lambda_g$ & $T$ \\
        \midrule
        \textbf{Scen. I.1} & 1 & 0.5 & 0.1 & 5 & 3 & 5 & 5 & 0.5  & 50 \\
        \midrule
        \textbf{Scen. I.2} & 1 & 0.3 & 0.1 & 5 & 3 & 5 & 5 & 0.5  & 50 \\
        \midrule
        \textbf{Scen. I.3} & 1 & 1 & 0.1 & 5 & 3 & 5 & 5 & 0.5  & 50 \\
        \midrule
        \textbf{Scen. II} & 3 & 0.5 & 0.1 & 5 & 3 & 5 & 5 & 0.5  & 50 \\
        \midrule
        \textbf{NYT weak} & 1 & - & 0.1 & 5 & 2 & 4 & 4 & 0.75 & 120 \\
        \midrule
        \textbf{NYT strong} & 1 & - & 0.1 & 5 & 2 & 4 & 4 & 0.75 & 120 \\
        \bottomrule
    \end{tabular}
\end{table}

For each scenario, we generate 630 distinct realizations of on-demand requests and gig workers. Any policy can be deployed in these scenario realizations, resulting in various episodes of experience. While state transitions may differ in these episodes of experiences, depending on the policy used to generate them, the system's exogenous stochasticity remains constant across them. We divide these 630 realizations into three sets: a training set of 480 realizations, a testing set of 120 realizations, and a validation set of 30 realizations. We use the training realizations to train the \gls{acr:mnl} model and the post-decision value function approximation, while the validation set helps determine the stopping point for the post-decision value function approximation training process, as well as to select the final model. We use the testing set exclusively to evaluate the performance of our algorithm. Table \ref{tab:scen_param} shows the exact parameters used to generate the simulations of each scenario. 

While these datasets exhibit a certain level of complexity, they still lack certain attributes that characterize real-world data. In real-world scenarios, arrivals do not necessarily follow a Poisson Point Process and the distribution among pickup and dropoff locations are not necessarily uniformly distributed. Lastly, significant heterogeneity can exist between different instances. In the following section, we extend our experimental setup using real-world data from the \cite{NYTdata} dataset in order to explore the algorithm’s performance in such heterogeneous settings.

\subsection{New York Taxi (NYT) data}
\label{sec:nyt_data}

\begin{figure}[b!]%
    \centering
    \fontsize{10}{10}\selectfont
    \begin{minipage}[b]{0.3\textwidth}
        \centering
        \subfloat[\centering Manhattan segmentation to regions]{\includegraphics[width=\textwidth]{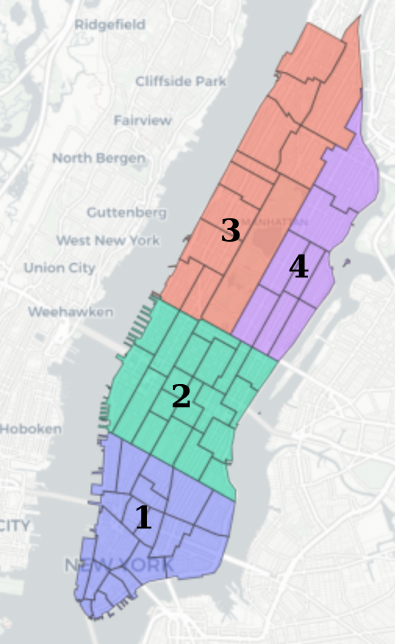}\label{fig:manhattan}}
        \vspace*{-0.8cm}
    \end{minipage}%
    \hspace*{0.2cm}
    \begin{minipage}[b]{0.6\textwidth}
        \centering
        \subfloat[\centering Utilities by pickup location\newline(weak location preferences)]{{\includegraphics[width=0.49\textwidth]{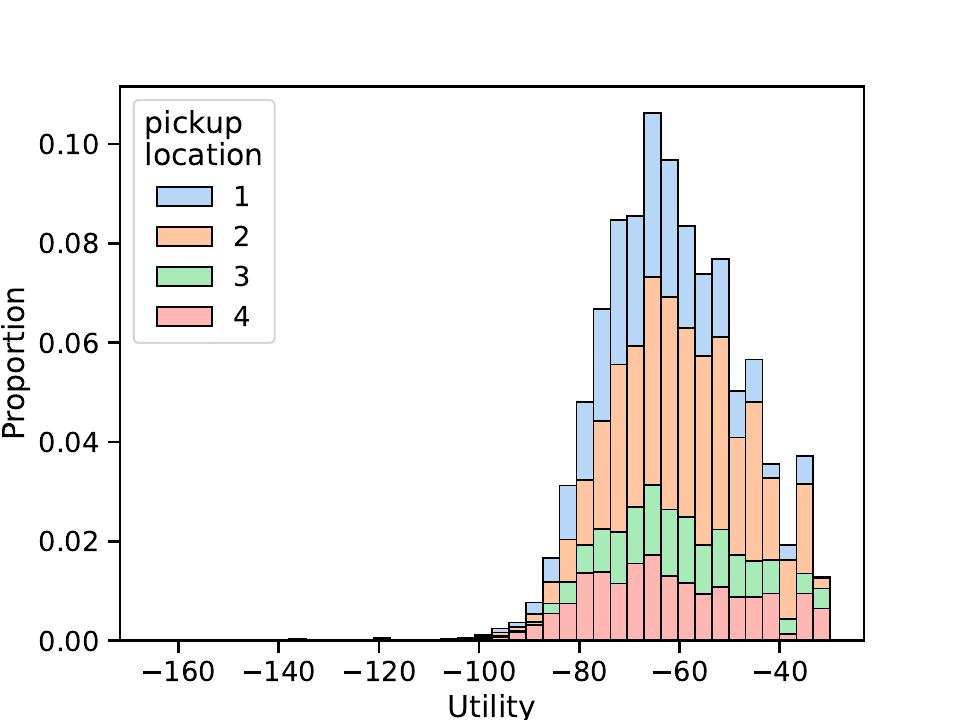}}\label{fig:util_nyt_weak_a}}%
        \hspace*{0.3cm}
        \subfloat[\centering Utilities by dropoff location\newline(weak location preferences)]{{\includegraphics[width=0.49\textwidth]{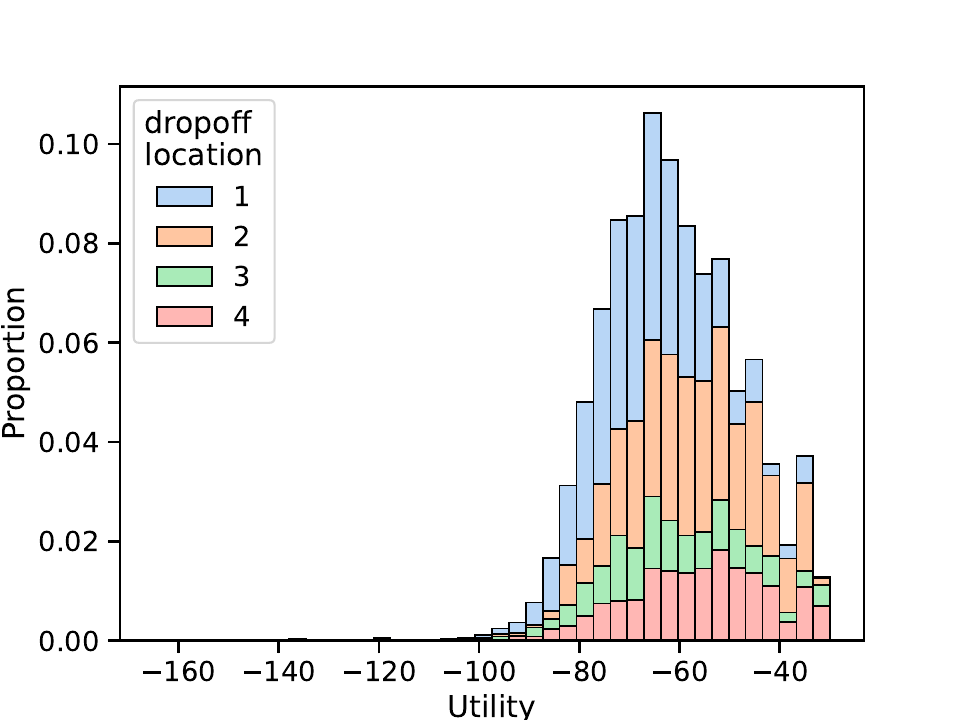}}\label{fig:util_nyt_weak_b}}%
        \vspace*{-0.4cm}
        
        \subfloat[\centering Utilities by pickup location\newline(strong location preferences)]{{\includegraphics[width=0.49\textwidth]{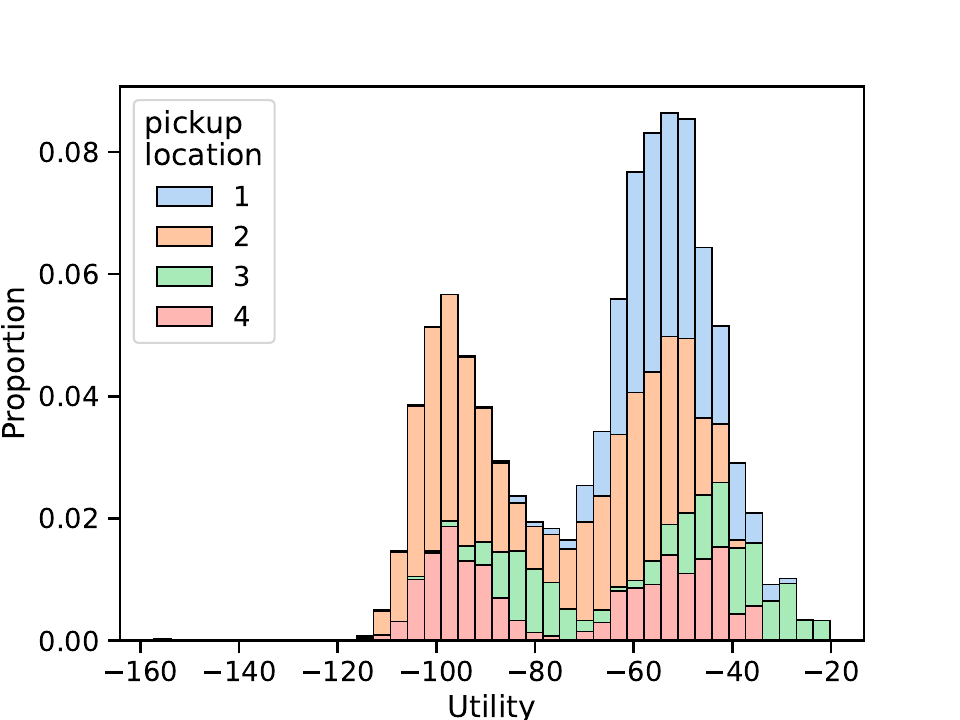}}\label{fig:util_nyt_strong_a}}%
        \hspace*{0.3cm}
        \subfloat[\centering Utilities by dropoff location\newline(strong location preferences)]{{\includegraphics[width=0.49\textwidth]{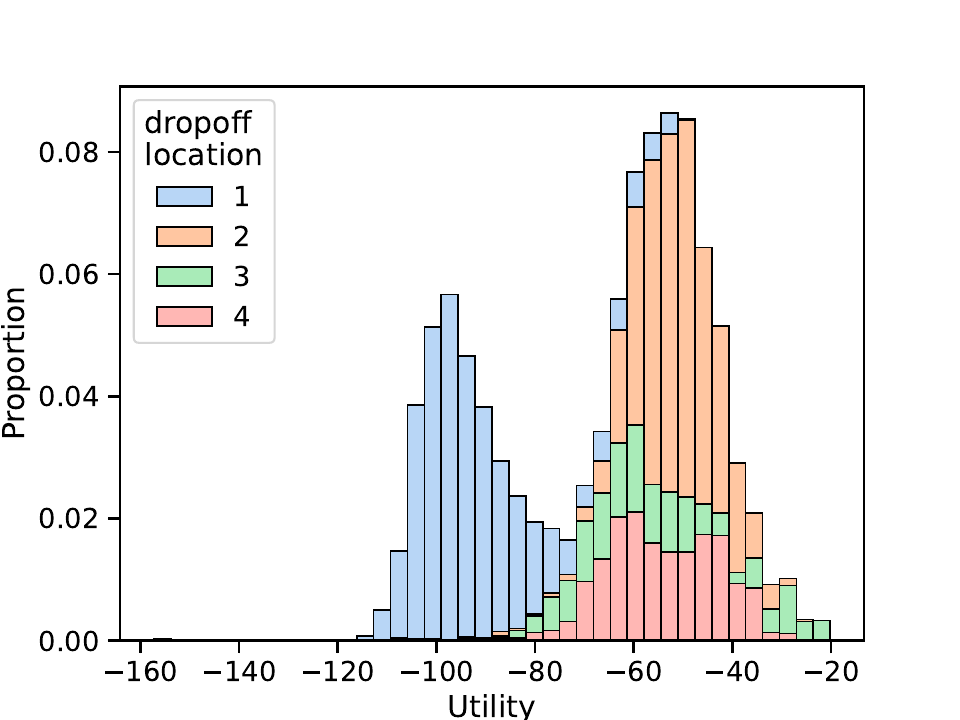}}\label{fig:util_nyt_strong_b}}%
    \end{minipage}
    \caption{\textnormal{Manhattan segmentation (a) and histograms of request utilities grouped by pickup and dropoff locations for gig workers with weak location preferences (b-c) and strong location preferences (d-e).}}
    \label{fig:manhattan_utilities}%
\end{figure}

The \cite{NYTdata} dataset includes detailed trip records for yellow taxis operating in New York City in the year 2018. The dataset includes features such as pick-up and drop-off dates/times, pick-up and drop-off taxi zone locations and trip distances. Due to the substantial volume of trip requests in New York City, we constrained the dataset to requests originating in the Manhattan region on Mondays from 11 a.m. to 12 p.m., and further reduced the dataset size by selecting 30\% of the total on-demand requests. Each episode represents a 30-minute interval, either from 11 a.m. to 11:30 a.m. or from 11:30 a.m. to 12 p.m. Considering the 53 Mondays in 2018, this selection resulted in a total of 106 instances. We select a horizon length $T$ of 120 time steps, corresponding to one decision every 15 seconds, a time frame deemed realistic for gig worker decision-making processes. In this scenario, we assume that the gig worker population exhibits homogeneous preferences. Due to the unavailability of real-world data on gig worker decisions, we simulate gig worker utility synthetically. To this end, we divide Manhattan into four distinct regions, following a similar approach as \cite{yahia2021book}. These regions roughly correspond to: (1) Lower Manhattan, (2) Midtown Manhattan, (3) Upper West Side, and (4) Upper East Side as shown in Figure \ref{fig:manhattan}. We explore two variants of gig worker preferences with respect to region specific pickup and/or dropoff location preferences. In the first variant we generate the location preferences following the same procedure as previously described, i.e., where weights are uniformly selected; in the second variant we explicitly impose a strong preference for specific locations. These will be referred to as weak and strong location preference gig workers, respectively. Figures \ref{fig:manhattan_utilities}b\&c present histograms illustrating the utility distribution of gig workers with weak location preferences across all requests. We display the utility grouped by the requests' pickup locations (Figure \ref{fig:util_nyt_weak_a}) and dropoff locations (Figure \ref{fig:util_nyt_weak_b}). As can be seen, gig workers do not exhibit a strong preference for any particular pickup or dropoff location, resulting in similar utility distributions across locations. Analogously, Figures \ref{fig:manhattan_utilities}d\&e present the utility distributions for the scenario where gig workers have strong location preferences. In this scenario, requests with pickup location 1 are highly preferred, whereas those with dropoff location 1 are less preferred. To increase the complexity of the data, we also introduce two arbitrary characteristics for each request, and assign rewards and penalties similarly to the fully synthetic dataset. The resulting 106 realizations are divided into three subsets: a training set comprising 76 realizations, a testing set of 20 realizations, and a validation set containing the remaining 10 realizations.

\subsection{Benchmark policies \& performance measure}
\label{sec:benchmarks}
To assess the performance of our algorithm, we compare it against several benchmark policies. These benchmarks include policies that calculate compensation based on factors like reward, travel distance, and urgency, reflecting common practices in crowdsourced delivery platforms \citep{alnaggar2021crowdsourced}, as well as theoretical upper bounds to help identify the strengths and limitations of our approach. Additionally, to understand the effect of the \gls{acr:mnl} model on performance, we create benchmarks using our algorithm with both the true \gls{acr:mnl} parameters and perturbed versions of them. We now provide details on the benchmark policies used to evaluate our algorithm’s performance. 
~\\
\textbf{\gls{acr:pp}:} The reward percentage compensation policy sets a compensation for a request $i$ based on a percentage of its total reward. We perform hyperparameter tuning via gridsearch to select the best percentage ranging from $40-100\%$. 
~\\
\textbf{\gls{acr:fp}:} The formula-based compensation policy sets a compensation for a request $i$ based on the following formula:
\begin{align}
\text{FP}(i) = v_1 \cdot r_i + v_2 \cdot td_i + v_3 \cdot \beta_i + (v_4 \cdot r_i) \cdot \mathbbm{1}_{i \in \mathcal{R}\hspace{0.05em}_t^{\mathrm{exp}}}
\end{align}
where $r_i$ is the reward, $td_i$ is the travel distance and $\beta_i$ is the penalty. The value $v_4$ can be considered as an extra boost on the compensation in cases where the request is expiring in the current time step. 
We conduct hyperparameter optimization via grid search over a selection of weights for each attribute and refer to Table \ref{table:fbp_val} in Appendix \ref{sec:fbp_app} for details.
~\\
\textbf{Full information:} The full information policy is an offline policy that assumes complete information about all requests and gig workers, including all stochastic elements, i.e., request and gig worker arrival times and the realization of the random element of the gig worker's utility. This policy solves the full-information problem optimally and yields an upper bound. We refer to Appendix \ref{sec:fi_app} for details on the upper bound computation. 
~\\ 
\textbf{\gls{acr:p-pd-vfa}:} This policy uses perturbed versions of the true weight vectors \(\mathbf{w_d}^{pert} = \mathbf{w_d} + \delta_{dw} \) for \( d \in [1,\dots,D] \) and the Gumbel parameter \( \mu_d^{pert} = \mu_d + \delta_{d\mu} \) of the gig worker utility. We generate the combined perturbation vector \( \delta_d = (\delta_{dw}, \delta_{d\mu}) \) to ensure that both the perturbation and the resulting perturbed vector are constrained within a maximum Euclidean distance $\epsilon$ and a minimum distance of $\max(0,\epsilon - 1)$ from the true vector, formally \(\max(0,\epsilon - 1) \leq \|\delta_d\|_2 \leq \epsilon \) for \( d \in [1,\dots,D] \). The special case where \( \epsilon = 0 \) indicates perfect knowledge of the deterministic term of the gig worker utility, as the true weight vectors \( \mathbf{w_d} \) and Gumbel parameter $\mu_d$ for \( d \in [1,\dots,D] \) are exactly known. Therefore, this version has a competitive advantage over the standard algorithm by leveraging exact environmental values which are not available to the standard version.

In order to evaluate the performance of each algorithm we calculate a performance ratio as follows.
~\\
\textbf{Performance ratio:}
Let $I$ be the set of all considered instances. The performance ratio of the solution of an algorithm for an instance $i \in I$ is equal to:
$$
\left(1 - \frac{\rho^{opt}_i - \rho^{alg}_i}{\rho^{opt}_i - \sum_{j=1}^{N_i} \beta_{ij}} \right) \cdot 100
$$
\noindent where $\rho^{\text{opt}}_i $,  is the total reward acquired on instance $i \in I$ in the full information solution, $\rho^{alg}_i$ is the total reward acquired on instance $i \in I$ by the considered algorithm, $\beta_{ij}$ is the penalty of the $j$-th request in instance $i \in I$ and $N_i$ is the total number of requests in instance $i \in I$. Essentially, the performance ratio evaluates the algorithm’s performance as a percentage of how close the algorithm’s solution is to the optimal solution, while also considering a baseline where no requests are accepted, and only penalties are incurred. Higher values indicate better performance, while lower values reflect poorer performance.

\section{Results}
\label{sec:res}

We organize our results discussion in three parts: In Section \ref{sec:res_syn}, we compare the performance of our algorithm to that of the benchmark policies on the synthetic dataset with a homogeneous gig worker population. In Section \ref{sec:res_syn_2}, we examine the performance of our algorithm on the synthetic dataset where the gig worker population is heterogeneous, analyzing the impact of correctly modeling multiple gig worker groups versus incorrectly assuming a homogeneous population. Lastly, Section \ref{sec:res_nyt} focuses on the performance of our algorithm using the NYT datasets, for gig workers with weak and strong location preferences. 

\subsection{Homogeneous gig worker population}
\label{sec:res_syn}
We first focus on a simplified case where the gig worker population is homogeneous, represented by a single group (D = 1). We examine three sub-scenarios (I.1–I.3), each with different arrival rates of on-demand requests. The benchmarks include the \gls{acr:pp}, \gls{acr:fp}, and Pert. ($\epsilon=0$), alongside our algorithm, employing a single \gls{acr:mnl} model in alignment with the environmental conditions. 

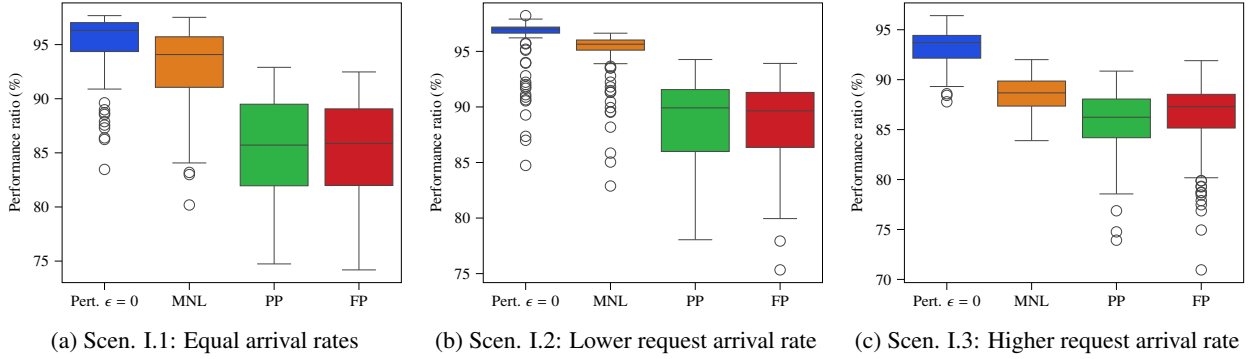
\begin{figure}[t]%
    \centering
    \fontsize{9}{9}\selectfont
    \subfloat[\footnotesize \centering Scen. I.1: Equal arrival rates]{{\adjustbox{width=0.32\textwidth}{
\begin{tikzpicture}

\definecolor{chocolate22312431}{RGB}{223,124,31}
\definecolor{darkgray176}{RGB}{176,176,176}
\definecolor{darkslategray68}{RGB}{68,68,68}
\definecolor{firebrick2032937}{RGB}{203,29,37}
\definecolor{limegreen4717970}{RGB}{47,179,70}
\definecolor{royalblue3378223}{RGB}{33,78,223}

\begin{axis}[
tick align=outside,
tick pos=left,
x grid style={darkgray176},
xmin=-0.5, xmax=3.5,
xtick style={color=black},
xtick={0,1,2,3},
xticklabels={Pert. \(\displaystyle \epsilon=0\),MNL,PP,FP},
y grid style={darkgray176},
ylabel={Performance ratio (\%)},
ymin=73.0126816492352, ymax=98.851891984976,
ytick style={color=black},
ytick={70,75,80,85,90,95,100},
yticklabels={
  \(\displaystyle {70}\),
  \(\displaystyle {75}\),
  \(\displaystyle {80}\),
  \(\displaystyle {85}\),
  \(\displaystyle {90}\),
  \(\displaystyle {95}\),
  \(\displaystyle {100}\)
}
]
\path [draw=darkslategray68, fill=royalblue3378223]
(axis cs:-0.4,94.3583668716296)
--(axis cs:0.4,94.3583668716296)
--(axis cs:0.4,97.0304994279352)
--(axis cs:-0.4,97.0304994279352)
--(axis cs:-0.4,94.3583668716296)
--cycle;
\addplot [darkslategray68]
table {%
0 94.3583668716296
0 90.8923771825346
};
\addplot [darkslategray68]
table {%
0 97.0304994279352
0 97.6773824242606
};
\addplot [darkslategray68]
table {%
-0.2 90.8923771825346
0.2 90.8923771825346
};
\addplot [darkslategray68]
table {%
-0.2 97.6773824242606
0.2 97.6773824242606
};
\addplot [black, mark=o, mark size=3, mark options={solid,fill opacity=0,draw=darkslategray68}, only marks]
table {%
0 87.8817328799838
0 87.5646932209679
0 89.6215083940763
0 87.3032659192475
0 88.5449661317598
0 86.2373659689477
0 86.3858890120452
0 83.4682240609051
0 88.7055814669542
0 88.9774803758999
};
\path [draw=darkslategray68, fill=chocolate22312431]
(axis cs:0.6,91.0542915481331)
--(axis cs:1.4,91.0542915481331)
--(axis cs:1.4,95.7218502384598)
--(axis cs:0.6,95.7218502384598)
--(axis cs:0.6,91.0542915481331)
--cycle;
\addplot [darkslategray68]
table {%
1 91.0542915481331
1 84.0683351615054
};
\addplot [darkslategray68]
table {%
1 95.7218502384598
1 97.5190151596301
};
\addplot [darkslategray68]
table {%
0.8 84.0683351615054
1.2 84.0683351615054
};
\addplot [darkslategray68]
table {%
0.8 97.5190151596301
1.2 97.5190151596301
};
\addplot [black, mark=o, mark size=3, mark options={solid,fill opacity=0,draw=darkslategray68}, only marks]
table {%
1 80.1832295616118
1 82.9924018720299
1 83.1940641618163
};
\path [draw=darkslategray68, fill=limegreen4717970]
(axis cs:1.6,81.9577093234914)
--(axis cs:2.4,81.9577093234914)
--(axis cs:2.4,89.493561829944)
--(axis cs:1.6,89.493561829944)
--(axis cs:1.6,81.9577093234914)
--cycle;
\addplot [darkslategray68]
table {%
2 81.9577093234914
2 74.7397800293973
};
\addplot [darkslategray68]
table {%
2 89.493561829944
2 92.900743210251
};
\addplot [darkslategray68]
table {%
1.8 74.7397800293973
2.2 74.7397800293973
};
\addplot [darkslategray68]
table {%
1.8 92.900743210251
2.2 92.900743210251
};
\path [draw=darkslategray68, fill=firebrick2032937]
(axis cs:2.6,81.9848971547146)
--(axis cs:3.4,81.9848971547146)
--(axis cs:3.4,89.0634049663873)
--(axis cs:2.6,89.0634049663873)
--(axis cs:2.6,81.9848971547146)
--cycle;
\addplot [darkslategray68]
table {%
3 81.9848971547146
3 74.1871912099507
};
\addplot [darkslategray68]
table {%
3 89.0634049663873
3 92.4757144002047
};
\addplot [darkslategray68]
table {%
2.8 74.1871912099507
3.2 74.1871912099507
};
\addplot [darkslategray68]
table {%
2.8 92.4757144002047
3.2 92.4757144002047
};
\addplot [darkslategray68]
table {%
-0.4 96.3142210188277
0.4 96.3142210188277
};
\addplot [darkslategray68]
table {%
0.6 94.0807111742749
1.4 94.0807111742749
};
\addplot [darkslategray68]
table {%
1.6 85.7174748323543
2.4 85.7174748323543
};
\addplot [darkslategray68]
table {%
2.6 85.8620096566213
3.4 85.8620096566213
};
\draw (axis cs:0,66.5528790652999) node[
  scale=0.75,
  anchor=base,
  text=black,
  rotate=0.0
]{\bfseries 95.17};
\draw (axis cs:1,66.5528790652999) node[
  scale=0.75,
  anchor=base,
  text=black,
  rotate=0.0
]{\bfseries 92.93};
\draw (axis cs:2,66.5528790652999) node[
  scale=0.75,
  anchor=base,
  text=black,
  rotate=0.0
]{\bfseries 85.56};
\draw (axis cs:3,66.5528790652999) node[
  scale=0.75,
  anchor=base,
  text=black,
  rotate=0.0
]{\bfseries 85.18};
\draw (axis cs:-1,66.8112711686574) node[
  scale=0.75,
  text=black,
  rotate=0.0
]{\bfseries \textbf{Average:}};
\end{axis}

\end{tikzpicture}} }}%
    \hspace*{0.1cm} 
    \subfloat[\footnotesize \centering Scen. I.2: Lower request arrival rate]{{\adjustbox{width=0.32\textwidth}{
\begin{tikzpicture}

\definecolor{chocolate22312431}{RGB}{223,124,31}
\definecolor{darkgray176}{RGB}{176,176,176}
\definecolor{darkslategray68}{RGB}{68,68,68}
\definecolor{firebrick2032937}{RGB}{203,29,37}
\definecolor{limegreen4717970}{RGB}{47,179,70}
\definecolor{royalblue3378223}{RGB}{33,78,223}

\begin{axis}[
tick align=outside,
tick pos=left,
x grid style={darkgray176},
xmin=-0.5, xmax=3.5,
xtick style={color=black},
xtick={0,1,2,3},
xticklabels={Pert. \(\displaystyle \epsilon=0\),MNL,PP,FP},
y grid style={darkgray176},
ylabel={Performance ratio (\%)},
ymin=74.1910997484927, ymax=99.361593772276,
ytick style={color=black},
ytick={70,75,80,85,90,95,100},
yticklabels={
  \(\displaystyle {70}\),
  \(\displaystyle {75}\),
  \(\displaystyle {80}\),
  \(\displaystyle {85}\),
  \(\displaystyle {90}\),
  \(\displaystyle {95}\),
  \(\displaystyle {100}\)
}
]
\path [draw=darkslategray68, fill=royalblue3378223]
(axis cs:-0.4,96.6419615346797)
--(axis cs:0.4,96.6419615346797)
--(axis cs:0.4,97.1868975892247)
--(axis cs:-0.4,97.1868975892247)
--(axis cs:-0.4,96.6419615346797)
--cycle;
\addplot [darkslategray68]
table {%
0 96.6419615346797
0 96.2165776607347
};
\addplot [darkslategray68]
table {%
0 97.1868975892247
0 97.9052317733693
};
\addplot [darkslategray68]
table {%
-0.2 96.2165776607347
0.2 96.2165776607347
};
\addplot [darkslategray68]
table {%
-0.2 97.9052317733693
0.2 97.9052317733693
};
\addplot [black, mark=o, mark size=3, mark options={solid,fill opacity=0,draw=darkslategray68}, only marks]
table {%
0 91.8279480049551
0 84.7358075575047
0 95.731481858829
0 87.3523052390947
0 90.8844376264629
0 91.6782689356234
0 92.8037957744955
0 90.5444487946375
0 89.2767045807754
0 95.6738294834345
0 86.9985047098216
0 90.6443018209354
0 92.2229984354157
0 92.0129322350107
0 95.1941046744618
0 93.9416729650652
0 91.1006434706477
0 95.1077941488231
0 93.9942138781203
0 98.2174804075586
};
\path [draw=darkslategray68, fill=chocolate22312431]
(axis cs:0.6,95.110363317606)
--(axis cs:1.4,95.110363317606)
--(axis cs:1.4,96.0240261224948)
--(axis cs:0.6,96.0240261224948)
--(axis cs:0.6,95.110363317606)
--cycle;
\addplot [darkslategray68]
table {%
1 95.110363317606
1 93.8924768411228
};
\addplot [darkslategray68]
table {%
1 96.0240261224948
1 96.6358324586154
};
\addplot [darkslategray68]
table {%
0.8 93.8924768411228
1.2 93.8924768411228
};
\addplot [darkslategray68]
table {%
0.8 96.6358324586154
1.2 96.6358324586154
};
\addplot [black, mark=o, mark size=3, mark options={solid,fill opacity=0,draw=darkslategray68}, only marks]
table {%
1 82.8842530770368
1 89.5204426414265
1 90.4651890543327
1 91.4780454323682
1 89.5787084160496
1 88.180842255625
1 85.8443944298772
1 91.8057237139984
1 91.3251492021376
1 93.5077074695329
1 93.6575289638538
1 92.2301017442514
1 89.9438044704687
1 93.5124748810001
1 85.0386548103656
1 91.3128459352645
1 92.7947374136883
};
\path [draw=darkslategray68, fill=limegreen4717970]
(axis cs:1.6,85.9939607721336)
--(axis cs:2.4,85.9939607721336)
--(axis cs:2.4,91.5599409422826)
--(axis cs:1.6,91.5599409422826)
--(axis cs:1.6,85.9939607721336)
--cycle;
\addplot [darkslategray68]
table {%
2 85.9939607721336
2 78.0485963504318
};
\addplot [darkslategray68]
table {%
2 91.5599409422826
2 94.26559936098
};
\addplot [darkslategray68]
table {%
1.8 78.0485963504318
2.2 78.0485963504318
};
\addplot [darkslategray68]
table {%
1.8 94.26559936098
2.2 94.26559936098
};
\path [draw=darkslategray68, fill=firebrick2032937]
(axis cs:2.6,86.358498706691)
--(axis cs:3.4,86.358498706691)
--(axis cs:3.4,91.3063709310569)
--(axis cs:2.6,91.3063709310569)
--(axis cs:2.6,86.358498706691)
--cycle;
\addplot [darkslategray68]
table {%
3 86.358498706691
3 79.9520251212715
};
\addplot [darkslategray68]
table {%
3 91.3063709310569
3 93.9166913358089
};
\addplot [darkslategray68]
table {%
2.8 79.9520251212715
3.2 79.9520251212715
};
\addplot [darkslategray68]
table {%
2.8 93.9166913358089
3.2 93.9166913358089
};
\addplot [black, mark=o, mark size=3, mark options={solid,fill opacity=0,draw=darkslategray68}, only marks]
table {%
3 75.3352131132101
3 77.9293968694246
};
\addplot [darkslategray68]
table {%
-0.4 96.9900443614698
0.4 96.9900443614698
};
\addplot [darkslategray68]
table {%
0.6 95.6474432911195
1.4 95.6474432911195
};
\addplot [darkslategray68]
table {%
1.6 89.9159767633996
2.4 89.9159767633996
};
\addplot [darkslategray68]
table {%
2.6 89.6544379872342
3.4 89.6544379872342
};
\draw (axis cs:0,67.8984762425468) node[
  scale=0.75,
  anchor=base,
  text=black,
  rotate=0.0
]{\bfseries 96.19};
\draw (axis cs:1,67.8984762425468) node[
  scale=0.75,
  anchor=base,
  text=black,
  rotate=0.0
]{\bfseries 94.90};
\draw (axis cs:2,67.8984762425468) node[
  scale=0.75,
  anchor=base,
  text=black,
  rotate=0.0
]{\bfseries 88.64};
\draw (axis cs:3,67.8984762425468) node[
  scale=0.75,
  anchor=base,
  text=black,
  rotate=0.0
]{\bfseries 88.58};
\draw (axis cs:-1,68.1501811827847) node[
  scale=0.75,
  text=black,
  rotate=0.0
]{\bfseries \textbf{Average:}};
\end{axis}

\end{tikzpicture}} }}%
    \hspace*{0.1cm}
    \subfloat[\footnotesize \centering Scen. I.3: Higher request arrival rate]{{\adjustbox{width=0.32\textwidth}{
\begin{tikzpicture}

\definecolor{chocolate22312431}{RGB}{223,124,31}
\definecolor{darkgray176}{RGB}{176,176,176}
\definecolor{darkslategray68}{RGB}{68,68,68}
\definecolor{firebrick2032937}{RGB}{203,29,37}
\definecolor{limegreen4717970}{RGB}{47,179,70}
\definecolor{royalblue3378223}{RGB}{33,78,223}

\begin{axis}[
tick align=outside,
tick pos=left,
x grid style={darkgray176},
xmin=-0.5, xmax=3.5,
xtick style={color=black},
xtick={0,1,2,3},
xticklabels={Pert. \(\displaystyle \epsilon=0\),MNL,PP,FP},
y grid style={darkgray176},
ylabel={Performance ratio (\%)},
ymin=69.680673639012, ymax=97.6849116827857,
ytick style={color=black},
ytick={65,70,75,80,85,90,95,100},
yticklabels={
  \(\displaystyle {65}\),
  \(\displaystyle {70}\),
  \(\displaystyle {75}\),
  \(\displaystyle {80}\),
  \(\displaystyle {85}\),
  \(\displaystyle {90}\),
  \(\displaystyle {95}\),
  \(\displaystyle {100}\)
}
]
\path [draw=darkslategray68, fill=royalblue3378223]
(axis cs:-0.4,92.1550598873663)
--(axis cs:0.4,92.1550598873663)
--(axis cs:0.4,94.4320875086815)
--(axis cs:-0.4,94.4320875086815)
--(axis cs:-0.4,92.1550598873663)
--cycle;
\addplot [darkslategray68]
table {%
0 92.1550598873663
0 89.3141091510825
};
\addplot [darkslategray68]
table {%
0 94.4320875086815
0 96.411991771705
};
\addplot [darkslategray68]
table {%
-0.2 89.3141091510825
0.2 89.3141091510825
};
\addplot [darkslategray68]
table {%
-0.2 96.411991771705
0.2 96.411991771705
};
\addplot [black, mark=o, mark size=3, mark options={solid,fill opacity=0,draw=darkslategray68}, only marks]
table {%
0 88.4261860970828
0 88.5921491576229
0 87.7830848219175
};
\path [draw=darkslategray68, fill=chocolate22312431]
(axis cs:0.6,87.3560116562822)
--(axis cs:1.4,87.3560116562822)
--(axis cs:1.4,89.860462167285)
--(axis cs:0.6,89.860462167285)
--(axis cs:0.6,87.3560116562822)
--cycle;
\addplot [darkslategray68]
table {%
1 87.3560116562822
1 83.8944048329985
};
\addplot [darkslategray68]
table {%
1 89.860462167285
1 92.0054887151522
};
\addplot [darkslategray68]
table {%
0.8 83.8944048329985
1.2 83.8944048329985
};
\addplot [darkslategray68]
table {%
0.8 92.0054887151522
1.2 92.0054887151522
};
\path [draw=darkslategray68, fill=limegreen4717970]
(axis cs:1.6,84.1874064264844)
--(axis cs:2.4,84.1874064264844)
--(axis cs:2.4,88.0602755325824)
--(axis cs:1.6,88.0602755325824)
--(axis cs:1.6,84.1874064264844)
--cycle;
\addplot [darkslategray68]
table {%
2 84.1874064264844
2 78.5603102520858
};
\addplot [darkslategray68]
table {%
2 88.0602755325824
2 90.8559884182014
};
\addplot [darkslategray68]
table {%
1.8 78.5603102520858
2.2 78.5603102520858
};
\addplot [darkslategray68]
table {%
1.8 90.8559884182014
2.2 90.8559884182014
};
\addplot [black, mark=o, mark size=3, mark options={solid,fill opacity=0,draw=darkslategray68}, only marks]
table {%
2 74.7604296747939
2 76.8727589975359
2 73.9267979744025
};
\path [draw=darkslategray68, fill=firebrick2032937]
(axis cs:2.6,85.1576943161778)
--(axis cs:3.4,85.1576943161778)
--(axis cs:3.4,88.5320969158398)
--(axis cs:2.6,88.5320969158398)
--(axis cs:2.6,85.1576943161778)
--cycle;
\addplot [darkslategray68]
table {%
3 85.1576943161778
3 80.1794945097269
};
\addplot [darkslategray68]
table {%
3 88.5320969158398
3 91.9023604993431
};
\addplot [darkslategray68]
table {%
2.8 80.1794945097269
3.2 80.1794945097269
};
\addplot [darkslategray68]
table {%
2.8 91.9023604993431
3.2 91.9023604993431
};
\addplot [black, mark=o, mark size=3, mark options={solid,fill opacity=0,draw=darkslategray68}, only marks]
table {%
3 77.4704878638738
3 78.6815953415637
3 79.8698078235877
3 77.889348193281
3 79.3237171490251
3 79.8853904517332
3 78.7327256715275
3 79.2652786401799
3 70.9535935500926
3 76.8683844974797
3 78.4509166524302
3 74.9461209549858
};
\addplot [darkslategray68]
table {%
-0.4 93.7052431465881
0.4 93.7052431465881
};
\addplot [darkslategray68]
table {%
0.6 88.678655660778
1.4 88.678655660778
};
\addplot [darkslategray68]
table {%
1.6 86.2381039465492
2.4 86.2381039465492
};
\addplot [darkslategray68]
table {%
2.6 87.2927229959253
3.4 87.2927229959253
};
\draw (axis cs:0,62.6796141280686) node[
  scale=0.75,
  anchor=base,
  text=black,
  rotate=0.0
]{\bfseries 93.25};
\draw (axis cs:1,62.6796141280686) node[
  scale=0.75,
  anchor=base,
  text=black,
  rotate=0.0
]{\bfseries 88.51};
\draw (axis cs:2,62.6796141280686) node[
  scale=0.75,
  anchor=base,
  text=black,
  rotate=0.0
]{\bfseries 85.78};
\draw (axis cs:3,62.6796141280686) node[
  scale=0.75,
  anchor=base,
  text=black,
  rotate=0.0
]{\bfseries 86.18};
\draw (axis cs:-1,62.9596565085063) node[
  scale=0.75,
  text=black,
  rotate=0.0
]{\bfseries \textbf{Average:}};
\end{axis}

\end{tikzpicture}} }}%
    \caption{\textnormal{Performance ratios for synthetic scenarios I.1, I.2, and I.3, which correspond to equal, lower, and higher request arrival rates compared to gig worker arrival rates.}}%
    \label{fig:perf_syn}%
\end{figure}

\noindent \textbf{Performance:} The boxplots in Figure \ref{fig:perf_syn} show the performance ratios across all test set realizations for each of the three sub-scenarios. In Scenario I.1, where the arrival rates of requests and gig workers are equal, our algorithm achieves an average performance ratio of 92.9\% (±3.4\%), compared to 95.1\% (±2.8\%) for the Pert. ($\epsilon=0$) benchmark, which has an inherent advantage over all other benchmarks due to prior and accurate knowledge of gig workers' utilities. The rule based benchmarks achieve a performance of 85.5\% (±4.4\%) for \gls{acr:pp}, and 85.1\% (±4.6\%) for \gls{acr:fp}. Accordingly, our algorithm yields approximately an average improvement of 7.5\% over both benchmarks, while at the same time showing a lower standard deviation. This indicates that our algorithm is particularly effective when the arrival rates are balanced highlighting the stability of our method in this scenario.

In Scenario I.2, where the arrival rate of requests is lower than that of gig workers, our algorithm achieves an average performance ratio of 94.7\% (±2.3\%), compared to 96.1\% (±2.3\%) for the Pert. ($\epsilon=0$), 88.6\% (±3.8\%) for the \gls{acr:pp}, and 88.5\% (±3.7\%) for the \gls{acr:fp} benchmark. Here, the average performance improvement of 6.2\% over \gls{acr:pp} and \gls{acr:fp} highlights our algorithm's capability to maintain a high performance ratio in situations where requests are less frequent. Additionally, our algorithm's standard deviation remains lower than that of both \gls{acr:pp} and \gls{acr:fp}, indicating consistent performance in this scenario as well.

Lastly, in Scenario I.3, where the arrival rate of requests exceeds that of gig workers, our algorithm achieves an average performance ratio of 88.5\% (±1.8\%), compared to 93.2\% (±1.7\%) for the Pert. ($\epsilon=0$), 85.7\% (±3.1\%) for the \gls{acr:pp}, and 86.1\% (±3.7\%) for the \gls{acr:fp} benchmark. In this case, the algorithm only outperforms the \gls{acr:pp} and \gls{acr:fp} benchmarks by an average of 2.5\%. This scenario reveals a lower performance ratio, indicating that while our algorithm remains beneficial, it is less effective in scenarios where the arrival of requests significantly exceeds that of gig workers. We attribute this reduced effectiveness to potential estimation errors in our algorithm, which relies on the \gls{acr:mnl} model for utility predictions. With a high volume of requests, even minor utility estimation errors may amplify across numerous requests, causing greater deviations in performance. This challenge does not affect the Pert. ($\epsilon=0$) benchmark since this benchmark leverages prior knowledge of gig workers’ utilities and remains more robust under these conditions.

Overall, our algorithm consistently outperforms both the \gls{acr:pp} and \gls{acr:fp} benchmarks across all scenarios, yielding improvements between 2.5-7.5\%, with the degree of improvement varying based on the specific arrival rate conditions. We observe a negative gap of 1.4-4.7\% between our algorithm's performance and that of the Pert. ($\epsilon=0$) benchmark. This observed discrepancy can be attributed primarily to errors in estimating the gig worker's utility, and it is particularly pronounced in Scenario I.3, where the lower accuracy of the utility estimator impacts performance more than in the other two scenarios. As expected, the quality of the utility estimator directly influences our algorithm's performance. To further investigate the effect of gig worker utility estimation quality on our algorithm's performance, we conduct a sensitivity analysis over $\epsilon$ in the following.

\noindent \textbf{Result 1:} In a homogeneous gig worker population, our algorithm achieved performance ratios between 88.5\% and 94.7\% across various request arrival rates, consistently outperforming both the \gls{acr:pp} and \gls{acr:fp} benchmarks by 2.5-7.5\%. 

\begin{figure}[t!]%
    \centering
    \fontsize{10}{10}\selectfont
    \includegraphics[width=\textwidth]{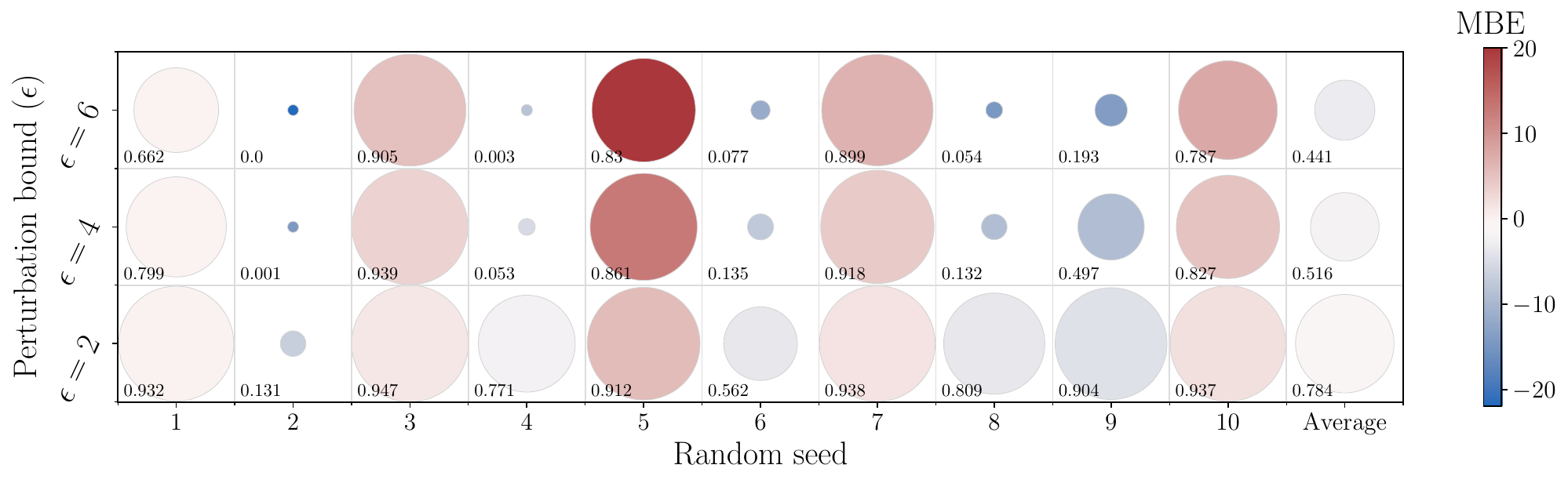}%
    \caption{\textnormal{Performance ratio and \gls{acr:mbe} for \gls{acr:p-pd-vfa} under various perturbation bounds $\epsilon$. The y-axis represents the different perturbation bounds, while the x-axis shows 10 perturbations, each sampled using a unique random seed. The numerical value inside each cell and the size of the circle indicate the achieved performance ratio. Meanwhile, the color gradient of the circles reflects the corresponding MBE of the estimated utilities for each perturbation. The final column shows the average performance ratio and MBE across all random seeds for each perturbation bound.}}
    \label{fig:sens_syn}%
\end{figure}

\noindent \textbf{Sensitivity analysis:} Figure \ref{fig:sens_syn} presents the results from a sensitivity analysis with respect to the estimates of the \gls{acr:mnl} model. Specifically, we evaluate the performance of the \gls{acr:p-pd-vfa} method under varying levels of perturbation. We vary the perturbation bound $\epsilon \in \{2, 4, 6\}$ across 10 different random seeds, each producing a distinct perturbation vector. Our results show that our method remains relatively stable against perturbations that cause underestimation of gig workers' utilities, as characterized by a positive \gls{acr:mbe}. For more details, we refer to Appendix \ref{sec:mbe}. However, when perturbations lead to substantial overestimations, indicated by a strongly negative \gls{acr:mbe}, performance can degrade significantly. In extreme cases almost no requests are accepted, leading to a performance ratio around zero. This aligns with expectations: if the utility model overestimates the utilities, it incorrectly assumes that requests are more attractive to gig workers than they actually are, resulting in insufficient compensation to incentivize acceptance. Conversely, underestimation provides enough compensation to ensure acceptance but reduces revenue, as a lower offer could have sufficed for gig worker incentivization. As expected, increasing the perturbation bound deteriorates performance in all cases, with a more pronounced effect when utilities are overestimated rather than underestimated.

We argue that in practical applications of our algorithm, where \gls{acr:mnl} parameters are estimated using real data, it is more likely for the estimation errors to result in underestimations rather than overestimations. This argument is based on the premise that for a model to systematically overestimate utilities, gig workers must consistently accept requests for compensations lower than their baseline utility for each request. Such occurrences are only observable due to the natural variability in gig workers' utility functions, stemming from the stochastic elements of their decision-making. In contrast, underestimating utility functions is more probable, as this can occur when data is collected under an improper policy, specifically, one that systematically over-offers compensation, prompting gig workers to select the most profitable options. In these scenarios, it becomes challenging to determine the true baseline of gig worker utilities from the data. 

\noindent \textbf{Result 2:} Perturbations to the true \gls{acr:mnl} parameters lead to performance degradation, with more pronounced effects when gig workers' utilities are overestimated.

\subsection{Heterogeneous gig worker population}
\label{sec:res_syn_2}
In the following, we evaluate the impact on algorithmic performance when incorporating knowledge about different gig worker groups compared to cases without such knowledge. To this end, we test two variations of our algorithm. In the first variation, we assume that the platform has no information about the existence of different gig worker groups, thus applying a single \gls{acr:mnl} model across all gig workers. In the second variation, we assume that the platform is aware of different gig worker groups, allowing us to use three separate \gls{acr:mnl} models (Multi \gls{acr:mnl}) to better reflect group-specific preferences. 

\begin{figure}[b!]
    \begin{minipage}{0.62\textwidth}
        \centering
        \subfloat[\centering Single \gls{acr:mnl} model]{{\adjustbox{width=0.49\textwidth}{\includegraphics{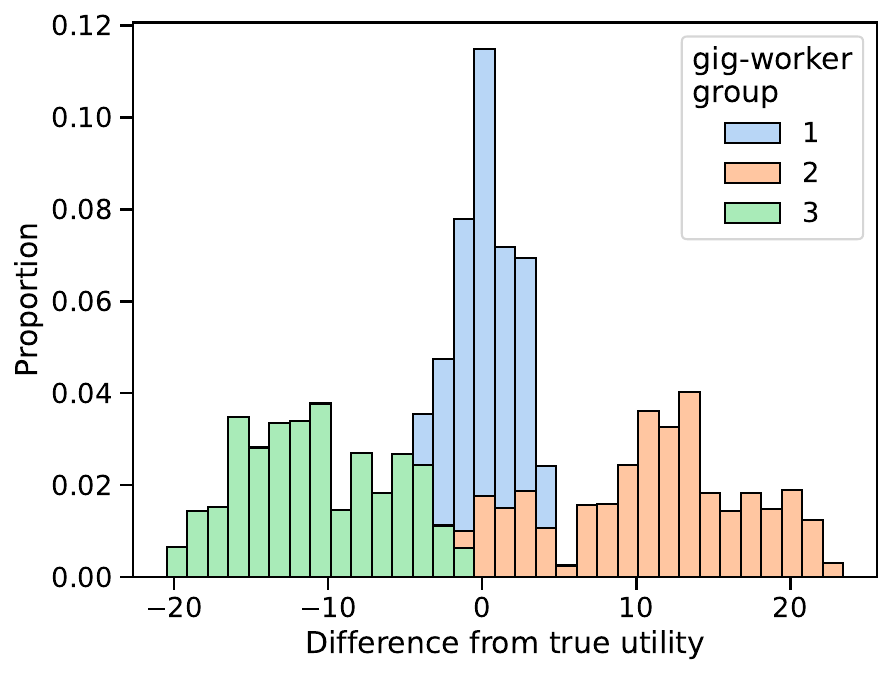}} }}%
        \hspace*{0.1cm} 
        \subfloat[\centering Multiple \gls{acr:mnl} models]{{\adjustbox{width=0.49\textwidth}{\includegraphics{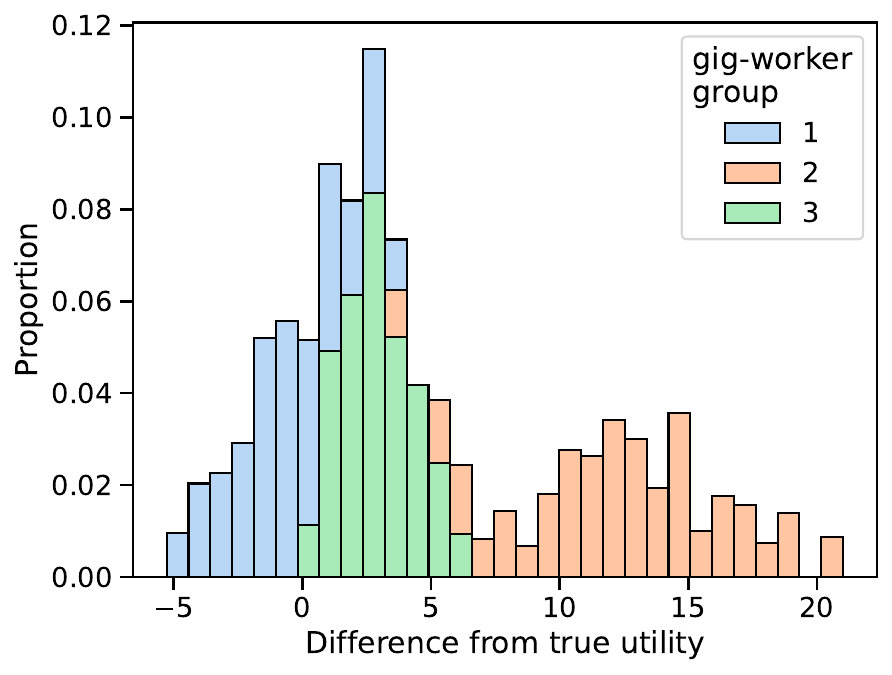}} }}%
        \caption{\textnormal{Difference between true utility and estimated utility when using a single \gls{acr:mnl} model (a) or multiple \gls{acr:mnl} models (b).}}
        \label{fig:mnl_momnl}
    \end{minipage}%
    \hspace*{0.02\textwidth} 
    \raisebox{0.1cm}{\begin{minipage}{0.34\textwidth}
        \centering
        \fontsize{10}{10}\selectfont
        \adjustbox{width=0.9\textwidth}{
\begin{tikzpicture}

\definecolor{chocolate22312431}{RGB}{223,124,31}
\definecolor{darkgray176}{RGB}{176,176,176}
\definecolor{darkorchid13765203}{RGB}{137,65,203}
\definecolor{darkslategray68}{RGB}{68,68,68}
\definecolor{firebrick2032937}{RGB}{203,29,37}
\definecolor{limegreen4717970}{RGB}{47,179,70}
\definecolor{royalblue3378223}{RGB}{33,78,223}

\begin{axis}[
tick align=outside,
tick pos=left,
x grid style={darkgray176},
xmin=-0.5, xmax=4.5,
xtick style={color=black},
xtick={0,1,2,3,4},
xticklabel style={rotate=15.0},
xticklabels={Pert. \(\displaystyle \epsilon=0\),Multi MNL,MNL,PP,FP},
y grid style={darkgray176},
ylabel={Performance ratio (\%)},
ymin=46.9641574196683, ymax=99.3026105344177,
ytick style={color=black},
ytick={40,50,60,70,80,90,100},
yticklabels={
  \(\displaystyle {40}\),
  \(\displaystyle {50}\),
  \(\displaystyle {60}\),
  \(\displaystyle {70}\),
  \(\displaystyle {80}\),
  \(\displaystyle {90}\),
  \(\displaystyle {100}\)
}
]
\path [draw=darkslategray68, fill=royalblue3378223]
(axis cs:-0.4,90.7945482505667)
--(axis cs:0.4,90.7945482505667)
--(axis cs:0.4,95.0250109392938)
--(axis cs:-0.4,95.0250109392938)
--(axis cs:-0.4,90.7945482505667)
--cycle;
\addplot [darkslategray68]
table {%
0 90.7945482505667
0 84.605982889528
};
\addplot [darkslategray68]
table {%
0 95.0250109392938
0 96.9235899382927
};
\addplot [darkslategray68]
table {%
-0.2 84.605982889528
0.2 84.605982889528
};
\addplot [darkslategray68]
table {%
-0.2 96.9235899382927
0.2 96.9235899382927
};
\addplot [black, mark=o, mark size=3, mark options={solid,fill opacity=0,draw=darkslategray68}, only marks]
table {%
0 83.8823287736362
};
\path [draw=darkslategray68, fill=chocolate22312431]
(axis cs:0.6,87.7259901591438)
--(axis cs:1.4,87.7259901591438)
--(axis cs:1.4,91.8675688551704)
--(axis cs:0.6,91.8675688551704)
--(axis cs:0.6,87.7259901591438)
--cycle;
\addplot [darkslategray68]
table {%
1 87.7259901591438
1 81.8339690381899
};
\addplot [darkslategray68]
table {%
1 91.8675688551704
1 94.4436556843885
};
\addplot [darkslategray68]
table {%
0.8 81.8339690381899
1.2 81.8339690381899
};
\addplot [darkslategray68]
table {%
0.8 94.4436556843885
1.2 94.4436556843885
};
\addplot [black, mark=o, mark size=3, mark options={solid,fill opacity=0,draw=darkslategray68}, only marks]
table {%
1 75.1728426862789
};
\path [draw=darkslategray68, fill=limegreen4717970]
(axis cs:1.6,70.1600619168688)
--(axis cs:2.4,70.1600619168688)
--(axis cs:2.4,81.9879878856474)
--(axis cs:1.6,81.9879878856474)
--(axis cs:1.6,70.1600619168688)
--cycle;
\addplot [darkslategray68]
table {%
2 70.1600619168688
2 56.7600264906113
};
\addplot [darkslategray68]
table {%
2 81.9879878856474
2 93.0645193576326
};
\addplot [darkslategray68]
table {%
1.8 56.7600264906113
2.2 56.7600264906113
};
\addplot [darkslategray68]
table {%
1.8 93.0645193576326
2.2 93.0645193576326
};
\addplot [black, mark=o, mark size=3, mark options={solid,fill opacity=0,draw=darkslategray68}, only marks]
table {%
2 49.3431780157933
2 50.5735419693593
};
\path [draw=darkslategray68, fill=firebrick2032937]
(axis cs:2.6,75.8449688825698)
--(axis cs:3.4,75.8449688825698)
--(axis cs:3.4,82.4022993199884)
--(axis cs:2.6,82.4022993199884)
--(axis cs:2.6,75.8449688825698)
--cycle;
\addplot [darkslategray68]
table {%
3 75.8449688825698
3 68.8724611986297
};
\addplot [darkslategray68]
table {%
3 82.4022993199884
3 87.3404238482497
};
\addplot [darkslategray68]
table {%
2.8 68.8724611986297
3.2 68.8724611986297
};
\addplot [darkslategray68]
table {%
2.8 87.3404238482497
3.2 87.3404238482497
};
\addplot [black, mark=o, mark size=3, mark options={solid,fill opacity=0,draw=darkslategray68}, only marks]
table {%
3 63.8000644930302
};
\path [draw=darkslategray68, fill=darkorchid13765203]
(axis cs:3.6,76.2521834388413)
--(axis cs:4.4,76.2521834388413)
--(axis cs:4.4,84.6809458763638)
--(axis cs:3.6,84.6809458763638)
--(axis cs:3.6,76.2521834388413)
--cycle;
\addplot [darkslategray68]
table {%
4 76.2521834388413
4 68.6061398541417
};
\addplot [darkslategray68]
table {%
4 84.6809458763638
4 90.4525199239857
};
\addplot [darkslategray68]
table {%
3.8 68.6061398541417
4.2 68.6061398541417
};
\addplot [darkslategray68]
table {%
3.8 90.4525199239857
4.2 90.4525199239857
};
\addplot [black, mark=o, mark size=3, mark options={solid,fill opacity=0,draw=darkslategray68}, only marks]
table {%
4 63.4092235156811
};
\addplot [darkslategray68]
table {%
-0.4 93.6450521490036
0.4 93.6450521490036
};
\addplot [darkslategray68]
table {%
0.6 90.486263555085
1.4 90.486263555085
};
\addplot [darkslategray68]
table {%
1.6 75.777345484792
2.4 75.777345484792
};
\addplot [darkslategray68]
table {%
2.6 79.9065168022921
3.4 79.9065168022921
};
\addplot [darkslategray68]
table {%
3.6 81.2345523599997
4.4 81.2345523599997
};
\draw (axis cs:0,33.879544140981) node[
  scale=0.75,
  anchor=base,
  text=black,
  rotate=0.0
]{\bfseries 92.63};
\draw (axis cs:1,33.879544140981) node[
  scale=0.75,
  anchor=base,
  text=black,
  rotate=0.0
]{\bfseries 89.57};
\draw (axis cs:2,33.879544140981) node[
  scale=0.75,
  anchor=base,
  text=black,
  rotate=0.0
]{\bfseries 75.53};
\draw (axis cs:3,33.879544140981) node[
  scale=0.75,
  anchor=base,
  text=black,
  rotate=0.0
]{\bfseries 79.24};
\draw (axis cs:4,33.879544140981) node[
  scale=0.75,
  anchor=base,
  text=black,
  rotate=0.0
]{\bfseries 80.48};
\draw (axis cs:-1,34.4029286721285) node[
  scale=0.75,
  text=black,
  rotate=0.0
]{\bfseries \textbf{Average:}};
\end{axis}

\end{tikzpicture}}
        \caption{\textnormal{Performance ratios in Scenario II}}%
        \label{fig:perf_syn_2}%
    \end{minipage}}
\end{figure}

Figure \ref{fig:mnl_momnl} illustrates the difference between the true utility and estimated utility when using either a single \gls{acr:mnl} or three separate \gls{acr:mnl} models. As expected, it is hard to accurately capture the true underlying utility model within a singular \gls{acr:mnl} model (Figure \ref{fig:mnl_momnl}a). The \gls{acr:rmse} between the estimated utility and the true utility is equal to 1.8, 12.7, and 11.7 for each respective group of gig worker, indicating that the predictive accuracy varies significantly among different gig worker groups. When employing a distinct \gls{acr:mnl} model for each group of gig worker, the \gls{acr:rmse} between the estimated utility and the true utility is equal to 2.2, 12.9, and 3.2 for each respective group of gig worker (Figure \ref{fig:mnl_momnl}b). The accuracy of utility estimation for gig worker group 3 has significantly improved, while the performance for gig worker group 1 remains comparable to that of the singular model. However, similar to the singular \gls{acr:mnl} model, the performance for gig worker group 2 remains relatively low, suggesting an inherent difficulty in estimating this group's utility. This difficulty stems from the fact that the groups have varying intrinsic utility levels, and accordingly require varying levels of compensation to be incentivized: specifically, group 2 demonstrates the highest average utility for requests compared to the other two groups (Figure~\ref{fig:utility_scen_2}), indicating that, they require the least compensation on average. As a result, when applying a uniform pricing policy across all gig worker groups during data collection for training the \gls{acr:mnl} model, we tend to overcompensate this group on average to ensure adequate incentives for the other two groups.

\noindent \textbf{Performance:} Figure~\ref{fig:perf_syn_2} shows the distribution of performance ratios across all test set realizations, including benchmarks \gls{acr:pp}, \gls{acr:fp}, and Pert. ($\epsilon=0$), alongside the two variations of our algorithm. When utilizing the correct number of \gls{acr:mnl} models, our algorithm demonstrates an average performance ratio of 89.5\% (±3.1\%), compared to 92.6\% (±3.1\%) for the Pert. ($\epsilon=0$), 79.2\% (±4.3\%) for the \gls{acr:pp}, and 80.4\% (±5.3\%) for the \gls{acr:fp} benchmark. In contrast, when employing a single \gls{acr:mnl} model, the average performance ratio drops significantly to 75.5\% (±9.0\%), and is inferior to all benchmark policies. Our results underscore the importance of having a reliable utility estimator, which requires the platform to understand the characteristics of gig workers to effectively differentiate between various groups. Notably, even though the multi-\gls{acr:mnl} estimator exhibits a relatively high error in estimating the utility for a specific gig worker group, it still manages to achieve satisfactory overall performance. This finding suggests that while having multiple models improves the algorithm's ability to capture utility variations, the platform must also focus on refining the accuracy of the utility estimations across different gig worker groups to further enhance overall performance.

\begin{figure}[t!]%
    \centering
    \fontsize{10}{10}\selectfont
    \includegraphics[width=\textwidth]{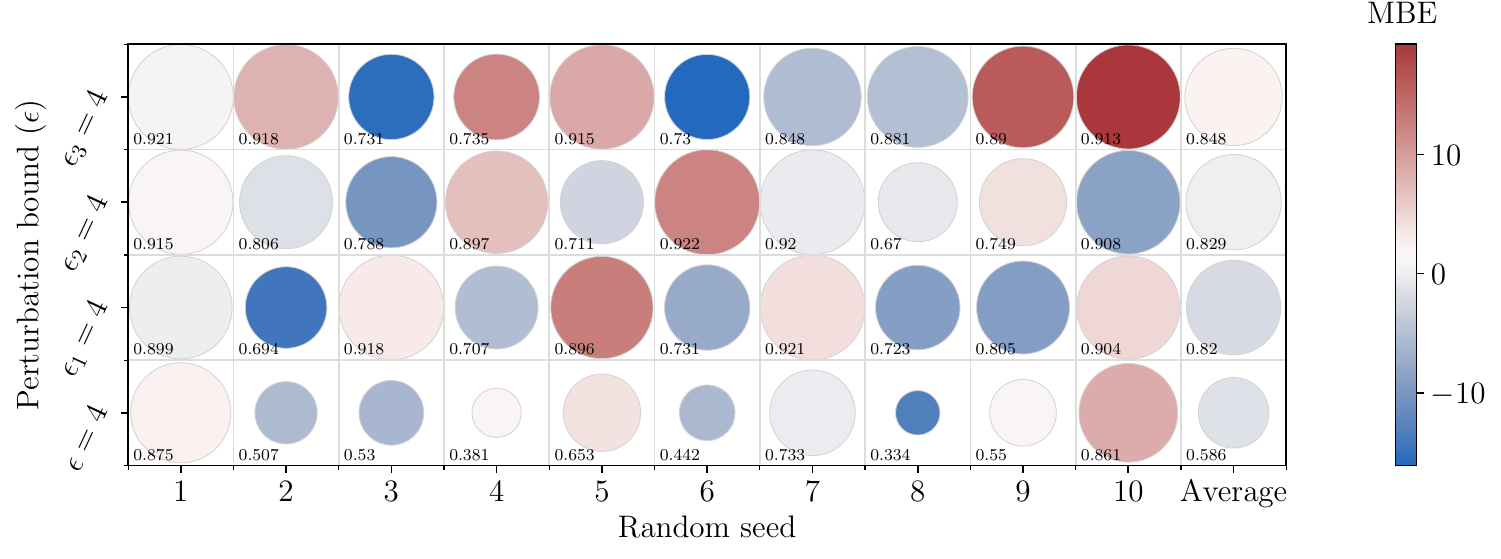}%
    \caption{\textnormal{Performance ratio and \gls{acr:mbe} for \gls{acr:p-pd-vfa} under various perturbation bounds $\epsilon$.  We denote the bound as $\epsilon_d$ in the case where only the parameters of gig worker group $d \in \{1,2,3\}$ are perturbed. The y-axis represents the different perturbation bounds, while the x-axis shows 10 perturbations, each sampled using a unique random seed. The numerical value inside each cell and the size of the circle indicate the achieved performance ratio. Meanwhile, the color gradient of the circles reflects the corresponding MBE of the estimated utilities for each perturbation. The final column shows the average performance ratio and MBE across all random seeds for each perturbation bound}}
    \label{fig:sens_syn_2}%
\end{figure}

\noindent \textbf{Result 3:} In a heterogeneous population, our algorithm achieved an average performance ratio of 89.5\%, outperforming both the \gls{acr:pp} and \gls{acr:fp} benchmarks by 9-10\%, when employing the correct number of \gls{acr:mnl} models. Using a single \gls{acr:mnl} model resulted in lower performance (75.5\%), further underscoring the importance of accurate utility estimation when dealing with diverse gig worker groups. 

\noindent \textbf{Sensitivity analysis:} In Figure \ref{fig:sens_syn_2}, we conduct a sensitivity analysis on the \gls{acr:mnl} model estimates by evaluating the performance of \gls{acr:p-pd-vfa} under varying perturbation bounds $\epsilon$. We denote the bound as $\epsilon_d$ in the case where only the parameters of gig worker group $d \in \{1,2,3\}$ are perturbed. Alternatively, we denote the bound as $\epsilon$ when the parameters of all groups are perturbed. As expected, perturbing the parameters of all gig worker groups leads to greater performance degradation, while perturbing the parameters of only one group results in relatively robust, albeit reduced, performance. Although perturbing the utility estimates of individual gig worker groups does negatively impact performance, the degree of degradation is similar across all groups. Consistent with the results from Section \ref{sec:res_syn}, perturbations that lead to overestimation of gig worker utilities have a more severe impact on performance compared to underestimations.

\noindent \textbf{Result 4:} Performance degrades as the accuracy of the gig worker utility estimator worsens, but the extent of degradation is relatively consistent across groups.

\subsection{New York Taxi Data}
\label{sec:res_nyt}
We now apply our algorithm to real-world data from New York taxi services, to test the robustness of the algorithm under more complex, realistic conditions. For gig workers with weak location preferences, the algorithm operates in an environment where gig workers are indifferent with regards to specific locations, thus representing simpler, non-rigid settings. In contrast, strong location preferences introduce additional complexity, as gig workers prioritize particular areas. This analysis provides deeper insights into the algorithm's behavior in cases where gig worker preferences over specific locations play a more decisive role in request acceptance.

\begin{figure}[b!]
    \centering
    \begin{minipage}{0.48\textwidth}
        \centering
        \fontsize{10}{10}\selectfont
        \subfloat[\centering Grouped by pickup location (weak preference)]{{\includegraphics[width=0.48\textwidth]{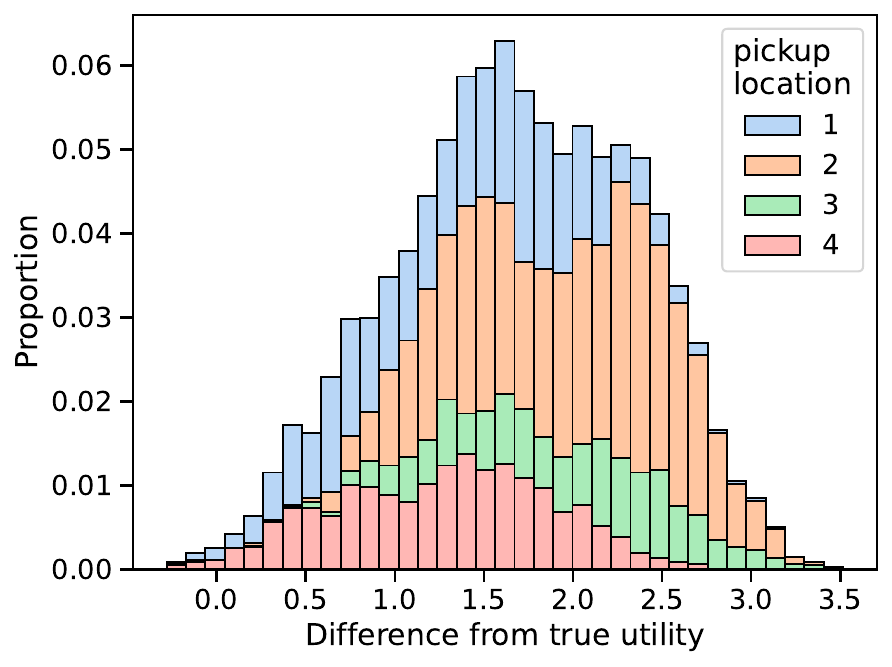}}}%
        \hspace*{0.1cm}
        \subfloat[\centering Grouped by dropoff location (weak preference)]{{\includegraphics[width=0.48\textwidth]{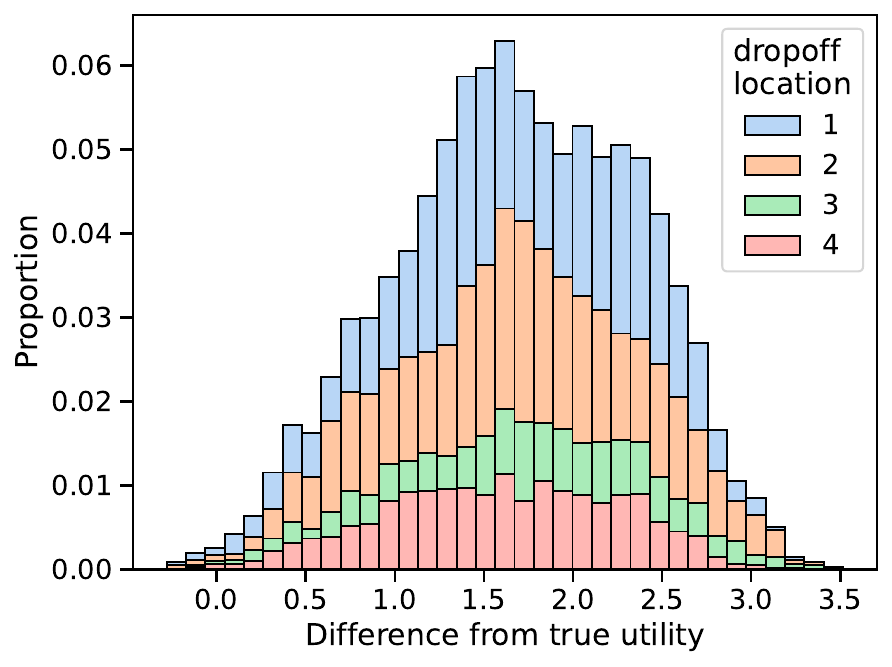}}}%
    \end{minipage}\hfill
    \begin{minipage}{0.48\textwidth}
        \centering
        \fontsize{10}{10}\selectfont
        \subfloat[\centering Grouped by pickup location (strong preference)]{{\includegraphics[width=0.48\textwidth]{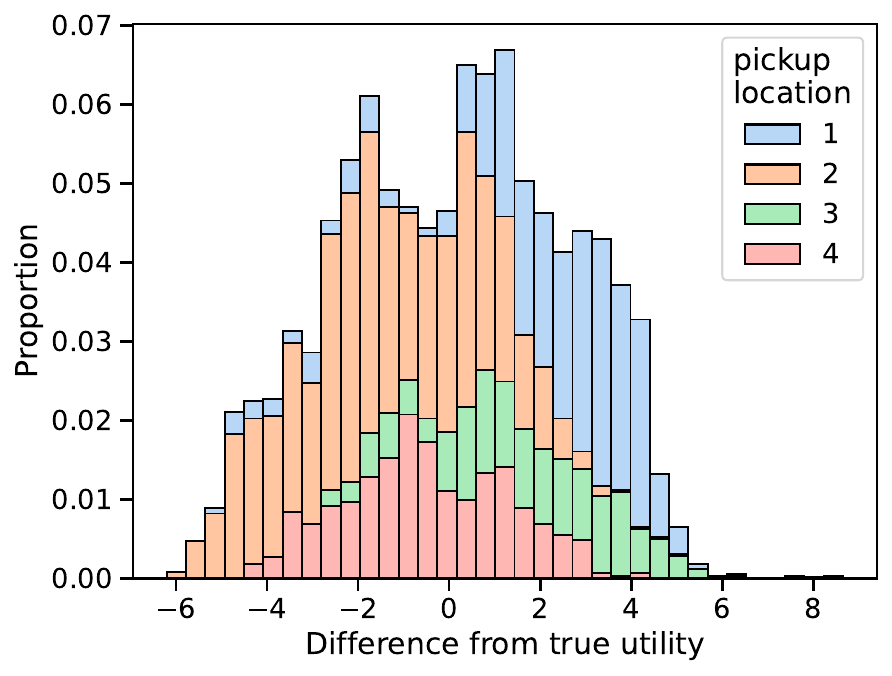}}}%
        \hspace*{0.1cm}
        \subfloat[\centering Grouped by dropoff location (strong preference)]{{\includegraphics[width=0.48\textwidth]{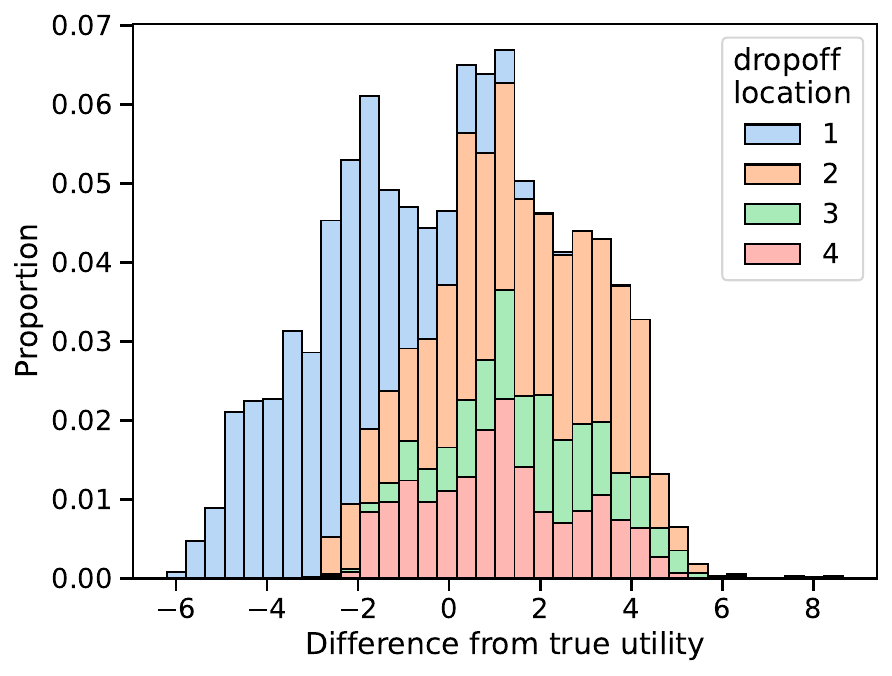}}}%
    \end{minipage}
    \caption{\textnormal{Difference between true utility and estimated utility grouped by pickup and dropoff locations for NYT data, with gig workers exhibiting weak location preference in (a) and (b), and strong location preference in (c) and (d).}}
    \label{fig:nyt_combined_mnl}
\end{figure}

The trained MNL model’s utility estimation accuracy varies across preference scenarios. Figure \ref{fig:nyt_combined_mnl} show histograms depicting the difference between the true and the estimated utility for NYT data with weak and strong location preference gig workers. In the weak preferences scenario (see Figure \ref{fig:nyt_combined_mnl}a\&b), the \gls{acr:mnl} model achieves a \gls{acr:rmse} of 1.8 and shows a slight average underestimation of request utilities with an overall \gls{acr:mbe} of 1.6. We do not observe any distinct pattern where requests with specific pickup or dropoff locations are significantly over- or under-estimated. In the strong preference scenario (Figure \ref{fig:nyt_combined_mnl}c\&d), the model faces greater complexity in estimating the \gls{acr:mnl} parameters. Despite the \gls{acr:mnl} model achieving a reasonably good \gls{acr:rmse} of 2.5 and an overall \gls{acr:mbe} of 0.08, we observe varying biases in the utility estimation of different subgroups of requests. For instance, we observe an overestimation of utilities for requests with dropoff location 1, whereas we observe an underestimation for other dropoff locations (Figure \ref{fig:nyt_combined_mnl}d). This scenario is one of the most challenging cases for accurate utility estimation, as it has the most divergent distribution in terms of utilities.

\begin{figure}[t!]
    \centering
    \captionsetup[subfloat]{labelformat=empty}
    \fontsize{9}{9}\selectfont
    \begin{minipage}{0.32\textwidth}
        \centering
        \subfloat[\vspace{-0.05cm}]{\adjustbox{width=\textwidth}{ 
\begin{tikzpicture}

\definecolor{chocolate22312431}{RGB}{223,124,31}
\definecolor{darkgray176}{RGB}{176,176,176}
\definecolor{darkslategray68}{RGB}{68,68,68}
\definecolor{firebrick2032937}{RGB}{203,29,37}
\definecolor{limegreen4717970}{RGB}{47,179,70}
\definecolor{royalblue3378223}{RGB}{33,78,223}

\begin{axis}[
tick align=outside,
tick pos=left,
x grid style={darkgray176},
xmin=-0.5, xmax=3.5,
xtick style={color=black},
xtick={0,1,2,3},
xticklabels={Pert. \(\displaystyle \epsilon=0\),MNL,PP,FP},
y grid style={darkgray176},
ylabel={Performance ratio (\%)},
ymin=72.5467977566071, ymax=96.2003769214691,
ytick style={color=black},
ytick={70,75,80,85,90,95,100},
yticklabels={
  \(\displaystyle {70}\),
  \(\displaystyle {75}\),
  \(\displaystyle {80}\),
  \(\displaystyle {85}\),
  \(\displaystyle {90}\),
  \(\displaystyle {95}\),
  \(\displaystyle {100}\)
}
]
\path [draw=darkslategray68, fill=royalblue3378223]
(axis cs:-0.4,93.0765634615927)
--(axis cs:0.4,93.0765634615927)
--(axis cs:0.4,94.8837217605645)
--(axis cs:-0.4,94.8837217605645)
--(axis cs:-0.4,93.0765634615927)
--cycle;
\addplot [darkslategray68]
table {%
0 93.0765634615927
0 91.8523982442826
};
\addplot [darkslategray68]
table {%
0 94.8837217605645
0 95.1252142321572
};
\addplot [darkslategray68]
table {%
-0.2 91.8523982442826
0.2 91.8523982442826
};
\addplot [darkslategray68]
table {%
-0.2 95.1252142321572
0.2 95.1252142321572
};
\path [draw=darkslategray68, fill=chocolate22312431]
(axis cs:0.6,92.0224894960145)
--(axis cs:1.4,92.0224894960145)
--(axis cs:1.4,93.6162686656397)
--(axis cs:0.6,93.6162686656397)
--(axis cs:0.6,92.0224894960145)
--cycle;
\addplot [darkslategray68]
table {%
1 92.0224894960145
1 90.2346005578498
};
\addplot [darkslategray68]
table {%
1 93.6162686656397
1 94.2259197773107
};
\addplot [darkslategray68]
table {%
0.8 90.2346005578498
1.2 90.2346005578498
};
\addplot [darkslategray68]
table {%
0.8 94.2259197773107
1.2 94.2259197773107
};
\addplot [black, mark=o, mark size=3, mark options={solid,fill opacity=0,draw=darkslategray68}, only marks]
table {%
1 89.347001997502
};
\path [draw=darkslategray68, fill=limegreen4717970]
(axis cs:1.6,80.7656222923379)
--(axis cs:2.4,80.7656222923379)
--(axis cs:2.4,85.3874874430389)
--(axis cs:1.6,85.3874874430389)
--(axis cs:1.6,80.7656222923379)
--cycle;
\addplot [darkslategray68]
table {%
2 80.7656222923379
2 77.6341964522543
};
\addplot [darkslategray68]
table {%
2 85.3874874430389
2 88.7043764630758
};
\addplot [darkslategray68]
table {%
1.8 77.6341964522543
2.2 77.6341964522543
};
\addplot [darkslategray68]
table {%
1.8 88.7043764630758
2.2 88.7043764630758
};
\addplot [black, mark=o, mark size=3, mark options={solid,fill opacity=0,draw=darkslategray68}, only marks]
table {%
2 73.621960445919
};
\path [draw=darkslategray68, fill=firebrick2032937]
(axis cs:2.6,82.5021621714283)
--(axis cs:3.4,82.5021621714283)
--(axis cs:3.4,87.0227157031759)
--(axis cs:2.6,87.0227157031759)
--(axis cs:2.6,82.5021621714283)
--cycle;
\addplot [darkslategray68]
table {%
3 82.5021621714283
3 77.6603881814643
};
\addplot [darkslategray68]
table {%
3 87.0227157031759
3 89.0406873356241
};
\addplot [darkslategray68]
table {%
2.8 77.6603881814643
3.2 77.6603881814643
};
\addplot [darkslategray68]
table {%
2.8 89.0406873356241
3.2 89.0406873356241
};
\addplot [darkslategray68]
table {%
-0.4 94.3656415323535
0.4 94.3656415323535
};
\addplot [darkslategray68]
table {%
0.6 92.7418460931945
1.4 92.7418460931945
};
\addplot [darkslategray68]
table {%
1.6 84.0058825970647
2.4 84.0058825970647
};
\addplot [darkslategray68]
table {%
2.6 85.0975996089361
3.4 85.0975996089361
};
\draw (axis cs:0,66.6334029653916) node[
  scale=0.75,
  anchor=base,
  text=black,
  rotate=0.0
]{\bfseries 94.00};
\draw (axis cs:1,66.6334029653916) node[
  scale=0.75,
  anchor=base,
  text=black,
  rotate=0.0
]{\bfseries 92.61};
\draw (axis cs:2,66.6334029653916) node[
  scale=0.75,
  anchor=base,
  text=black,
  rotate=0.0
]{\bfseries 83.05};
\draw (axis cs:3,66.6334029653916) node[
  scale=0.75,
  anchor=base,
  text=black,
  rotate=0.0
]{\bfseries 84.65};
\draw (axis cs:-1,66.8699387570403) node[
  scale=0.75,
  text=black,
  rotate=0.0
]{\bfseries \textbf{Average:}};
\end{axis}

\end{tikzpicture}}}
        \caption{\textnormal{Performance ratios for NYT data (weak preference)}}
        \label{fig:perf_nyt_weak}
    \end{minipage}
    \captionsetup[subfloat]{labelformat=parens} 
    \hspace*{0.1cm}
    \begin{minipage}{0.64\textwidth}
        \centering
        \subfloat[\centering Difference in offered compensation]{{\adjustbox{width=0.49\textwidth}{
\begin{tikzpicture}

\definecolor{chocolate22312431}{RGB}{223,124,31}
\definecolor{darkgray176}{RGB}{176,176,176}
\definecolor{darkslategray68}{RGB}{68,68,68}
\definecolor{firebrick2032937}{RGB}{203,29,37}
\definecolor{limegreen4717970}{RGB}{47,179,70}
\definecolor{royalblue3378223}{RGB}{33,78,223}

\begin{axis}[
tick align=outside,
tick pos=left,
x grid style={darkgray176},
xmin=-0.5, xmax=3.5,
xtick style={color=black},
xtick={0,1,2,3},
xticklabels={Pert. \(\displaystyle \epsilon=0\),MNL,PP,FP},
y grid style={darkgray176},
ylabel={Monetary units},
ymin=-6.96864948272705, ymax=54.4579187393188,
ytick style={color=black},
ytick={-10,0,10,20,30,40,50,60},
yticklabels={
  \(\displaystyle {\ensuremath{-}10}\),
  \(\displaystyle {0}\),
  \(\displaystyle {10}\),
  \(\displaystyle {20}\),
  \(\displaystyle {30}\),
  \(\displaystyle {40}\),
  \(\displaystyle {50}\),
  \(\displaystyle {60}\)
}
]
\path [draw=darkslategray68, fill=royalblue3378223]
(axis cs:-0.4,2.36447906494141)
--(axis cs:0.4,2.36447906494141)
--(axis cs:0.4,3.32570266723633)
--(axis cs:-0.4,3.32570266723633)
--(axis cs:-0.4,2.36447906494141)
--cycle;
\addplot [darkslategray68]
table {%
0 2.36447906494141
0 0.945648193359375
};
\addplot [darkslategray68]
table {%
0 3.32570266723633
0 4.60414123535156
};
\addplot [darkslategray68]
table {%
-0.2 0.945648193359375
0.2 0.945648193359375
};
\addplot [darkslategray68]
table {%
-0.2 4.60414123535156
0.2 4.60414123535156
};
\addplot [black, mark=o, mark size=3, mark options={solid,fill opacity=0,draw=darkslategray68}, only marks]
table {%
0 0.0480270385742188
0 -0.58233642578125
0 0.590114593505859
0 0.0474777221679688
0 0.473167419433594
0 0.646642684936523
0 0.159904479980469
0 0.678043365478516
0 0.229476928710938
0 -4.17653274536133
0 0.416534423828125
0 0.024566650390625
0 0.635429382324219
0 0.466636657714844
0 0.329971313476562
0 -0.340896606445312
0 -0.608551025390625
0 0.40887451171875
0 -0.849254608154297
0 -1.67988586425781
0 0.225910186767578
0 0.609138488769531
0 0.257419586181641
0 -1.42390441894531
0 0.68231201171875
0 0.364372253417969
0 -0.628273010253906
0 0.598533630371094
0 0.765205383300781
0 -0.412887573242188
0 0.670192718505859
0 0.381237030029297
0 -0.164932250976562
0 0.352832794189453
0 0.890525817871094
0 0.329719543457031
0 -0.994338989257812
0 0.459976196289062
0 -0.279548645019531
0 -0.632007598876953
0 0.3521728515625
0 -0.725673675537109
0 0.0569686889648438
0 0.359874725341797
0 0.562919616699219
0 -1.02316665649414
0 0.559417724609375
0 -0.417884826660156
0 -2.62622451782227
0 -0.507026672363281
0 0.073944091796875
0 -0.413215637207031
0 0.770946502685547
0 0.546897888183594
0 0.541557312011719
0 0.452713012695312
0 0.446205139160156
0 -1.08180236816406
0 0.589023590087891
0 0.850990295410156
};
\path [draw=darkslategray68, fill=chocolate22312431]
(axis cs:0.6,2.72096633911133)
--(axis cs:1.4,2.72096633911133)
--(axis cs:1.4,4.15716171264648)
--(axis cs:0.6,4.15716171264648)
--(axis cs:0.6,2.72096633911133)
--cycle;
\addplot [darkslategray68]
table {%
1 2.72096633911133
1 0.582527160644531
};
\addplot [darkslategray68]
table {%
1 4.15716171264648
1 5.81782531738281
};
\addplot [darkslategray68]
table {%
0.8 0.582527160644531
1.2 0.582527160644531
};
\addplot [darkslategray68]
table {%
0.8 5.81782531738281
1.2 5.81782531738281
};
\addplot [black, mark=o, mark size=3, mark options={solid,fill opacity=0,draw=darkslategray68}, only marks]
table {%
1 -0.0558090209960938
1 -1.10079956054688
1 0.554561614990234
1 -0.0296554565429688
1 0.462055206298828
1 -0.490493774414062
1 -0.433330535888672
1 -0.299003601074219
1 -1.60445404052734
1 -0.516349792480469
1 0.526008605957031
1 -0.238037109375
1 -0.240951538085938
1 -0.696323394775391
1 -0.255561828613281
1 0.0371932983398438
1 -1.59751510620117
1 0.489463806152344
1 -1.9185905456543
1 -2.25302886962891
1 -0.595813751220703
1 -0.969982147216797
1 -1.12958526611328
1 -1.92495727539062
1 -0.715530395507812
1 0.260841369628906
1 -0.304996490478516
1 -0.114124298095703
1 0.531196594238281
1 0.0920562744140625
};
\path [draw=darkslategray68, fill=limegreen4717970]
(axis cs:1.6,5.74074172973633)
--(axis cs:2.4,5.74074172973633)
--(axis cs:2.4,13.4214458465576)
--(axis cs:1.6,13.4214458465576)
--(axis cs:1.6,5.74074172973633)
--cycle;
\addplot [darkslategray68]
table {%
2 5.74074172973633
2 -3.60192108154297
};
\addplot [darkslategray68]
table {%
2 13.4214458465576
2 24.7544403076172
};
\addplot [darkslategray68]
table {%
1.8 -3.60192108154297
2.2 -3.60192108154297
};
\addplot [darkslategray68]
table {%
1.8 24.7544403076172
2.2 24.7544403076172
};
\addplot [black, mark=o, mark size=3, mark options={solid,fill opacity=0,draw=darkslategray68}, only marks]
table {%
2 31.841136932373
2 27.4061622619629
2 25.0264587402344
2 28.3835411071777
2 25.0276145935059
2 41.994384765625
2 26.4535522460938
2 29.6583023071289
2 28.4854202270508
2 30.5247650146484
2 31.6587448120117
2 28.4067611694336
2 25.2874603271484
2 26.3846015930176
2 27.0558319091797
2 30.2211036682129
2 28.8731575012207
};
\path [draw=darkslategray68, fill=firebrick2032937]
(axis cs:2.6,4.91100311279297)
--(axis cs:3.4,4.91100311279297)
--(axis cs:3.4,12.253246307373)
--(axis cs:2.6,12.253246307373)
--(axis cs:2.6,4.91100311279297)
--cycle;
\addplot [darkslategray68]
table {%
3 4.91100311279297
3 -4.00156402587891
};
\addplot [darkslategray68]
table {%
3 12.253246307373
3 22.9434814453125
};
\addplot [darkslategray68]
table {%
2.8 -4.00156402587891
3.2 -4.00156402587891
};
\addplot [darkslategray68]
table {%
2.8 22.9434814453125
3.2 22.9434814453125
};
\addplot [black, mark=o, mark size=3, mark options={solid,fill opacity=0,draw=darkslategray68}, only marks]
table {%
3 24.5218086242676
3 26.3654632568359
3 26.3880462646484
3 27.5759887695312
3 51.6658020019531
3 28.2884521484375
3 23.5855407714844
3 32.3109359741211
3 26.2455978393555
3 25.526725769043
3 25.5008926391602
3 23.408088684082
3 24.3065795898438
3 28.7025527954102
3 24.1070938110352
3 31.8087310791016
3 29.7452926635742
3 23.9744262695312
3 33.1860847473145
3 36.9113616943359
3 25.5370178222656
3 23.3439292907715
3 26.7060852050781
3 23.284065246582
3 38.4702339172363
};
\addplot [darkslategray68]
table {%
-0.4 2.95130920410156
0.4 2.95130920410156
};
\addplot [darkslategray68]
table {%
0.6 3.50237274169922
1.4 3.50237274169922
};
\addplot [darkslategray68]
table {%
1.6 9.6194953918457
2.4 9.6194953918457
};
\addplot [darkslategray68]
table {%
2.6 8.51543426513672
3.4 8.51543426513672
};
\end{axis}

\end{tikzpicture}}}}%
        \hspace*{0.1cm} 
        \subfloat[\centering Percentage of utilized gig workers]{{\adjustbox{width=0.49\textwidth}{
\begin{tikzpicture}

\definecolor{chocolate22312431}{RGB}{223,124,31}
\definecolor{darkgray176}{RGB}{176,176,176}
\definecolor{darkslategray68}{RGB}{68,68,68}
\definecolor{firebrick2032937}{RGB}{203,29,37}
\definecolor{limegreen4717970}{RGB}{47,179,70}
\definecolor{royalblue3378223}{RGB}{33,78,223}

\begin{axis}[
tick align=outside,
tick pos=left,
x grid style={darkgray176},
xmin=-0.5, xmax=3.5,
xtick style={color=black},
xtick={0,1,2,3},
xticklabels={Pert. \(\displaystyle \epsilon=0\),MNL,PP,FP},
y grid style={darkgray176},
ylabel={Percentage},
ymin=0.131064102564103, ymax=0.987653846153846,
ytick style={color=black},
ytick={0,0.2,0.4,0.6,0.8,1},
yticklabels={
  \(\displaystyle {0.0}\),
  \(\displaystyle {0.2}\),
  \(\displaystyle {0.4}\),
  \(\displaystyle {0.6}\),
  \(\displaystyle {0.8}\),
  \(\displaystyle {1.0}\)
}
]
\path [draw=darkslategray68, fill=royalblue3378223]
(axis cs:-0.4,0.409082500924898)
--(axis cs:0.4,0.409082500924898)
--(axis cs:0.4,0.697389063200359)
--(axis cs:-0.4,0.697389063200359)
--(axis cs:-0.4,0.409082500924898)
--cycle;
\addplot [darkslategray68]
table {%
0 0.409082500924898
0 0.17
};
\addplot [darkslategray68]
table {%
0 0.697389063200359
0 0.948717948717949
};
\addplot [darkslategray68]
table {%
-0.2 0.17
0.2 0.17
};
\addplot [darkslategray68]
table {%
-0.2 0.948717948717949
0.2 0.948717948717949
};
\path [draw=darkslategray68, fill=chocolate22312431]
(axis cs:0.6,0.407476686185209)
--(axis cs:1.4,0.407476686185209)
--(axis cs:1.4,0.684222321828776)
--(axis cs:0.6,0.684222321828776)
--(axis cs:0.6,0.407476686185209)
--cycle;
\addplot [darkslategray68]
table {%
1 0.407476686185209
1 0.17
};
\addplot [darkslategray68]
table {%
1 0.684222321828776
1 0.923076923076923
};
\addplot [darkslategray68]
table {%
0.8 0.17
1.2 0.17
};
\addplot [darkslategray68]
table {%
0.8 0.923076923076923
1.2 0.923076923076923
};
\path [draw=darkslategray68, fill=limegreen4717970]
(axis cs:1.6,0.400401214487096)
--(axis cs:2.4,0.400401214487096)
--(axis cs:2.4,0.659233527566114)
--(axis cs:1.6,0.659233527566114)
--(axis cs:1.6,0.400401214487096)
--cycle;
\addplot [darkslategray68]
table {%
2 0.400401214487096
2 0.17
};
\addplot [darkslategray68]
table {%
2 0.659233527566114
2 0.825581395348837
};
\addplot [darkslategray68]
table {%
1.8 0.17
2.2 0.17
};
\addplot [darkslategray68]
table {%
1.8 0.825581395348837
2.2 0.825581395348837
};
\path [draw=darkslategray68, fill=firebrick2032937]
(axis cs:2.6,0.404603122966818)
--(axis cs:3.4,0.404603122966818)
--(axis cs:3.4,0.665313299232737)
--(axis cs:2.6,0.665313299232737)
--(axis cs:2.6,0.404603122966818)
--cycle;
\addplot [darkslategray68]
table {%
3 0.404603122966818
3 0.17
};
\addplot [darkslategray68]
table {%
3 0.665313299232737
3 0.872093023255814
};
\addplot [darkslategray68]
table {%
2.8 0.17
3.2 0.17
};
\addplot [darkslategray68]
table {%
2.8 0.872093023255814
3.2 0.872093023255814
};
\addplot [darkslategray68]
table {%
-0.4 0.544820336391437
0.4 0.544820336391437
};
\addplot [darkslategray68]
table {%
0.6 0.539299242424242
1.4 0.539299242424242
};
\addplot [darkslategray68]
table {%
1.6 0.519677033492823
2.4 0.519677033492823
};
\addplot [darkslategray68]
table {%
2.6 0.527935606060606
3.4 0.527935606060606
};
\end{axis}

\end{tikzpicture}}}}%
        \caption{\textnormal{Difference in offered compensation from true utility (a) and percentage of utilized gig workers (b) for NYT data (weak preference).}}
        \label{fig:comp_nyt_weak}
    \end{minipage}
\end{figure}
\begin{figure}[t!]
    \centering
    \captionsetup[subfloat]{labelformat=empty}
    \fontsize{9}{9}\selectfont
    \begin{minipage}{0.32\textwidth}
        \centering
        \subfloat[\vspace{-0.05cm}]{\adjustbox{width=\textwidth}{
\begin{tikzpicture}

\definecolor{chocolate22312431}{RGB}{223,124,31}
\definecolor{darkgray176}{RGB}{176,176,176}
\definecolor{darkslategray68}{RGB}{68,68,68}
\definecolor{firebrick2032937}{RGB}{203,29,37}
\definecolor{limegreen4717970}{RGB}{47,179,70}
\definecolor{royalblue3378223}{RGB}{33,78,223}

\begin{axis}[
tick align=outside,
tick pos=left,
x grid style={darkgray176},
xmin=-0.5, xmax=3.5,
xtick style={color=black},
xtick={0,1,2,3},
xticklabels={Pert. \(\displaystyle \epsilon=0\),MNL,PP,FP},
y grid style={darkgray176},
ylabel={Performance ratio (\%)},
ymin=51.7327333589047, ymax=95.705322244542,
ytick style={color=black},
ytick={50,60,70,80,90,100},
yticklabels={
  \(\displaystyle {50}\),
  \(\displaystyle {60}\),
  \(\displaystyle {70}\),
  \(\displaystyle {80}\),
  \(\displaystyle {90}\),
  \(\displaystyle {100}\)
}
]
\path [draw=darkslategray68, fill=royalblue3378223]
(axis cs:-0.4,90.450776630759)
--(axis cs:0.4,90.450776630759)
--(axis cs:0.4,92.4829044807425)
--(axis cs:-0.4,92.4829044807425)
--(axis cs:-0.4,90.450776630759)
--cycle;
\addplot [darkslategray68]
table {%
0 90.450776630759
0 88.1720209517941
};
\addplot [darkslategray68]
table {%
0 92.4829044807425
0 93.4733527349877
};
\addplot [darkslategray68]
table {%
-0.2 88.1720209517941
0.2 88.1720209517941
};
\addplot [darkslategray68]
table {%
-0.2 93.4733527349877
0.2 93.4733527349877
};
\path [draw=darkslategray68, fill=chocolate22312431]
(axis cs:0.6,83.370784409605)
--(axis cs:1.4,83.370784409605)
--(axis cs:1.4,87.9421723344371)
--(axis cs:0.6,87.9421723344371)
--(axis cs:0.6,83.370784409605)
--cycle;
\addplot [darkslategray68]
table {%
1 83.370784409605
1 78.3111257451702
};
\addplot [darkslategray68]
table {%
1 87.9421723344371
1 93.7065682042857
};
\addplot [darkslategray68]
table {%
0.8 78.3111257451702
1.2 78.3111257451702
};
\addplot [darkslategray68]
table {%
0.8 93.7065682042857
1.2 93.7065682042857
};
\path [draw=darkslategray68, fill=limegreen4717970]
(axis cs:1.6,58.1983222317439)
--(axis cs:2.4,58.1983222317439)
--(axis cs:2.4,61.9153945878929)
--(axis cs:1.6,61.9153945878929)
--(axis cs:1.6,58.1983222317439)
--cycle;
\addplot [darkslategray68]
table {%
2 58.1983222317439
2 53.731487399161
};
\addplot [darkslategray68]
table {%
2 61.9153945878929
2 67.4878542291351
};
\addplot [darkslategray68]
table {%
1.8 53.731487399161
2.2 53.731487399161
};
\addplot [darkslategray68]
table {%
1.8 67.4878542291351
2.2 67.4878542291351
};
\addplot [black, mark=o, mark size=3, mark options={solid,fill opacity=0,draw=darkslategray68}, only marks]
table {%
2 72.1973492131034
};
\path [draw=darkslategray68, fill=firebrick2032937]
(axis cs:2.6,64.2016330129495)
--(axis cs:3.4,64.2016330129495)
--(axis cs:3.4,67.1314118422844)
--(axis cs:2.6,67.1314118422844)
--(axis cs:2.6,64.2016330129495)
--cycle;
\addplot [darkslategray68]
table {%
3 64.2016330129495
3 61.7273700318582
};
\addplot [darkslategray68]
table {%
3 67.1314118422844
3 70.0340493056789
};
\addplot [darkslategray68]
table {%
2.8 61.7273700318582
3.2 61.7273700318582
};
\addplot [darkslategray68]
table {%
2.8 70.0340493056789
3.2 70.0340493056789
};
\addplot [black, mark=o, mark size=3, mark options={solid,fill opacity=0,draw=darkslategray68}, only marks]
table {%
3 58.542062452697
3 59.6589528057897
3 58.8263570050731
3 58.3384168869763
3 71.833140233908
};
\addplot [darkslategray68]
table {%
-0.4 91.6217044635638
0.4 91.6217044635638
};
\addplot [darkslategray68]
table {%
0.6 84.7378846675944
1.4 84.7378846675944
};
\addplot [darkslategray68]
table {%
1.6 60.118417064117
2.4 60.118417064117
};
\addplot [darkslategray68]
table {%
2.6 66.0223193072873
3.4 66.0223193072873
};
\draw (axis cs:0,40.7395861374954) node[
  scale=0.75,
  anchor=base,
  text=black,
  rotate=0.0
]{\bfseries 91.34};
\draw (axis cs:1,40.7395861374954) node[
  scale=0.75,
  anchor=base,
  text=black,
  rotate=0.0
]{\bfseries 85.39};
\draw (axis cs:2,40.7395861374954) node[
  scale=0.75,
  anchor=base,
  text=black,
  rotate=0.0
]{\bfseries 60.33};
\draw (axis cs:3,40.7395861374954) node[
  scale=0.75,
  anchor=base,
  text=black,
  rotate=0.0
]{\bfseries 65.20};
\draw (axis cs:-1,41.1793120263518) node[
  scale=0.75,
  text=black,
  rotate=0.0
]{\bfseries \textbf{Average:}};
\end{axis}

\end{tikzpicture}}}
        \caption{\textnormal{Performance ratios for NYT data (strong preference)}}
        \label{fig:perf_nyt_strong}
    \end{minipage}%
    \captionsetup[subfloat]{labelformat=parens} 
    \hspace*{0.1cm}
    \begin{minipage}{0.64\textwidth}
        \centering
        \subfloat[\centering Difference in offered compensation]{{\adjustbox{width=0.49\textwidth}{
\begin{tikzpicture}

\definecolor{chocolate22312431}{RGB}{223,124,31}
\definecolor{darkgray176}{RGB}{176,176,176}
\definecolor{darkslategray68}{RGB}{68,68,68}
\definecolor{firebrick2032937}{RGB}{203,29,37}
\definecolor{limegreen4717970}{RGB}{47,179,70}
\definecolor{royalblue3378223}{RGB}{33,78,223}

\begin{axis}[
tick align=outside,
tick pos=left,
x grid style={darkgray176},
xmin=-0.5, xmax=3.5,
xtick style={color=black},
xtick={0,1,2,3},
xticklabels={Pert. \(\displaystyle \epsilon=0\),MNL,PP,FP},
y grid style={darkgray176},
ylabel={Monetary units},
ymin=-7.72179203033447, ymax=77.8589876174927,
ytick style={color=black},
ytick={-20,0,20,40,60,80},
yticklabels={
  \(\displaystyle {\ensuremath{-}20}\),
  \(\displaystyle {0}\),
  \(\displaystyle {20}\),
  \(\displaystyle {40}\),
  \(\displaystyle {60}\),
  \(\displaystyle {80}\)
}
]
\path [draw=darkslategray68, fill=royalblue3378223]
(axis cs:-0.4,2.9080810546875)
--(axis cs:0.4,2.9080810546875)
--(axis cs:0.4,3.87141418457031)
--(axis cs:-0.4,3.87141418457031)
--(axis cs:-0.4,2.9080810546875)
--cycle;
\addplot [darkslategray68]
table {%
0 2.9080810546875
0 1.46649169921875
};
\addplot [darkslategray68]
table {%
0 3.87141418457031
0 4.56305313110352
};
\addplot [darkslategray68]
table {%
-0.2 1.46649169921875
0.2 1.46649169921875
};
\addplot [darkslategray68]
table {%
-0.2 4.56305313110352
0.2 4.56305313110352
};
\addplot [black, mark=o, mark size=3, mark options={solid,fill opacity=0,draw=darkslategray68}, only marks]
table {%
0 1.31788635253906
0 1.15647888183594
0 -0.979782104492188
0 0.157768249511719
0 0.0550308227539062
0 0.712020874023438
0 0.814285278320312
0 0.712867736816406
0 0.907318115234375
0 1.04023742675781
0 0.919090270996094
0 1.3988037109375
0 1.16722869873047
0 1.21726989746094
0 0.101913452148438
0 0.631832122802734
0 1.10314178466797
0 1.32956695556641
0 0.535453796386719
0 1.34599304199219
0 0.836349487304688
0 1.05873107910156
0 0.913238525390625
0 -0.0850448608398438
0 -0.017578125
0 -0.0562362670898438
0 1.41719055175781
0 0.717041015625
0 1.29923248291016
0 1.07136917114258
0 -2.47638702392578
0 -0.09478759765625
0 0.164253234863281
0 -0.094573974609375
0 0.803794860839844
0 -0.842056274414062
0 -2.45515441894531
0 0.494766235351562
0 -0.927444458007812
0 0.393043518066406
0 0.828407287597656
0 1.25785827636719
0 1.23259735107422
0 1.05279541015625
0 0.796051025390625
0 1.30381774902344
0 0.734695434570312
0 1.30718231201172
};
\path [draw=darkslategray68, fill=chocolate22312431]
(axis cs:0.6,1.12636947631836)
--(axis cs:1.4,1.12636947631836)
--(axis cs:1.4,4.80079746246338)
--(axis cs:0.6,4.80079746246338)
--(axis cs:0.6,1.12636947631836)
--cycle;
\addplot [darkslategray68]
table {%
1 1.12636947631836
1 -3.25681304931641
};
\addplot [darkslategray68]
table {%
1 4.80079746246338
1 8.09512710571289
};
\addplot [darkslategray68]
table {%
0.8 -3.25681304931641
1.2 -3.25681304931641
};
\addplot [darkslategray68]
table {%
0.8 8.09512710571289
1.2 8.09512710571289
};
\path [draw=darkslategray68, fill=limegreen4717970]
(axis cs:1.6,11.7126359939575)
--(axis cs:2.4,11.7126359939575)
--(axis cs:2.4,38.8348731994629)
--(axis cs:1.6,38.8348731994629)
--(axis cs:1.6,11.7126359939575)
--cycle;
\addplot [darkslategray68]
table {%
2 11.7126359939575
2 -3.72576141357422
};
\addplot [darkslategray68]
table {%
2 38.8348731994629
2 71.0504455566406
};
\addplot [darkslategray68]
table {%
1.8 -3.72576141357422
2.2 -3.72576141357422
};
\addplot [darkslategray68]
table {%
1.8 71.0504455566406
2.2 71.0504455566406
};
\path [draw=darkslategray68, fill=firebrick2032937]
(axis cs:2.6,8.43429946899414)
--(axis cs:3.4,8.43429946899414)
--(axis cs:3.4,27.4146118164062)
--(axis cs:2.6,27.4146118164062)
--(axis cs:2.6,8.43429946899414)
--cycle;
\addplot [darkslategray68]
table {%
3 8.43429946899414
3 -3.83175659179688
};
\addplot [darkslategray68]
table {%
3 27.4146118164062
3 55.3362007141113
};
\addplot [darkslategray68]
table {%
2.8 -3.83175659179688
3.2 -3.83175659179688
};
\addplot [darkslategray68]
table {%
2.8 55.3362007141113
3.2 55.3362007141113
};
\addplot [black, mark=o, mark size=3, mark options={solid,fill opacity=0,draw=darkslategray68}, only marks]
table {%
3 73.9689521789551
3 56.1101303100586
};
\addplot [darkslategray68]
table {%
-0.4 3.49254608154297
0.4 3.49254608154297
};
\addplot [darkslategray68]
table {%
0.6 3.0895824432373
1.4 3.0895824432373
};
\addplot [darkslategray68]
table {%
1.6 24.8769016265869
2.4 24.8769016265869
};
\addplot [darkslategray68]
table {%
2.6 18.0128326416016
3.4 18.0128326416016
};
\end{axis}

\end{tikzpicture}}}}%
        \hspace*{0.1cm} 
        \subfloat[\centering Percentage of utilized gig workers]{{\adjustbox{width=0.49\textwidth}{
\begin{tikzpicture}

\definecolor{chocolate22312431}{RGB}{223,124,31}
\definecolor{darkgray176}{RGB}{176,176,176}
\definecolor{darkslategray68}{RGB}{68,68,68}
\definecolor{firebrick2032937}{RGB}{203,29,37}
\definecolor{limegreen4717970}{RGB}{47,179,70}
\definecolor{royalblue3378223}{RGB}{33,78,223}

\begin{axis}[
tick align=outside,
tick pos=left,
x grid style={darkgray176},
xmin=-0.5, xmax=3.5,
xtick style={color=black},
xtick={0,1,2,3},
xticklabels={Pert. \(\displaystyle \epsilon=0\),MNL,PP,FP},
y grid style={darkgray176},
ylabel={Percentage},
ymin=0.0981080131208997, ymax=0.888185332708529,
ytick style={color=black},
ytick={0,0.1,0.2,0.3,0.4,0.5,0.6,0.7,0.8,0.9},
yticklabels={
  \(\displaystyle {0.0}\),
  \(\displaystyle {0.1}\),
  \(\displaystyle {0.2}\),
  \(\displaystyle {0.3}\),
  \(\displaystyle {0.4}\),
  \(\displaystyle {0.5}\),
  \(\displaystyle {0.6}\),
  \(\displaystyle {0.7}\),
  \(\displaystyle {0.8}\),
  \(\displaystyle {0.9}\)
}
]
\path [draw=darkslategray68, fill=royalblue3378223]
(axis cs:-0.4,0.423202614379085)
--(axis cs:0.4,0.423202614379085)
--(axis cs:0.4,0.697530864197531)
--(axis cs:-0.4,0.697530864197531)
--(axis cs:-0.4,0.423202614379085)
--cycle;
\addplot [darkslategray68]
table {%
0 0.423202614379085
0 0.175257731958763
};
\addplot [darkslategray68]
table {%
0 0.697530864197531
0 0.852272727272727
};
\addplot [darkslategray68]
table {%
-0.2 0.175257731958763
0.2 0.175257731958763
};
\addplot [darkslategray68]
table {%
-0.2 0.852272727272727
0.2 0.852272727272727
};
\path [draw=darkslategray68, fill=chocolate22312431]
(axis cs:0.6,0.38578431372549)
--(axis cs:1.4,0.38578431372549)
--(axis cs:1.4,0.589472624798712)
--(axis cs:0.6,0.589472624798712)
--(axis cs:0.6,0.38578431372549)
--cycle;
\addplot [darkslategray68]
table {%
1 0.38578431372549
1 0.144329896907216
};
\addplot [darkslategray68]
table {%
1 0.589472624798712
1 0.738636363636364
};
\addplot [darkslategray68]
table {%
0.8 0.144329896907216
1.2 0.144329896907216
};
\addplot [darkslategray68]
table {%
0.8 0.738636363636364
1.2 0.738636363636364
};
\path [draw=darkslategray68, fill=limegreen4717970]
(axis cs:1.6,0.382843137254902)
--(axis cs:2.4,0.382843137254902)
--(axis cs:2.4,0.611949812130972)
--(axis cs:1.6,0.611949812130972)
--(axis cs:1.6,0.382843137254902)
--cycle;
\addplot [darkslategray68]
table {%
2 0.382843137254902
2 0.134020618556701
};
\addplot [darkslategray68]
table {%
2 0.611949812130972
2 0.727272727272727
};
\addplot [darkslategray68]
table {%
1.8 0.134020618556701
2.2 0.134020618556701
};
\addplot [darkslategray68]
table {%
1.8 0.727272727272727
2.2 0.727272727272727
};
\path [draw=darkslategray68, fill=firebrick2032937]
(axis cs:2.6,0.346078431372549)
--(axis cs:3.4,0.346078431372549)
--(axis cs:3.4,0.566435532233883)
--(axis cs:2.6,0.566435532233883)
--(axis cs:2.6,0.346078431372549)
--cycle;
\addplot [darkslategray68]
table {%
3 0.346078431372549
3 0.154639175257732
};
\addplot [darkslategray68]
table {%
3 0.566435532233883
3 0.704545454545455
};
\addplot [darkslategray68]
table {%
2.8 0.154639175257732
3.2 0.154639175257732
};
\addplot [darkslategray68]
table {%
2.8 0.704545454545455
3.2 0.704545454545455
};
\addplot [darkslategray68]
table {%
-0.4 0.557369464639789
0.4 0.557369464639789
};
\addplot [darkslategray68]
table {%
0.6 0.487512487512488
1.4 0.487512487512488
};
\addplot [darkslategray68]
table {%
1.6 0.477522477522478
2.4 0.477522477522478
};
\addplot [darkslategray68]
table {%
2.6 0.487318563789152
3.4 0.487318563789152
};
\end{axis}

\end{tikzpicture}}}}%
        \caption{\textnormal{Difference in offered compensation from true utility (a) and percentage of utilized gig workers (b) for NYT data (strong preference).}}
        \label{fig:comp_nyt_strong}
    \end{minipage}
\end{figure}

\noindent \textbf{Performance:} Figures \ref{fig:perf_nyt_weak} and \ref{fig:perf_nyt_strong} present boxplots comparing the performance ratios across various benchmarks using weak and strong preference scenarios, respectively. For the weak preference scenario, our algorithm achieves an average performance ratio of 92.6\% (±1.3\%), demonstrating competitive results compared to the Pert. ($\epsilon=0$) benchmark, which attains the highest average performance ratio of 94.0\% (±1.1\%). In comparison, the \gls{acr:pp} benchmark achieves an average performance ratio of 83.0\% (±3.8\%), and the \gls{acr:fp} benchmark yields 84.6\% (±3.0\%). Therefore, our algorithm consistently outperforms the \gls{acr:pp} and \gls{acr:fp} benchmarks. Furthermore, it exhibits low variability and consistent performance, as evidenced by the small standard deviation, and achieves only slightly lower performance than the Pert. ($\epsilon=0$) benchmark. For the strong preference scenario, our algorithm achieves an average performance ratio of 85.3\% (±3.7\%), demonstrating substantial competitiveness, though lower than the Pert. ($\epsilon=0$) benchmark at 91.3\% (±1.6\%). In this scenario, the \gls{acr:pp} and \gls{acr:fp} benchmarks show significantly lower ratios of 60.3\% (±4.2\%) and 65.2\%~(±3.9\%), respectively. Notably, this case shows the greatest relative improvement over benchmark policies among all tested scenarios, highlighting the algorithm's adaptability to more complex preference structures. We attribute the observed gap between our algorithm and the Pert. ($\epsilon=0$) benchmark to challenges in estimating \gls{acr:mnl} parameters under stronger preferences; further refinement in utility estimation could likely enhance performance, aligning the results more closely with the benchmark.
\noindent \textbf{Result 5:} For NYT data, our algorithm achieves performance ratios of 92.6\% for weak and 85.3\% for strong preference scenarios, outperforming rule-based benchmarks by 8–9\% and 20–25\%, respectively.

\noindent \textbf{Compensation and utilized gig workers:} In the following, we examine the offered compensation and the percentage of utilized gig workers. Figures \ref{fig:comp_nyt_weak} and \ref{fig:comp_nyt_strong} display operational metrics for NYT data using gig workers with weak and strong location preferences, respectively. In each figure, part (a) shows the difference between the offered compensation and the deterministic part of the true utility for all accepted requests, while part (b) shows the percentage of utilized gig workers for each policy. We note that the negative values in Figures \ref{fig:comp_nyt_weak}a and \ref{fig:comp_nyt_strong}a are reasonable, as the analysis considers only the deterministic part of the utility, excluding the random component. Consequently, acceptance of requests is still possible if the addition of the stochastic component results in a positive overall utility, which is indeed the case for these observations.  

In the weak preference scenario, the average utilization rate of gig workers remains relatively consistent across all models, at approximately 55\%, with only minor variations of 1–2\%. However, the policies vary in terms of the difference in offered compensation from the true utility. Specifically, the Pert. ($\epsilon=0$) benchmark has a mean difference of only 2.7 monetary units, our algorithm has a mean difference of 3.3 monetary units, and the \gls{acr:pp} and \gls{acr:fp} algorithms have mean differences of 9.8 and 9 monetary units respectively. From these observations, we can conclude that the variation in total reward is mainly due to the lost revenue from offering more compensation than necessary, and not from utilizing a significantly larger amount of gig workers. Therefore, we observe that our algorithm achieves a balance between compensating gig workers and maintaining platform profitability, without the need to overcompensate gig workers to ensure their incentivation.

Considering the strong preference scenario, the average percentage of utilized gig workers is 55\% for the Pert.~($\epsilon=0$) benchmark, 48\% for both our algorithm and the \gls{acr:pp} benchmark, and 46\% for the \gls{acr:fp} benchmark. We observe a reduction in the number of utilized gig workers across all algorithms compared to the Pert.($\epsilon=0$) benchmark. For our algorithm, the lower percentage of utilized gig workers relative to Pert. ($\epsilon=0$) is likely due to an inability to sufficiently incentivize gig workers, particularly for requests where utilities are overestimated, specifically those with dropoff location 1. Despite this, our algorithm demonstrates a similar average difference in offered compensation for accepted requests compared to Pert. ($\epsilon=0$), with mean differences of 2.9 and 3.1 monetary units, respectively. In contrast, the \gls{acr:pp} and \gls{acr:fp} benchmarks show much larger differences of 25.6 and 18.3 monetary units, respectively. This large discrepancy in offered compensation results directly from the benchmarks ignoring gig worker utilities when making decisions.

\noindent \textbf{Result 6:} For the NYT scenario, our algorithm balances compensation and utilization, with mean differences in offered compensation of 3.3 units for weak and 3.1 units for strong preferences, effectively avoiding overcompensation.

\begin{figure}[t]%
    \centering
    \fontsize{10}{10}\selectfont
    \subfloat[\centering NYT with weak location preference gig workers]{\includegraphics[width=0.48\textwidth]{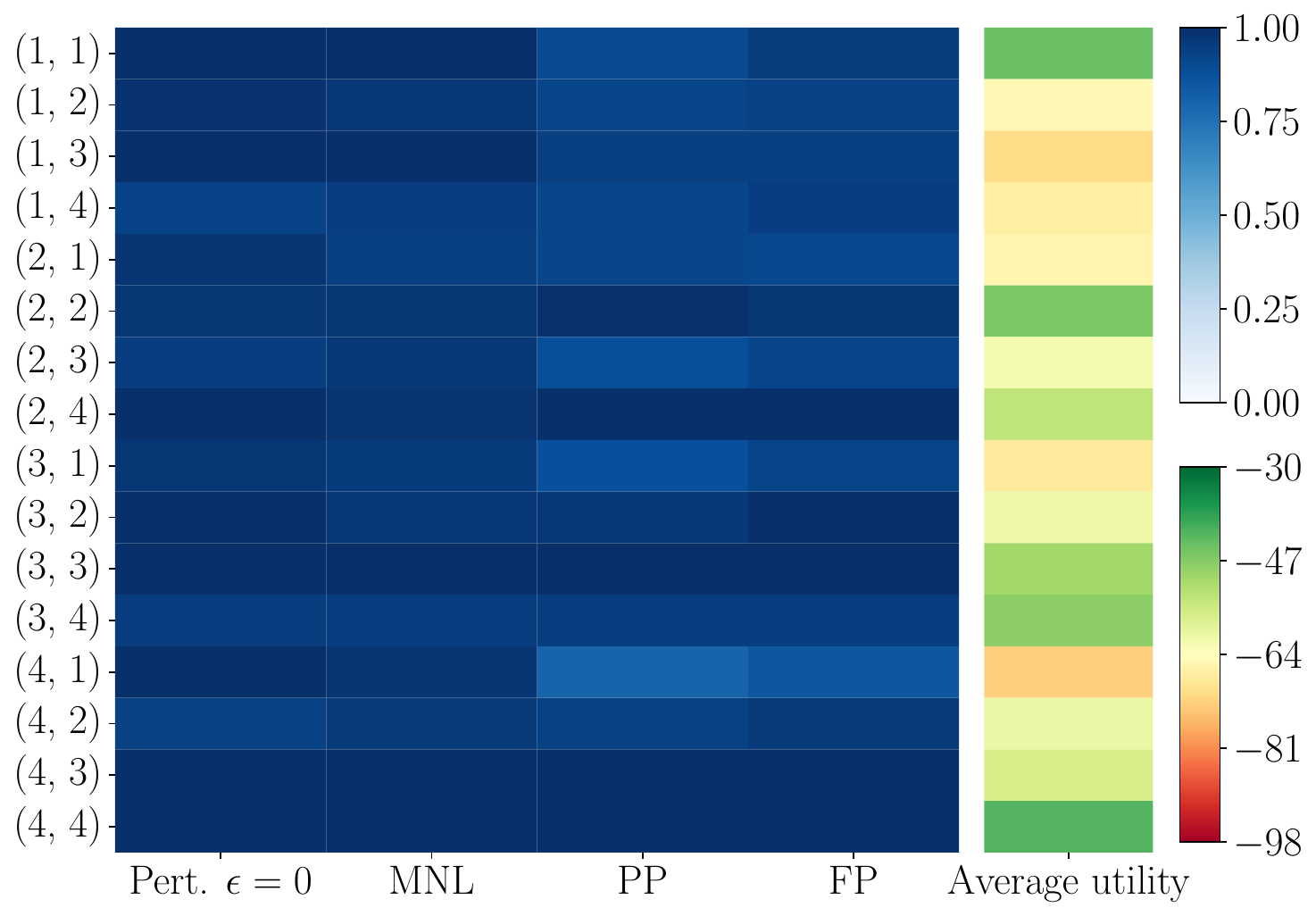}}%
    \hspace*{0.1cm}
    \subfloat[\centering  NYT with strong location preference gig workers]{\includegraphics[width=0.48\textwidth]{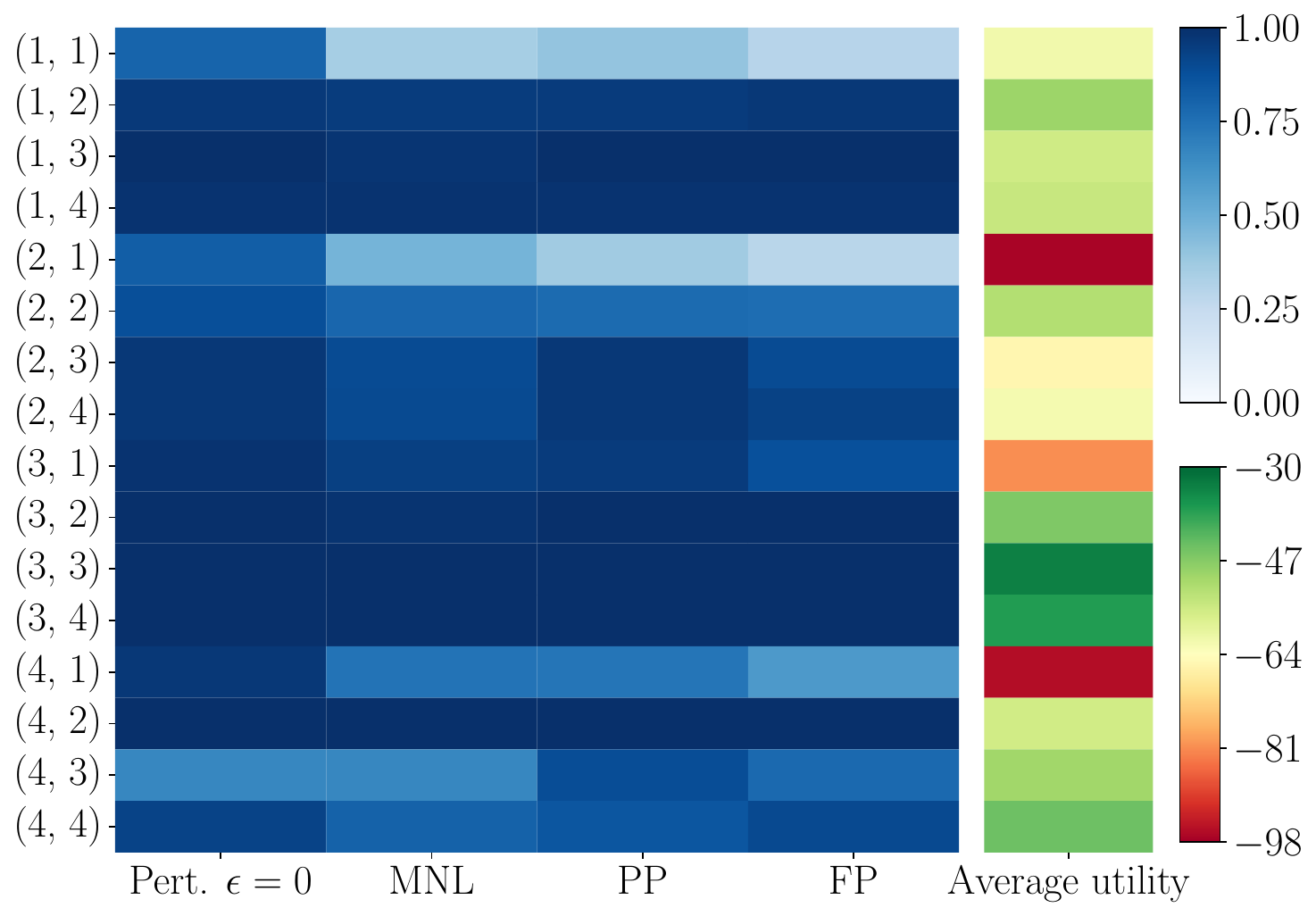}}%
    \caption{\textnormal{Percentage of accepted requests (white-blue matrix) and average utility (right-side column) for each pickup $(p)$ and dropoff $(d)$ location pair under weak (a) and strong (b) location preferences. Each row represents a different location pair $(p,d)$.}}%
    \label{fig:fair_nyt_combined}%
\end{figure}

\noindent \textbf{Acceptance rates across pickup and dropoff locations}: In the following, we examine the acceptance rates of the various algorithms in handling service requests, based on different pairs of pickup and dropoff locations. We evaluate the algorithms by analyzing the distribution of acceptance probabilities across all location pairs. We employ the Coefficient of Variation (CV) to quantify the balance of acceptance rates of each algorithm. A lower CV signifies a more balanced distribution of acceptance rates across locations. In Figure \ref{fig:fair_nyt_combined}, we observe the percentage of accepted requests and the average utility for each pickup $(p)$ and dropoff $(d)$ location pair $(p,d)$ for weak (Figure \ref{fig:fair_nyt_combined}a) and strong (Figure \ref{fig:fair_nyt_combined}b) location preference scenarios.

For the weak preference scenario, the Pert. ($\epsilon=0$) benchmark and our algorithm exhibit a more equal distribution of acceptance probabilities among all location pairs, with a CV equal to 2.4\% and 2.7\% correspondingly, compared to 5.9\% and 4.4\% for the \gls{acr:pp} and \gls{acr:fp} benchmarks. Even though the \gls{acr:pp} and \gls{acr:fp} benchmarks have a slightly higher CV, the acceptance rates remain relatively balanced among all algorithms. When considering the strong preference scenario, we observe a less balanced distribution of acceptance probabilities  across all algorithms. This indicates a strong influence of location preferences on request acceptance and suggests that strong preferences may lead to certain areas becoming underserved. The Pert. ($\epsilon=0$) benchmark achieves the most balanced distribution, with a CV of 10.3\%. Our algorithm and the \gls{acr:pp} benchmark exhibit similar CVs at 23\% and 23.8\%, respectively, while the \gls{acr:fp} benchmark displays the least balanced distribution, with a CV of 28\%. This result raises potential fairness concerns when using different algorithms, particularly when gig workers exhibit strong preferences for specific request characteristics. Notably, the $(p,d)$ pairs with dropoff location 1 show lower acceptance rates for both our algorithm and the \gls{acr:pp} and \gls{acr:fp} benchmarks. This effect is less pronounced in the Pert. ($\epsilon=0$) benchmark, reinforcing our earlier observation regarding the difficulty in incentivizing gig workers for certain requests. Consequently, we believe that both the percentage of utilized gig workers and the balance of acceptance rates across pickup and dropoff locations can be improved by refining the \gls{acr:mnl} model, ultimately approaching the performance of the Pert. ($\epsilon=0$) benchmark.

\noindent \textbf{Result 7:} For weak location preferences, acceptance rates are consistent across policies (CVs 2.4\%–5.9\%). For strong preferences, Pert. ($\epsilon=0$) achieves the most balanced rates with CV of 10.3\%, while our policy shows a CV of 23\%, comparable to the benchmark policies.

\section{Conclusion}
\label{sec:conc}

This paper presented a general algorithmic paradigm for optimizing revenue in on-demand crowdsourcing platforms by modeling the problem as an \gls{acr:mdp} and using the \gls{acr:mnl} model to capture gig worker behavior. By directly incorporating gig worker preferences, our algorithm balances creating attractive offers to ensure utilization while avoiding overcompensation to maximize platform profits.

Our algorithm consistently outperformed rule-based benchmarks and achieved results close to a perfect-prediction upper bound. Specifically, our algorithm showed improvements of up to 7.5\% in scenarios with homogeneous and of up to 10\% in scenarios with heterogeneous gig worker groups. On real-world data, where gig workers have location preferences, we observe improvements of up to 20\%. With these findings, we demonstrated that our algorithm is effective in multiple scenarios utilizing synthetic data with varying on-demand request arrival rates as well as real-world data. However, our results also reveal that the performance of such a partially model-based algorithm depends on the accuracy of the estimated \gls{acr:mnl} model parameters used to estimate the gig worker utilities. In this context, we show that our algorithm shows stability to parameter perturbations, performing well even when the \gls{acr:mnl} parameters are under underestimated, which can likely happen in practice. 

Future research could refine estimation techniques for the \gls{acr:mnl} parameters and improve robustness to parameter variability. Additionally, future work could investigate fairness issues in compensation strategies, especially concerning important gig worker groups. Future enhancements could also include extending the considered gig worker state to account for the gig worker queue, using attention mechanisms to generate gig worker state context vectors. Additionally, exploring priority-based gig worker queue management strategies as alternatives to FIFO could enhance decision-making.

\noindent
{\section*{Acknowledgements}}
\noindent This work was funded by the Deutsche Forschungsgemeinschaft (DFG, German Research Foundation) - Projektnummer 277991500.

\singlespacing{
\bibliographystyle{model5-names}

%
\newpage
\onehalfspacing
\begin{appendices}
	\normalsize
	\section{Proofs}

\subsection{Proof of Lemma 1:}
\label{sec:p_1}

From Equations \eqref{eq:imm_rew} and \eqref{eq:bell_post}, the Bellman equation of the pre-decision state $S\hspace{-0.1em}_t = (\mathcal{R}\hspace{0.05em}_{t},\mathcal{G}_t)$ at time step $t = 0,1,...,T$ is equal to:
\begin{align}
V_t(\mathcal{R}\hspace{0.05em}_{t},\mathcal{G}_t) = & \max_{\mathbf{c_t}} \{ \! \sum_{i \in \mathcal{R}\hspace{0.05em}_t^{\mathrm{exp}}}\!\beta_i + (1 - \mathbbm{1}_{|\mathcal{G}_t|=1})  V^{\mathrm{p}}_{t}(\mathcal{R}\hspace{0.05em}_t^{'}) + \mathbbm{1}_{|\mathcal{G}_t|=1}\sum_{i \in \mathcal{R}\hspace{0.05em}_{t}}{P^i_{t}(\mathbf{c_t})(r_i - c_t^i - \beta_i \mathbbm{1}_{i \in \mathcal{R}\hspace{0.05em}_t^{\mathrm{exp}}}}) \notag \\ 
& + \mathbbm{1}_{|\mathcal{G}_t|=1} \! \sum_{i \in \mathcal{R}\hspace{0.05em}_{t} \cup \{\emptyset\}} \! P^i_{t}(\mathbf{c_t}) V^{\mathrm{p}}_{t}(\mathcal{R}\hspace{0.05em}_t^{'} \setminus \{i\}) \}
\end{align}

\noindent We can reformulate this expression as follows: 
\begin{align}
V_t(\mathcal{R}\hspace{0.05em}_{t},\mathcal{G}_t) = & \max_{c_t} \{ \mathbbm{1}_{|\mathcal{G}_t|=1} [ \sum_{i \in \mathcal{R}\hspace{0.05em}_{t}}P^i_{t}(\mathbf{c_t})(r_i - c_t^i - \beta_i \mathbbm{1}_{i \in \mathcal{R}\hspace{0.05em}_t^{\mathrm{exp}}}) +  \sum_{i \in \mathcal{R}\hspace{0.05em}_{t}}P^i_{t}(\mathbf{c_t}) V^{\mathrm{p}}_{t}(\mathcal{R}\hspace{0.05em}_t^{'} \setminus \{i\}) + P^\emptyset_{t}(\mathbf{c_t}) V^{\mathrm{p}}_{t}(\mathcal{R}\hspace{0.05em}_t^{'}) ] \notag \\
&  +  (1 - \mathbbm{1}_{|\mathcal{G}_t|=1}) V^{\mathrm{p}}_{t}(\mathcal{R}\hspace{0.05em}_t^{'}) \} + \sum_{i \in \mathcal{R}\hspace{0.05em}_t^{\mathrm{exp}}}\beta_i \label{eq:using_p0}
\end{align}

\noindent We express $P^\emptyset_{t}(c_t^i)$ as $ 1 - \sum_{i \in \mathcal{R}\hspace{0.05em}_{t}}P^i_{t}(\mathbf{c_t})$ and therefore using Equation \eqref{eq:using_p0} we have:
\begin{align}
V_t(\mathcal{R}\hspace{0.05em}_{t},\mathcal{G}_t) = & max_{c_t} \{ \mathbbm{1}_{|\mathcal{G}_t|=1} [ \sum_{i \in \mathcal{R}\hspace{0.05em}_{t}}P^i_{t}(\mathbf{c_t})(r_i - c_t^i - \beta_i \mathbbm{1}_{i \in \mathcal{R}\hspace{0.05em}_t^{\mathrm{exp}}}) + \sum_{i \in \mathcal{R}\hspace{0.05em}_{t}}P^i_{t}(\mathbf{c_t}) V^{\mathrm{p}}_{t}(\mathcal{R}\hspace{0.05em}_t^{'} \setminus \{i\}) +  (1 - \sum_{i \in \mathcal{R}\hspace{0.05em}_{t}}P^i_{t}(\mathbf{c_t})) V^{\mathrm{p}}_{t}(\mathcal{R}\hspace{0.05em}_t^{'}) ] \notag \\
&   + (1 - \mathbbm{1}_{|\mathcal{G}_t|=1}) V^{\mathrm{p}}_{t}(\mathcal{R}\hspace{0.05em}_t^{'}) +\sum_{i \in \mathcal{R}\hspace{0.05em}_t^{\mathrm{exp}}} \beta_i \}  \\
& = max_{c_t} \{ \mathbbm{1}_{|\mathcal{G}_t|=1} \sum_{i \in \mathcal{R}\hspace{0.05em}_{t}}P^i_{t}(\mathbf{c_t})(r_i - c_t^i - \beta_i \mathbbm{1}_{i \in \mathcal{R}\hspace{0.05em}_t^{\mathrm{exp}}} + V^{\mathrm{p}}_{t}(\mathcal{R}\hspace{0.05em}_t^{'} \setminus \{i\}) - V^{\mathrm{p}}_{t}(\mathcal{R}\hspace{0.05em}_t^{'})) +  V^{\mathrm{p}}_{t}(\mathcal{R}\hspace{0.05em}_t^{'}) + \sum_{i \in \mathcal{R}\hspace{0.05em}_t^{\mathrm{exp}}}\! \beta_i \} \\
& = max_{c_t} \{ \mathbbm{1}_{|\mathcal{G}_t|=1} \sum_{i \in \mathcal{R}\hspace{0.05em}_{t}}P^i_{t}(\mathbf{c_t})(r_i - c_t^i - \beta_i \mathbbm{1}_{i \in \mathcal{R}\hspace{0.05em}_t^{\mathrm{exp}}} - \Delta^i_{V_{t}}(\mathcal{R}\hspace{0.05em}_t^{'}) ) + V^{\mathrm{p}}_{t}(\mathcal{R}\hspace{0.05em}_t^{'}) + \sum_{i \in \mathcal{R}\hspace{0.05em}_t^{\mathrm{exp}}}\! \beta_i \}
\end{align}

\noindent Therefore: 
\begin{align} 
V_t(\mathcal{R}\hspace{0.05em}_{t},\mathcal{G}_t) = \max_{c_t} \{ \phi_t(\mathcal{R}\hspace{0.05em}_{t},\mathcal{G}_t,\mathbf{c_t}) \} + V^{\mathrm{p}}_{t}(\mathcal{R}\hspace{0.05em}_t^{'})  + \sum_{i \in \mathcal{R}\hspace{0.05em}_t^{\mathrm{exp}}}\! \beta_i \end{align}
where $\phi_t(\mathcal{R}\hspace{0.05em}_{t},\mathcal{G}_t,\mathbf{c_t}) = \mathbbm{1}_{|\mathcal{G}_t|=1} \mathbb{E}_{i \sim P_t(\mathbf{c_t})} [r_i - c_t^i - \Delta^i_{V_{t}}(\mathcal{R}\hspace{0.05em}_t^{'}) -\beta_i\mathbbm{1}_{i \in \mathcal{R}\hspace{0.05em}_t^{\mathrm{exp}}}]$  and $ \Delta^i_{V_{t}}(\mathcal{R}\hspace{0.05em}_t^{'}) = V^{\mathrm{p}}_{t}(\mathcal{R}\hspace{0.05em}_t^{'}) - V^{\mathrm{p}}_{t}(\mathcal{R}\hspace{0.05em}_t^{'} \setminus \{i\})$

\subsection{Proof of Lemma 2:}
\label{sec:p_2}

Using expression \ref{eq:mnl3} we reformulate $\phi_t$ as a function of $P_t$ as follows:
\begin{align}
\phi_t(\mathcal{R}\hspace{0.05em}_{t},\mathcal{G}_t, P_t) = \mathbbm{1}_{|\mathcal{G}_t|=1} [\sum_{i \in \mathcal{R}\hspace{0.05em}_{t}}P^i_{t}(r_i + u_{ij} - u_0 - \mu_j \ln{P_t^i} + \mu_j \ln{P_t^{\emptyset}}  - \Delta^i_{V_{t}}(\mathcal{R}\hspace{0.05em}_t^{'}) -\beta_i\mathbbm{1}_{i \in \mathcal{R}\hspace{0.05em}_t^{\mathrm{exp}}} ) ]
\end{align}

\noindent The gradient of $\phi_t$ is given by:
\begin{align} \frac{\partial \phi_t}{\partial P^i_t} = \mathbbm{1}_{|\mathcal{G}_t|=1} (r_i + u_{ij} - u_0 - \mu_j \ln{P^i_t} + \mu_j \ln{P^\emptyset_t} - \Delta^i_{V_{t}}(\mathcal{R}\hspace{0.05em}_t^{'}) - \beta_i \mathbbm{1}_{i \in \mathcal{R}\hspace{0.05em}_t^{\mathrm{exp}}} - \mu_j) \text{, } \forall i \in \mathcal{R}\hspace{0.05em}_{t}, \end{align}
\begin{align} \frac{\partial \phi_t}{\partial P^\emptyset_t} = \frac{\mathbbm{1}_{|\mathcal{G}_t|=1} \mu_j \sum_{i \in \mathcal{R}\hspace{0.05em}_{t}} P^i_t}{P^\emptyset_t} \end{align}

\noindent Additionally:
\begin{align} \frac{\partial^2 \phi_t}{\partial P^i_t \partial P^i_t} = -\frac{\mathbbm{1}_{|\mathcal{G}_t|=1} \mu_j}{P^i_t} \text{, } \forall i \in \mathcal{R}\hspace{0.05em}_{t}, \end{align} 
\begin{align} \frac{\partial^2 \phi_t}{\partial P^i_t \partial P^j_t} = 0 \text{, } \forall i,j \in \mathcal{R}\hspace{0.05em}_{t}: i \neq j,\end{align}
\begin{align} \frac{\partial^2 \phi_t}{\partial P^\emptyset_t \partial P^\emptyset_t} = -\frac{\mathbbm{1}_{|\mathcal{G}_t|=1} \mu_j \sum_{i \in \mathcal{R}\hspace{0.05em}_{t}} P^i_t}{(P^\emptyset_t)^2}, \end{align} 
\begin{align} \frac{\partial^2 \phi_t}{\partial P^\emptyset_t \partial P^i_t} = \frac{\mathbbm{1}_{|\mathcal{G}_t|=1} \mu_j}{P^\emptyset_t} \text{, } \forall i \in \mathcal{R}\hspace{0.05em}_{t} \end{align}

\noindent Therefore the Hessian of $-\phi_t$ is given by:
\begin{align} \mathbb{H}_{-\phi_t} = \mathbbm{1}_{|\mathcal{G}_t|=1} \mu_j \begin{bmatrix}
\frac{1}{P^1_t} & 0 & \dots & 0 & -\frac{1}{P^\emptyset_t} \\
0 & \frac{1}{P^2_t} & \dots & 0 & -\frac{1}{P^\emptyset_t} \\
\vdots &  & \ddots &  & \vdots \\
0 & 0 & \dots & \frac{1}{P^{|\mathcal{R}\hspace{0.05em}_{t}|}_t} & -\frac{1}{P^\emptyset_t} \\
-\frac{1}{P^\emptyset_t} & -\frac{1}{P^\emptyset_t} & \dots & -\frac{1}{P^\emptyset_t} & \frac{\sum_{i \in \mathcal{R}\hspace{0.05em}_{t}} P^i_t}{(P^\emptyset_t)^2} 
\end{bmatrix}
\end{align}
\noindent Therefore:
\begin{align} y\mathbb{H}_{-\phi_t}y^T = \mathbbm{1}_{|\mathcal{G}_t|=1} \mu_j ( \sum_{i \in \mathcal{R}\hspace{0.05em}_{t}}\frac{y_i^2}{P^i_t} + \frac{y_{|\mathcal{R}\hspace{0.05em}_{t}|+1}}{(P^\emptyset_t)^2}(\sum_{i \in \mathcal{R}\hspace{0.05em}_{t}} P^i_t)) \geq 0 \text{ } \forall y \in \mathbb{R}^{|\mathcal{R}\hspace{0.05em}_{t}|+1} \end{align} 
As a result, we conclude that $\phi_t$ is a concave function of $P_t$ for $P_t \in (0,1)^{|\mathcal{R}\hspace{0.05em}_{t}|+1}$. Since $ \Psi = \{ (P^1,...,P^{|\mathcal{R}\hspace{0.05em}_{t}|},P^\emptyset) \in (0,1)^{|\mathcal{R}\hspace{0.05em}_{t}|+1} : \sum_{i \in \mathcal{R}\hspace{0.05em}_{t}} P^i_t = 1 \}$ is a convex subset of $(0,1)^{|\mathcal{R}\hspace{0.05em}_{t}|+1}$, therefore $\phi_t $ is concave in $\Psi$.

\subsection{Proof of Lemma 3:}
\label{sec:p_3}
To find the optimal $P^*_t \in \Psi$ that maximizes $\phi_t(\mathcal{R}\hspace{0.05em}_{t},\mathcal{G}_t,P_t)$ we solve the first order condition. To do so we replace $P^\emptyset_{t}(c_t^i)$ by $1 - \sum_{i \in \mathcal{R}\hspace{0.05em}_{t}}P^i_{t}(\mathbf{c_t})$ and take the gradient of $\phi_t(\mathcal{R}\hspace{0.05em}_{t},\mathcal{G}_t,P_t)$:
\begin{align}
\frac{\partial \phi_t}{\partial P^i_t} = & \mathbbm{1}_{|\mathcal{G}_t|=1} [r_i + u_{ij} - u_0 - \mu_j \ln{P^i_t} - \Delta^i_{V_{t}}(\mathcal{R}\hspace{0.05em}_t^{'}) + \mu_j \ln{(1 - \sum_{i \in \mathcal{R}\hspace{0.05em}_{t}}P_t^i)} - \beta_i \mathbbm{1}_{i \in \mathcal{R}\hspace{0.05em}_t^{\mathrm{exp}}} - \mu_j + \sum_{i \in \mathcal{R}\hspace{0.05em}_{t}}P_t^i( -\frac{\mu_j}{1 - \sum_{i \in \mathcal{R}\hspace{0.05em}_{t}}P_t^i}) ] \\
& = \mathbbm{1}_{|\mathcal{G}_t|=1} [r_i + u_{ij} - u_0 - \mu_j \ln{P^i_t} - \Delta^i_{V_{t}}(\mathcal{R}\hspace{0.05em}_t^{'}) + \mu_j \ln{(1 - \sum_{i \in \mathcal{R}\hspace{0.05em}_{t}}P_t^i)}  - \beta_i \mathbbm{1}_{i \in \mathcal{R}\hspace{0.05em}_t^{\mathrm{exp}}} -\frac{\mu_j}{1 - \sum_{i \in \mathcal{R}\hspace{0.05em}_{t}}P_t^i} ]
\end{align}

\noindent Therefore:
\begin{align}
\frac{\partial \phi_t}{\partial P^i_t} = 0  \Leftrightarrow & P^{i*}_t = \frac{\mu_j}{m_t} \exp\{ \frac{1}{\mu_j} (r_i + u_{ij} - u_0 - \Delta^i_{V_{t}}(\mathcal{R}\hspace{0.05em}_t^{'}) - \beta_i \mathbbm{1}_{i \in \mathcal{R}\hspace{0.05em}_t^{\mathrm{exp}}}  - m_t)\}, \forall i \in \mathcal{R}\hspace{0.05em}_{t} \label{eq:pi} 
\end{align}

\noindent where: 
\begin{align}
m_t = \frac{\mu_j}{1 - \sum_{i \in \mathcal{R}\hspace{0.05em}_{t}}P_t^{i*}} = \frac{\mu_j}{P_t^{\emptyset*}} \label{eq:mt1}
\end{align}
\noindent We have:
\begin{align}
& \sum_{i \in \mathcal{R}\hspace{0.05em}_{t}}P_t^{i*} =  1 - P_t^{\emptyset*} = 1 - \frac{\mu_j}{m_t} \Leftrightarrow \\ 
& 1 - \frac{\mu_j}{m_t} = \frac{\mu_j}{m_t} \sum_{i \in \mathcal{R}\hspace{0.05em}_{t}} \exp\{ \frac{1}{\mu_j} (r_i + u_{ij} - u_0 - \beta_i \mathbbm{1}_{i \in \mathcal{R}\hspace{0.05em}_t^{\mathrm{exp}}} - \Delta^i_{V_{t}}(\mathcal{R}\hspace{0.05em}_t^{'}) - m_t)\} \Leftrightarrow \\
&(\frac{m_t}{\mu_j} - 1) \exp\{ \frac{m_t}{\mu_j} - 1\} = \sum_{i \in \mathcal{R}\hspace{0.05em}_{t}} \exp\{ \frac{1}{\mu_j} (r_i + u_{ij} - u_0 - \beta_i \mathbbm{1}_{i \in \mathcal{R}\hspace{0.05em}_t^{\mathrm{exp}}} - \Delta^i_{V_{t}}(\mathcal{R}\hspace{0.05em}_t^{'}) - \mu_j)\} \Leftrightarrow \\
& m_t =  \mu_j \left( W_0(\sum_{i \in \mathcal{R}\hspace{0.05em}_{t}} \exp\{ \frac{1}{\mu_j} (r_i + u_{ij} - u_0 - \beta_i \mathbbm{1}_{i \in \mathcal{R}\hspace{0.05em}_t^{\mathrm{exp}}} - \Delta^i_{V_{t}}(\mathcal{R}\hspace{0.05em}_t^{'}) - \mu_j)\}) + 1 \right) \label{eq:mt2}
\end{align}

\noindent where $W_0$ is the Lambert's W function for k = 0.

\noindent From Equation \ref{eq:mnl3} using Equations \ref{eq:pi} and \ref{eq:mt1} we result in:
\begin{align}
& c^{i*}_t = -u_{ij} + u_0 - \mu_j \ln( \frac{\mu_j}{m_t} ) + \mu_j \ln ( \frac{\mu_j}{m_t} \exp\{ \frac{1}{\mu_j} (r_i + u_{ij} - u_0 - \beta_i \mathbbm{1}_{i \in \mathcal{R}\hspace{0.05em}_t^{\mathrm{exp}}} - \Delta^i_{V_{t}}(\mathcal{R}\hspace{0.05em}_t^{'}) - m_t)\} )  \Rightarrow \\
& c^{i*}_t = r_i - \beta_i \mathbbm{1}_{i \in \mathcal{R}\hspace{0.05em}_t^{\mathrm{exp}}} - \Delta^i_{V_{t}}(\mathcal{R}\hspace{0.05em}_t^{'}) - m_t
\end{align}

\noindent We have therefore derived an analytical solution. 

\section{Experimental Design}

\subsection{Neural Network Training Details}
\label{sec:nn_app}

In our implementation, we incorporate a target network parameterized by $\hat{\theta}$ for the post-decision value function approximation. This mitigates oscillating or divergent updates that can occur with frequent updates. The target network's parameters ($\hat{\theta}$) are periodically synchronized with those of the main network ($\theta$) at intervals defined by \(K\) (see Table \ref{tab:training-details}), leading to smoother value function updates and improved convergence during training.

\begin{table}[h]
    \centering
    \caption{Neural Network Training Details}
    \begin{tabular}{ll}
        \hline
        \textbf{Parameter} & \textbf{Value} \\
        \hline
        Optimizer (clip value) & Adam (0.5) \\
        Loss function & Huber \\
        Network size & [16,16] \\
        Attention embedding size & 32 \\
        Activation function & Swish \\
        Discount factor & 0.95 \\
        Batch size & 512 \\
        Initial learning rate & $1 \times 10^{-5}$ \\
        Learning rate decay rate & $1 \times 10^{-2}$ \\
        Learning rate decay steps & 10000 \\
        Epochs & 30 \\
        C ($\theta$ update frequency) & $80 \times 50$ \\
        K ($\hat{\theta}$ update frequency) & $80 \times 50 \times 5$ \\
        Initial $\delta$ (e-greedy parameter) & 10 \\
        $\delta$ decay per time step & $1 \times 10^{-4}$ \\
        \hline
    \end{tabular}
    \label{tab:training-details}
\end{table}

\subsection{Formula-based compensation (FP)}
\label{sec:fbp_app}

\begin{table}[h]
  \centering
  \caption{Weights for each attribute in the formula-based compensation strategy}
  \begin{tabular}{lc}
    \hline
    \textbf{Attribute} & \textbf{Weights} \\
    \hline
    Reward       & \( \{ 0, 0.5, 0.55, 0.6, 0.65, 0.7, 0.75, 0.8, 0.85, 0.9, 0.95 \} \) \\
    Distance     & \( \{ 0, 5, 10, 15, 20 \} \) \\
    Penalty      & \( \{ -0.1, -0.05, 0, 0.05, 0.1 \} \) \\
    Exp. Boost   & \( \{ 0, 0.1, 0.2, 0.3 \} \) \\
    \hline
  \end{tabular}
  \label{table:fbp_val}
\end{table}

\subsection{Full information solution}
\label{sec:fi_app}

Consider $\mathcal{R}$ and $\mathcal{G}$ to be the sets of all requests and gig workers that arrive within the full time horizon. Let $r_i,\beta_i, t_i^{\mathrm{arr}}, t_i^{\mathrm{exp}}$ be the reward, penalty, arrival time and expiration time of request $i \in \mathcal{R}$, $t_j^{\mathrm{arr}}$ be the arrival time of gig worker $ j \in \mathcal{G}$. Lastly, let $u_{ij} + \hat{e}_j$ be the attractiveness of request $i$ for gig worker $j$, where $\hat{e}_j$ is the realization of the gig worker's utility noise variable $e_j$. The full information solution is given by solving the following MILP: 
\begin{align}
    \text{maximize} \quad & \sum_{i \in \mathcal{R}} \sum_{j \in \mathcal{G}} \left( r_i - \beta_i - u_{ij} -  \hat{e}_j \right) x_{ij}  \\
    \text{subject to} \quad & \sum_{j \in \mathcal{G}} x_{ij} \leq 1 \quad \forall i \in \mathcal{R}  \\ 
    & \sum_{i \in \mathcal{R}} x_{ij} \leq 1 \quad \forall j \in \mathcal{G} \\ 
    & t_j^{\mathrm{arr}} x_{ij} \leq t_i^{\mathrm{exp}} \quad \forall i \in \mathcal{R}, \forall j \in \mathcal{G} \\ 
    & t_i^{\mathrm{arr}} x_{ij} \leq t_j^{\mathrm{arr}} \quad \forall i \in \mathcal{R}, \forall j \in \mathcal{G} \\ 
    & x_{ij} \in \{0, 1\} \quad \forall i \in \mathcal{R}, \forall j \in \mathcal{G}
\end{align}

\subsection{Mean Bias Error calculation and interpretation}
\label{sec:mbe}

For a set of requests $I$, we calculate the \gls{acr:mbe} between the predicted utilities of requests $ \{ \hat{u}_i \}_{i \in I} $ and the true  of requests $ \{ u_i \}_{i \in I} $ as:
\begin{align}
\text{MBE}(I) = \frac{1}{|I|} \sum_{i \in I} (u_i - \hat{u}_i)
\end{align}
A positive $\text{MBE}$ indicates that, on average, the estimated utility is lower than the true utility of the gig workers, suggesting an underestimation of the requests' attractiveness. Conversely, a negative $\text{MBE}$ suggests an overestimation of the attractiveness.
\end{appendices}
\end{document}